\documentclass[10pt]{amsart}

\usepackage[letterpaper,margin=0.85in]{geometry}
\usepackage{amsmath,amssymb,amsthm,mathtools}
\usepackage{bm}
\usepackage{booktabs}
\usepackage{enumitem}
\usepackage{microtype}
\usepackage{longtable}
\usepackage{xcolor}
\usepackage{tikz}
\usetikzlibrary{arrows.meta,positioning,fit,calc}
\usepackage{hyperref}
\usepackage{aliascnt}
\usepackage[nameinlink,capitalise]{cleveref}

\allowdisplaybreaks
\numberwithin{equation}{section}
\setlist{itemsep=0.2em,topsep=0.35em}
\hypersetup{
  colorlinks=true,
  linkcolor=blue!55!black,
  citecolor=blue!55!black,
  urlcolor=blue!55!black,
  pdftitle={Grokking and the Dynamics of Generalization},
  pdfauthor={Malik Amir}
}

\newtheorem{theorem}{Theorem}[section]
\newaliascnt{proposition}{theorem}
\newtheorem{proposition}[proposition]{Proposition}
\aliascntresetthe{proposition}
\newaliascnt{lemma}{theorem}

\aliascntresetthe{lemma}
\newaliascnt{corollary}{theorem}
\newtheorem{corollary}[corollary]{Corollary}
\aliascntresetthe{corollary}
\newaliascnt{quotedtheorem}{theorem}

\aliascntresetthe{quotedtheorem}
\theoremstyle{definition}
\newaliascnt{definition}{theorem}
\newtheorem{definition}[definition]{Definition}
\aliascntresetthe{definition}
\newaliascnt{assumption}{theorem}
\newtheorem{assumption}[assumption]{Standing assumption}
\aliascntresetthe{assumption}
\newaliascnt{example}{theorem}
\newtheorem{example}[example]{Example}
\aliascntresetthe{example}
\newaliascnt{problem}{theorem}
\newtheorem{problem}[problem]{Problem}
\aliascntresetthe{problem}
\theoremstyle{remark}
\newaliascnt{remark}{theorem}
\newtheorem{remark}[remark]{Remark}
\aliascntresetthe{remark}

\crefname{theorem}{Theorem}{Theorems}
\Crefname{theorem}{Theorem}{Theorems}
\crefname{proposition}{Proposition}{Propositions}
\Crefname{proposition}{Proposition}{Propositions}
\crefname{lemma}{Lemma}{Lemmas}
\Crefname{lemma}{Lemma}{Lemmas}
\crefname{corollary}{Corollary}{Corollaries}
\Crefname{corollary}{Corollary}{Corollaries}
\crefname{quotedtheorem}{Quoted theorem}{Quoted theorems}
\Crefname{quotedtheorem}{Quoted theorem}{Quoted theorems}
\crefname{definition}{Definition}{Definitions}
\Crefname{definition}{Definition}{Definitions}
\crefname{assumption}{Standing assumption}{Standing assumptions}
\Crefname{assumption}{Standing assumption}{Standing assumptions}
\crefname{example}{Example}{Examples}
\Crefname{example}{Example}{Examples}
\crefname{problem}{Problem}{Problems}
\Crefname{problem}{Problem}{Problems}
\crefname{remark}{Remark}{Remarks}
\Crefname{remark}{Remark}{Remarks}

\newcommand{\R}{\mathbb R}
\newcommand{\C}{\mathbb C}
\newcommand{\E}{\mathbb E}
\newcommand{\HS}{\mathcal H_S}
\newcommand{\HE}{\mathcal H_E}
\newcommand{\HP}{\mathcal H_\theta}

\newcommand{\Task}{\mathcal T}
\newcommand{\Id}{I}
\newcommand{\eps}{\varepsilon}
\newcommand{\dd}{\,\mathrm d}
\newcommand{\ip}[2]{\left\langle #1,#2\right\rangle}
\newcommand{\norm}[1]{\left\lVert #1\right\rVert}
\newcommand{\abs}[1]{\left\lvert #1\right\rvert}
\newcommand{\op}{\mathrm{op}}
\newcommand{\tr}{\operatorname{tr}}
\newcommand{\rank}{\operatorname{rank}}
\newcommand{\ran}{\operatorname{ran}}
\newcommand{\Ker}{\operatorname{ker}}

\newcommand{\1}{\mathbf 1}
\newcommand{\Pg}{\mathcal P}
\newcommand{\W}{{W_S^M}}
\newcommand{\G}{{G_E^M}}
\newcommand{\Salpha}{\mathsf S_\alpha}
\newcommand{\Ralpha}{\mathsf R_\alpha}
\newcommand{\Risk}{\mathcal R}
\newcommand{\Loss}{\mathcal L}

\newcommand{\Prog}{\mathfrak P}
\newcommand{\Eff}{\mathfrak e}
\newcommand{\Mob}{\mathfrak m}
\newcommand{\Frob}{\mathrm F}
\newcommand{\Geo}{\mathcal G}
\newcommand{\Err}{\operatorname{err}}

\DeclareMathOperator*{\argmax}{arg\,max}



\title[Grokking and Generalization]
{On Grokking and the Dynamics of Generalization}

\author{Malik Amir}

\thanks{Faculty of Medicine, Université de Montréal,
Montreal, Quebec, Canada.}

\thanks{Imaging and Engineering Axis, CRCHUM,
Montreal, Quebec, Canada.}

\thanks{Email: \href{mailto:malik.amir.math@gmail.com}
{malik.amir.math@gmail.com}.}

\thanks{Google Scholar : \href{https://scholar.google.com/citations?user=uG67CWYAAAAJ}
{Malik Amir}.}

\begin{document}

\begin{abstract}
We continue the development of a theory of grokking
based on the evolution of the signal channel introduced
by Litman and Guo. The theory distinguishes the
dissipation of training error from its transfer to
evaluation outputs and relates both to the evolution
of population risk. For preconditioned gradient flows
with time-dependent kernels, we show that a direction
in the training-output space may remain nearly invisible
under an evaluation distribution either because its
cumulative dissipation is small or because its transfer
to the evaluation space is weak. We obtain quantitative
bounds for the visibility of task subspaces and study
their stability under signal filtering and changes of
evaluation distribution. We describe how directions
initially lying outside the effective signal channel
may acquire a component in it through eigenvalue
activation or the rotation of spectral subspaces.
The emergence of a visible signal does not, in itself,
imply generalization. We establish necessary and
sufficient conditions for the transferred signal to
decrease squared population risk and derive bounds
for classification accuracy under margin assumptions.
In particular, we give conditions under which the
dependence of the transfer on the evaluation
distribution allows the same training trajectory
to generalize under one distribution while failing
under another. We prove that an arbitrarily long
delay between achieving low training loss and low
population risk can arise from differences in the
learning rates of spectral modes. We relate the
evolution of classification accuracy to the distribution
of the times at which classification margins change
sign. Using this relation, we construct a family of
gradient flows in which the delay between achieving
low training loss and high evaluation accuracy tends
to infinity, while the interval over which classification
accuracy increases from zero to one tends to zero.
Moreover, the variation of population risk over this
interval tends to zero. These phenomena occur in the
absence of feature evolution during training and
result from the separation of spectral learning rates
and the resulting concentration of margin crossing
times. Finally, we show that the occurrence and timing
of generalization cannot, in general, be inferred from
the training dynamics alone. They depend also on the
geometry of the transfer from the training space
to the evaluation space and on its relation to the
target function.
\end{abstract}

\maketitle

\tableofcontents

\clearpage

\section{The unresolved problem}
\label{sec:introduction}

An overparameterized model may fit its training data long before
it generalizes. This phenomenon, known as \emph{grokking},
was observed by Power et al.~\cite{power2022grokking}.
Several explanations have been proposed, based on weight decay
and norm dynamics~\cite{liu2022effective}, transitions between
learning regimes~\cite{kumar2024lazy}, the formation of circuits
\cite{nanda2023progress}, phase transitions~\cite{rubin2024phase},
optimizer dynamics~\cite{acharya2026variance}, and differences
between the rates of representation learning and readout
calibration~\cite{chou2026twospeeds,tan2026twoclocks}.

Litman and Guo~\cite{litmanguo2026} study grokking in the
space of training outputs. They introduce a cumulative
dissipation operator \(W_S\), which determines a decomposition
into a signal channel and a reservoir. The reservoir is
invisible to every compatible train-to-test transfer operator.
Starting from this description, we study the evolution
of the effective signal channel relative to a given
task, the dependence of its visibility on the evaluation
law, and the conditions under which the resulting
displacement decreases population risk.

We consider the following problem.

\begin{samepage}
\begin{problem}[Reservoir-to-generalization problem]
\label{prob:main}
Let a task direction initially lie outside the effective
signal channel. We seek to determine
\begin{enumerate}[label=(\roman*)]
\item how the direction enters the effective signal channel,
\item when its effect on evaluation outputs reaches a given threshold,
\item under what conditions this effect decreases population risk.
\end{enumerate}
\end{problem}
\end{samepage}

The last question cannot be answered from visibility alone.
Indeed, the norm of an evaluation displacement does not
determine its effect on population risk. The direction of
the displacement relative to the target must also be taken
into account.

We study these questions for preconditioned gradient
flows with time-dependent kernels. We extend the
reservoir theorem to general evaluation distributions
and obtain a factorization of normalized squared
evaluation displacement into training mobility and
transfer efficiency
(\cref{thm:reservoir,thm:domination,cor:factorization}).
A resolvent filter gives a continuous approximation of
the signal projection, with a quantitative error bound
(\cref{thm:resolvent}). We then obtain two-sided bounds
for the visibility of a task subspace. These bounds depend
on task alignment, visibility within and outside the
subspace, and coupling between the two
(\cref{thm:block-task,cor:task-gap,thm:soft-task-stability}).

We next relate visibility to population risk. For squared
loss, we derive an exact identity for risk variation and
a necessary and sufficient condition for its decrease.
Under suitable margin assumptions, this leads to bounds
on classification error
(\cref{thm:risk-balance,thm:classification}).
We also give sufficient conditions under which a single
training trajectory generalizes for one evaluation
distribution but not for another
(\cref{thm:abstract-mu-not-nu,thm:explicit-mu-not-nu}).

To study the time dependence, we derive an evolution
formula for evaluation visibility. It separates direct
signal transfer from changes in the effective signal
channel. The latter can arise through eigenvalue
activation or through the rotation of eigenspaces
(\cref{thm:endpoint-balance,thm:projector-transport}).
We give a two-mode model in which the separation
between the times required to achieve low training
loss and low population risk can be arbitrarily
large (\cref{thm:two-mode}). We then extend this
construction to a binary classification problem
and establish the existence of a family of gradient
flows for which the delay before generalization
tends to infinity while the interval over which
classification accuracy increases from zero to
one tends to zero. The variation of population
risk over this interval also tends to zero
(\cref{thm:two-mode-classification,cor:two-mode-classification-risk}).

Finally, \cref{sec:structures} describes the algebraic
and geometric structures associated with the theory.
These include the composition of training intervals,
quotient signal spaces, observability, Grassmannian
transport, symmetries, and determinantal degeneracy loci.

Our approach is related to recent work on spectral
transitions~\cite{xu2026spectral} and feature emergence
\cite{tian2025li2}. We consider these phenomena from
the point of view of the transfer operators associated
with a training trajectory, with particular attention
to their dependence on the evaluation law and to
the effect of the transferred signal on population risk.

\clearpage

\section{Preliminaries}
\label{sec:background}

We recall the definitions and mathematical results
on which the subsequent sections are based. We first
introduce the notions of memorization, delayed grokking,
and generalization under different evaluation laws,
and discuss their relation to spectral bias and
task--model alignment. We then recall the signal-channel
and reservoir decomposition of Litman and Guo, which
provides the starting point of the present work.
Finally, we collect some results from operator theory
concerning spectral perturbations, the evolution of
spectral projectors, and operator factorization.
These results will be used to study the transfer of
training directions to evaluation outputs and the
evolution of the effective signal channel.

\subsection{Grokking, kernels, and task alignment}

The phenomenon known as \emph{grokking} was observed by
Power et al.~\cite{power2022grokking} on small algorithmic
data sets. It consists in a substantial delay between
the fitting of the training data and the improvement
of evaluation accuracy. We use the term independently
of the mechanism responsible for this delay.

\begin{definition}[Memorization and grokking]
\label{def:grokking}
Let \(\Loss_S(t)\) be the empirical loss at time \(t\),
and let \(A_\xi(t)\) be the accuracy under an evaluation
law \(\xi\). Given \(\varepsilon>0\) and \(0<\delta<1\),
put
\[
    t_{\rm mem}(\varepsilon)
    =\inf\{t\geq 0:\Loss_S(t)\leq\varepsilon\},
\]
and
\[
    t_{\rm grok}^{\xi}(\varepsilon,\delta)
    =\inf\{t\geq t_{\rm mem}(\varepsilon):
                  A_\xi(t)\geq 1-\delta\}.
\]
We use the convention \(\inf\varnothing=+\infty\).

There is \emph{delayed grokking} if
\(t_{\rm grok}^{\xi}(\varepsilon,\delta)\) is finite
and substantially larger than \(t_{\rm mem}(\varepsilon)\).

Let \(\mu\) and \(\nu\) be two evaluation laws.
We say that a trajectory exhibits
\emph{\(\mu\)-grokking without \(\nu\)-grokking}
at time \(T\) if
\[
    A_\mu(T)\geq 1-\delta
    \quad\text{and}\quad
    A_\nu(T)<1-\delta.
\]
\end{definition}

Several mechanisms have been proposed to account for
the difference between the two times in
\cref{def:grokking}. These include effective regularization
\cite[Sections~3.2 and~4]{liu2022effective}, circuit formation
\cite{nanda2023progress}, transitions between learning
regimes~\cite{kumar2024lazy}, and phase transitions
\cite{rubin2024phase}.

Other works distinguish the rates of representation
learning and readout calibration~\cite{chou2026twospeeds},
or the time scales associated with empirical loss and
the structure of the learned representation
\cite{tan2026twoclocks}. Related questions have been studied through the
spectrum of rolling-window parameter-update Gramians
\cite{xu2026spectral} and the emergence of
symmetry-adapted features in group-arithmetic
learning problems~\cite{tian2025li2}.

Our question concerns the relation between a learned
training direction and its effect on evaluation outputs.
We seek to determine when this effect becomes significant
and when it decreases population risk.

The distinction between empirical fit and population
risk is classical. Neural networks can interpolate
random labels~\cite{zhang2017rethinking}, and test error
need not be monotone in the capacity of the model
\cite{belkin2019reconciling} or in the number of
training samples~\cite{nakkiran2020deep}. Under
suitable covariance assumptions, minimum-norm
interpolation in linear regression is compatible with
small excess prediction risk~\cite{bartlett2020benign}.
Our concern here is with the evolution of population
risk along a training trajectory.

Let \(F(\theta,z)\in\R^p\) denote the output of a model
with parameters \(\theta\) at an input \(z\).
For a training set \(S\), let \(J_S(\theta)\) be the
Jacobian of the training outputs with respect to
\(\theta\). The empirical neural tangent kernel is
\[
    K_{SS}(\theta)=J_S(\theta)J_S(\theta)^*.
\]
This kernel plays a central role in the infinite-width
analysis of Jacot, Gabriel, and Hongler
\cite{jacot2018ntk}.

When the Jacobian is approximately constant, the
dynamics are described by the linearization of the
model at initialization~\cite{lee2019wide}.
This is the lazy-learning regime studied by Chizat,
Oyallon, and Bach~\cite{chizat2019lazy}.
We make no such assumption. In what follows,
the Jacobians and kernels are evaluated along the
training trajectory.

For a fixed kernel, the learning of a target depends
on the eigenvalues of the kernel and on the components
of the target in its eigenspaces
\cite{bordelon2020spectrum}. The dependence of the
learning rates on the eigenvalues is referred to as
\emph{spectral bias}. The compatibility of the target
with the kernel, measured by the cumulative
distribution of target power over its eigenfunctions,
is referred to as \emph{task--model alignment}
\cite[Eqs.~(5)--(6)]{canatar2021alignment}.

We shall use a related notion for evolving kernels.
A task is represented by a closed subspace of the
training-output gradient space, chosen according
to the target rule. Its definition does not require
a fixed kernel eigenbasis.

\subsection{The signal channel and the reservoir}

We recall a result of Litman and Guo
\cite[Def.~3.1 and Prop.~3.2]{litmanguo2026},
which is the starting point of the present work.
We state it in finite-dimensional training-output space.

\begin{theorem}[Litman and Guo \cite{litmanguo2026}]
\label{quoted:reservoir}
Let \(S\) and \(Q\) be finite training and evaluation
sets, respectively, and let \(J_S(t)\) and \(J_Q(t)\)
be the corresponding output Jacobians along a common
training trajectory. The associated kernels are
given by
\[
    K_{SS}(t)=J_S(t)J_S(t)^*,
    \qquad
    K_{QS}(t)=J_Q(t)J_S(t)^*.
\]
Let
\[
    g(t)=\Pg(t,s)g(s),
\]
be the propagated output gradient. We assume that
the operator-valued integrands below are Bochner
integrable on \([s,T]\). Following
\cite[Def.~3.1 and Eq.~(11)]{litmanguo2026}, put
\[
    W=\int_s^T
    \Pg(\tau,s)^*K_{SS}(\tau)\Pg(\tau,s)
    \,\dd\tau,
\]
and
\[
    G_Q=\int_s^T K_{QS}(\tau)\Pg(\tau,s)
    \,\dd\tau,
\]
where \(Q\) is a finite evaluation set.

The operator \(W\) is called the cumulative
dissipation operator, and \(G_Q\) is the corresponding
transfer operator. The subspaces
\[
    \ran W
    \qquad\text{and}\qquad
    \Ker W,
\]
are called the signal channel and the reservoir,
respectively. One has
\[
    \Ker W\subseteq\Ker G_Q.
\]

Let \(W^\dagger\) denote the Moore--Penrose
pseudoinverse of \(W\). Following
\cite[Eqs.~(49) and (178)]{litmanguo2026}, define the
normalized visibility Gramian associated with \(Q\) by
\[
    \Gamma_Q=(W^\dagger)^{1/2}
    G_Q^*G_Q(W^\dagger)^{1/2}.
\]
It is positive semidefinite, and one has
\[
    G_Q^*G_Q=W^{1/2}\Gamma_Q W^{1/2}.
\]
In particular,
\[
    G_Q^*G_Q\preceq\|\Gamma_Q\|_{\op}W.
\]
\end{theorem}

Thus every direction in the reservoir is invisible
under transfer to \(Q\). Since \(W\) is self-adjoint
and the training-output space is finite-dimensional,
one has the orthogonal decomposition
\[
    \ran W\oplus\Ker W.
\]

In the fixed-kernel case, Litman and Guo obtain
a spectral filter of the form
\[
    1-\exp\left(-\frac{t\sigma_j^2}{n}\right),
\]
where \(\sigma_j\) is a training singular value
\cite[Thm.~H.1 and Cor.~H.2]{litmanguo2026}.
Modes corresponding to small values of \(\sigma_j\)
are learned on longer time scales. The relation
between this spectral filter and grokking is
discussed in \cite[Rem.~H.6]{litmanguo2026}.

A related question concerns the implicit bias of
gradient descent. For homogeneous linear classification
with separable data and logistic-type losses, the
direction of the parameter vector converges to that
of the hard-margin solution
\cite[Thm.~3]{soudry2018implicit}.

We shall consider how the signal channel evolves
when the kernel is not fixed, how its directions
are transferred to an evaluation distribution,
and how this transfer affects population risk.

\subsection{Some results on operators}

We shall use the spectral calculus for self-adjoint
operators~\cite{kato1995} and the variational
characterization of eigenvalues in finite
dimension~\cite[Chapter~III]{bhatia1997}.

Let \(A\) be a bounded self-adjoint operator on a
Hilbert space, and let \(\Task\) be a nonzero
closed subspace. Put
\[
    \lambda_{\min}^{\Task}(A)
    =
    \inf_{\substack{v\in\Task\\\|v\|=1}}
    \langle v,Av\rangle,
\]
and
\[
    \lambda_{\max}^{\Task}(A)
    =
    \sup_{\substack{v\in\Task\\\|v\|=1}}
    \langle v,Av\rangle.
\]
If \(\Task\) is finite-dimensional, these are the
smallest and largest eigenvalues of the compression
of \(A\) to \(\Task\).

We shall use the following elementary consequence
of the min--max principle.

\begin{theorem}
\label{quoted:weyl}
Let \(A\) and \(B\) be bounded self-adjoint operators.
One has
\[
    \left|
    \lambda_{\min}^{\Task}(A)
    -
    \lambda_{\min}^{\Task}(B)
    \right|
    \leq \|A-B\|_{\op},
\]
and
\[
    \left|
    \lambda_{\max}^{\Task}(A)
    -
    \lambda_{\max}^{\Task}(B)
    \right|
    \leq \|A-B\|_{\op}.
\]
\end{theorem}

For the study of evolving spectral subspaces,
we use the perturbation results of Davis and
Kahan~\cite{daviskahan1970} and Kato~\cite{kato1995}.

\begin{theorem}[Davis--Kahan and Kato \cite{daviskahan1970,kato1995}]
\label{quoted:dk-kato}
Let \(W\) be self-adjoint and suppose that a part
of its spectrum is separated from the remainder
by a gap \(\Delta>0\). Let \(P\) denote the
corresponding spectral projector.

For a self-adjoint perturbation \(E\) with
\(\|E\|_{\op}<\Delta/2\), let \(\widetilde P\)
be the projector onto the corresponding spectral
subspace of \(W+E\). Then
\[
    \|\widetilde P-P\|_{\op}
    \leq C\frac{\|E\|_{\op}}{\Delta},
\]
where \(C\) is an absolute constant.

Suppose now that \(W=W(T)\) is continuously
differentiable in operator norm. Let \(\Gamma\)
be a positively oriented contour separating a
part of the spectrum from its complement for
\(T\) in some interval. The corresponding Riesz
projector is
\[
    P(T)=\frac{1}{2\pi i}
    \oint_\Gamma(zI-W(T))^{-1}\,\dd z,
\]
and its derivative is
\[
    \dot P(T)=\frac{1}{2\pi i}
    \oint_\Gamma
    (zI-W(T))^{-1}
    \dot W(T)
    (zI-W(T))^{-1}
    \,\dd z.
\]

In finite dimension, suppose that the eigenvalues
\(\lambda_i(T)\) are simple, with corresponding
differentiable orthonormal eigenvectors \(u_i(T)\).
For \(i\ne j\), one has
\[
    \langle u_i,\dot u_j\rangle
    =
    \frac{\langle u_i,\dot W u_j\rangle}
         {\lambda_j-\lambda_i}.
\]
\end{theorem}

The formula for \(\dot P(T)\) follows by
differentiating the resolvent under the contour
integral. Indeed, one has
\[
    \frac{\dd}{\dd T}(zI-W(T))^{-1}
    =
    (zI-W(T))^{-1}\dot W(T)(zI-W(T))^{-1}.
\]
The formula for the derivatives of the eigenvectors
is obtained by differentiating
\[
    W(T)u_j(T)=\lambda_j(T)u_j(T),
\]
and taking the inner product with \(u_i(T)\)
for \(i\ne j\).

We shall also use the factorization theorem of
Douglas~\cite{douglas1966}. For bounded operators
\(A\) and \(B\) with a common codomain, and for
\(c\geq 0\), the inequality
\[
    AA^*\preceq cBB^*,
\]
is equivalent to the existence of a bounded
operator \(D\) such that
\[
    A=BD,
    \qquad
    \|D\|_{\op}\leq\sqrt{c}.
\]
In \cref{cor:factorization}, we give a direct
construction of the factor associated with an
evaluation distribution.

\clearpage

\section{Evaluation laws and output-space dynamics}
\label{sec:setting}

We introduce the mathematical framework used throughout
the paper. We first define the training and evaluation
spaces in a manner that allows different evaluation
laws to be compared on a common space. Starting from
a preconditioned gradient flow, we derive the equations
governing the evolution of training and evaluation
outputs without assuming that the kernels remain
constant. We then introduce the propagator, the
cumulative dissipation operator, and the evaluation-transfer
operator, and establish exact identities relating them
to the decrease of training loss and the displacement
of evaluation outputs. Finally, we examine the case
of constant kernels and distinguish directions with
small cumulative dissipation from those belonging
to the exact reservoir.

\subsection{Evaluation laws}

Let \((\mathcal Z,\mathcal A)\) be a measurable space
endowed with a probability measure \(\lambda\), and let
\[
    \HE=L^2(\lambda;\R^p),
    \qquad
    \ip{v}{w}_{\HE}
    =\int_{\mathcal Z}v(z)^\top w(z)\,\dd\lambda(z).
\]
By an evaluation law we mean a probability measure \(\mu\)
absolutely continuous with respect to \(\lambda\), whose
Radon--Nikodym derivative
\[
    r_\mu=\frac{\dd\mu}{\dd\lambda},
\]
belongs to \(L^\infty(\lambda)\).

The multiplication operator \(C_\mu\) on \(\HE\), defined by
\[
    (C_\mu v)(z)=r_\mu(z)v(z),
\]
is bounded, self-adjoint, and positive semidefinite, and
its operator norm is given by
\[
    \|C_\mu\|_{\op}
    =\|r_\mu\|_{L^\infty(\lambda)}.
\]
For \(v\in\HE\), put
\begin{equation}
\label{eq:common-evaluation-metric}
    \norm v_{C_\mu}^2
    :=\ip{v}{C_\mu v}_{\HE}
    =\int_{\mathcal Z}\norm{v(z)}_2^2\,\dd\mu(z)
    =\norm{C_\mu^{1/2}v}_{\HE}^2.
\end{equation}
This defines a seminorm on \(\HE\), since a set of positive
\(\lambda\)-measure may have zero \(\mu\)-measure, and
coincides with the usual \(L^2(\mu;\R^p)\) norm on the
corresponding equivalence classes.

The advantage of introducing a reference measure is that
the operators associated with different evaluation laws
act on a common space, so that expressions such as
\[
    G^*C_\mu G-G^*C_\nu G,
\]
are well defined. Given two probability measures \(\mu\)
and \(\nu\), one may always take
\[
    \lambda=\frac12(\mu+\nu),
\]
in which case the corresponding densities satisfy
\[
    0\leq r_\mu,r_\nu\leq 2,
\]
almost everywhere with respect to \(\lambda\), whether
or not the two measures have disjoint supports.

\subsection{Parameters, outputs, and loss}

Let \(\HP\) be a real Hilbert space and let
\[
    F:\HP\times\mathcal Z\longrightarrow\R^p,
\]
be the model. Given a training sample
\(S=(z_1,\ldots,z_n)\), we denote by
\[
    \HS=(\R^p)^n,
\]
the space of training outputs, endowed with the inner
product
\[
    \ip{v}{w}_{\HS}
    =\sum_{i=1}^n v_i^\top w_i.
\]

The training and evaluation output maps are given by
\[
    U_S(\theta)
    =\bigl(F(\theta,z_1),\ldots,F(\theta,z_n)\bigr),
    \qquad
    U_E(\theta)(z)=F(\theta,z),
\]
where \(U_E\) is assumed to take values in \(\HE\) on
an open neighborhood of the training trajectory.
Their Fr\'echet derivatives are denoted by
\[
    J_S(\theta):\HP\longrightarrow\HS,
    \qquad
    J_E(\theta):\HP\longrightarrow\HE.
\]

Let \(\Phi_S:\HS\longrightarrow\R\) be a \(C^2\)
function incorporating the training targets, and define
the empirical loss by
\begin{equation}
\label{eq:empirical-composite-loss}
    \Loss_S(\theta)=\Phi_S(U_S(\theta)).
\end{equation}
The normalization of the empirical loss is included
in the definition of \(\Phi_S\), rather than in the
inner product on \(\HS\) or in the kernels introduced
below. For instance, the choice
\[
    \Phi_S(u)=\frac12\|u-y\|^2,
\]
gives
\[
    \nabla\Phi_S(u)=u-y,
    \qquad
    \nabla^2\Phi_S(u)=I,
\]
whereas the choice
\[
    \Phi_S(u)=\frac1{2n}\|u-y\|^2,
\]
gives
\[
    \nabla\Phi_S(u)=\frac{u-y}{n},
    \qquad
    \nabla^2\Phi_S(u)=\frac1n I.
\]
Unless otherwise specified, we use the first convention
in the squared-loss calculations of the following sections.

\begin{assumption}[Regularity]
\label{ass:regularity}
On each interval \([s,T]\) under consideration, we make
the following assumptions.
\begin{enumerate}[label=(\roman*)]
\item The parameter trajectory \(\theta(t)\) is absolutely
continuous and satisfies the preconditioned gradient
flow almost everywhere.

\item The maps \(U_S\) and \(U_E\) are continuously
Fr\'echet differentiable on an open neighborhood of
the trajectory, and \(\Phi_S\) is \(C^2\).

\item The operators \(J_S(t)\), \(J_E(t)\), and
\[
    B(t)=\nabla^2\Phi_S(U_S(\theta(t))),
\]
are strongly measurable and uniformly bounded.

\item The preconditioner \(M(t)\in\mathcal L(\HP)\)
is strongly measurable, uniformly bounded,
self-adjoint, and positive semidefinite.

\item All operator-valued integrals are Bochner
integrals.
\end{enumerate}
\end{assumption}

These assumptions suffice for the differentiations and
integrations used below, with the understanding that
identities involving time derivatives hold almost everywhere
unless otherwise stated. No convexity assumption is made
on the empirical loss in parameter space, nor do we impose
any restriction on the width of the model or on the size
of the parameter displacement. The flow considered here
does not include additional momentum variables or explicit
regularization drift, which would require further terms
in the equations.

\subsection{The parameter and output equations}

Along the training trajectory, put
\[
    u(t)=U_S(\theta(t)),
    \qquad
    g(t)=\nabla\Phi_S(u(t)),
    \qquad
    B(t)=\nabla^2\Phi_S(u(t)).
\]
For a parameter displacement \(v\in\HP\), the chain
rule gives
\[
    D\Loss_S(\theta)[v]
    =\ip{g}{J_Sv}_{\HS}
    =\ip{J_S^*g}{v}_{\HP},
\]
from which it follows that
\[
    \nabla_\theta\Loss_S=J_S^*g,
\]
and hence that the preconditioned gradient flow is
given by
\begin{equation}
\label{eq:parameter-flow}
    \dot\theta(t)=-M(t)J_S(t)^*g(t).
\end{equation}

We associate with this flow two operators, the training
kernel and the cross-evaluation kernel, defined by
\begin{align}
    K_{SS}^M(t)
    &=J_S(t)M(t)J_S(t)^*,
    \label{eq:train-kernel}\\
    K_{ES}^M(t)
    &=J_E(t)M(t)J_S(t)^*,
    \label{eq:cross-kernel}
\end{align}
whose domains and codomains are given by
\[
    K_{SS}^M(t):\HS\longrightarrow\HS,
    \qquad
    K_{ES}^M(t):\HS\longrightarrow\HE.
\]

The operator \(K_{SS}^M(t)\) is self-adjoint and positive
semidefinite, since for every \(h\in\HS\) one has
\begin{equation}
\label{eq:kernel-positive-square}
    \ip{h}{K_{SS}^M(t)h}
    =\norm{M(t)^{1/2}J_S(t)^*h}_{\HP}^2
    \geq 0.
\end{equation}
There is no corresponding positivity property for
\(K_{ES}^M(t)\), which is an operator between two
different Hilbert spaces.

Applying the chain rule to \eqref{eq:parameter-flow},
we obtain
\begin{equation}
\label{eq:output-flow}
    \dot u(t)=-K_{SS}^M(t)g(t),
    \qquad
    \dot g(t)=-B(t)K_{SS}^M(t)g(t),
\end{equation}
and
\begin{equation}
\label{eq:evaluation-velocity}
    \frac{\dd}{\dd t}U_E(\theta(t))
    =-K_{ES}^M(t)g(t).
\end{equation}

The second identity in \eqref{eq:output-flow} follows
from the relation \(\dot g=B\dot u\) and contains no
derivative of \(J_S\), since \(g\) denotes the gradient
with respect to the training outputs rather than the
gradient with respect to the parameters. Note also
that the order of the factors \(B\) and \(K_{SS}^M\)
cannot in general be reversed.

\begin{proposition}[Dissipation of training loss]
\label{prop:instantaneous-training-dissipation}
Under \cref{ass:regularity}, one has
\begin{equation}
\label{eq:instantaneous-training-dissipation}
    \frac{\dd}{\dd t}\Phi_S(u(t))
    =-\ip{g(t)}{K_{SS}^M(t)g(t)}
    =-\norm{M(t)^{1/2}J_S(t)^*g(t)}_{\HP}^2
    \leq 0.
\end{equation}
\end{proposition}

\begin{proof}
It follows from \eqref{eq:output-flow} that
\[
    \frac{\dd}{\dd t}\Phi_S(u(t))
    =\ip{g(t)}{\dot u(t)}
    =-\ip{g(t)}{K_{SS}^M(t)g(t)},
\]
and the assertion is a consequence of
\eqref{eq:kernel-positive-square}.
\end{proof}

Thus the empirical loss is a nonincreasing function
of time, whereas no analogous conclusion can be drawn
for the population risk without further assumptions
on the evaluation outputs and the target. We shall
return to this question in \cref{sec:risk}.

\subsection{The propagator}

Let \(\Pg(t,s)\) be the fundamental solution associated
with the equation for the output gradient in
\eqref{eq:output-flow}, characterized by
\begin{equation}
\label{eq:propagator}
    \partial_t\Pg(t,s)
    =-B(t)K_{SS}^M(t)\Pg(t,s),
    \qquad
    \Pg(s,s)=\Id,
\end{equation}
or, equivalently, by the integral equation
\begin{equation}
\label{eq:propagator-integral}
    \Pg(t,s)=\Id-
    \int_s^t
    B(\tau)K_{SS}^M(\tau)\Pg(\tau,s)\,\dd\tau.
\end{equation}

Since \(\HS\) is finite-dimensional and the coefficients
are bounded and measurable, this equation admits a unique
absolutely continuous solution satisfying, by Gronwall's
inequality,
\[
    \norm{\Pg(t,s)}_{\op}
    \leq
    \exp\left(
    \int_s^t
    \norm{B(\tau)K_{SS}^M(\tau)}_{\op}
    \,\dd\tau
    \right).
\]

\begin{proposition}
\label{prop:exact-gradient-propagation}
For \(s\leq t\leq T\), one has
\begin{equation}
\label{eq:exact-gradient-propagation}
    g(t)=\Pg(t,s)g(s).
\end{equation}
Moreover, the propagators satisfy the composition law
\[
    \Pg(t,r)\Pg(r,s)=\Pg(t,s),
\]
for \(s\leq r\leq t\).
\end{proposition}

\begin{proof}
The two sides of \eqref{eq:exact-gradient-propagation}
satisfy the same differential equation and have the
same initial value at \(s\), so that the first assertion
follows from uniqueness. Applying the same argument on
the interval \([r,t]\), with initial value \(\Pg(r,s)\),
gives the composition law.
\end{proof}

\begin{remark}
\label{rem:pathwise-propagator}
The representation \eqref{eq:exact-gradient-propagation}
is exact along the given training trajectory, even when
the coefficients \(B(t)\) and \(K_{SS}^M(t)\) depend
nonlinearly on the parameters. In general, the operators
\(B(t)K_{SS}^M(t)\) at different times do not commute,
so that the propagator cannot be written as
\[
    \Pg(t,s)
    =
    \exp\left(
    -\int_s^t B(\tau)K_{SS}^M(\tau)\,\dd\tau
    \right).
\]

For an arbitrary \(h\in\HS\), the vector \(\Pg(t,s)h\)
is obtained by solving the linear equation with the
coefficients of the given trajectory and initial value
\(h\). These coefficients need not coincide with those
of a second training trajectory whose initial output
gradient is \(h\). Thus the propagator describes the
evolution of output-gradient directions along a fixed
trajectory and is not, in general, the derivative of
the parameter flow with respect to its initial condition.
\end{remark}

\subsection{The dissipation and transfer operators}

\begin{definition}
\label{def:WG}
For \(s\leq T\), we define
\begin{align}
    \W(s,T)
    &=
    \int_s^T
    \Pg(\tau,s)^*K_{SS}^M(\tau)
    \Pg(\tau,s)\,\dd\tau,
    \label{eq:W}\\
    \G(T,s)
    &=
    \int_s^T
    K_{ES}^M(\tau)\Pg(\tau,s)\,\dd\tau,
    \label{eq:G}
\end{align}
and call \(\W(s,T)\) the cumulative dissipation operator
and \(\G(T,s)\) the evaluation-transfer operator,
respectively.
\end{definition}

The first operator acts on \(\HS\), whereas the second
takes values in \(\HE\). When the evaluation law is
\(\mu\), the corresponding weighted transfer operator
on the common reference space is
\[
    C_\mu^{1/2}\G(T,s).
\]

The propagators appearing in \eqref{eq:W} and
\eqref{eq:G} express the contributions at different
times in terms of a common initial direction at time
\(s\). In particular, for every \(h\in\HS\), one has
\begin{equation}
\label{eq:W-integrated-energy}
    \ip{h}{\W(s,T)h}
    =
    \int_s^T
    \norm{M(\tau)^{1/2}J_S(\tau)^*
    \Pg(\tau,s)h}_{\HP}^2
    \,\dd\tau,
\end{equation}
which shows that \(\W(s,T)\) is self-adjoint and
positive semidefinite. Its quadratic form measures
the cumulative dissipative energy associated with the
propagated direction \(h\), rather than the squared
norm of the parameter displacement, which involves
a different integration in time.

\begin{proposition}[Finite-window identities]
\label{prop:window-identities}
Let \(g_s=g(s)\). Under \cref{ass:regularity}, one has
\begin{align}
    \Phi_S(u(s))-\Phi_S(u(T))
    &=\ip{g_s}{\W(s,T)g_s},
    \label{eq:finite-window-training-dissipation}\\
    U_E(\theta(T))-U_E(\theta(s))
    &=-\G(T,s)g_s.
    \label{eq:window-output-displacement}
\end{align}
In particular, for every evaluation law \(\mu\),
\begin{equation}
\label{eq:actual-evaluation-visible-energy}
    \norm{U_E(\theta(T))-U_E(\theta(s))}_{C_\mu}^2
    =
    \ip{g_s}{\G(T,s)^*C_\mu\G(T,s)g_s}.
\end{equation}
\end{proposition}

\begin{proof}
Integrating \eqref{eq:instantaneous-training-dissipation}
and substituting \(g(\tau)=\Pg(\tau,s)g_s\), we obtain
\[
\begin{aligned}
    \Phi_S(u(s))-\Phi_S(u(T))
    &=
    \int_s^T
    \ip{g(\tau)}{K_{SS}^M(\tau)g(\tau)}
    \,\dd\tau\\
    &=
    \ip{g_s}{
    \left(
    \int_s^T
    \Pg(\tau,s)^*K_{SS}^M(\tau)
    \Pg(\tau,s)\,\dd\tau
    \right)g_s}\\
    &=\ip{g_s}{\W(s,T)g_s},
\end{aligned}
\]
which proves \eqref{eq:finite-window-training-dissipation}.
Similarly, integration of \eqref{eq:evaluation-velocity}
gives
\[
\begin{aligned}
    U_E(\theta(T))-U_E(\theta(s))
    &=
    -\int_s^T K_{ES}^M(\tau)g(\tau)\,\dd\tau\\
    &=
    -\int_s^T K_{ES}^M(\tau)
    \Pg(\tau,s)g_s\,\dd\tau\\
    &=-\G(T,s)g_s.
\end{aligned}
\]
The identity \eqref{eq:actual-evaluation-visible-energy}
then follows by taking the squared \(C_\mu\)-seminorm
of both sides.
\end{proof}

\begin{remark}
\label{rem:sign-and-evaluation-law}
The displacement of the evaluation outputs is given by
\(-\G(T,s)g_s\), rather than by \(\G(T,s)g_s\), a sign
which will be important in the population-risk identity
of \cref{thm:risk-balance}.

Although the displacement itself is independent of
the evaluation law, its size depends on the choice of
\(C_\mu\). Thus the operator
\[
    \G(T,s)^*C_\mu\G(T,s),
\]
determines the squared size of the displacement under
\(\mu\), but contains no information about its direction
relative to the target. It cannot therefore determine,
by itself, whether the displacement decreases population
risk.
\end{remark}

\subsection{The case of constant kernels}

Suppose that the loss is given by
\[
    \Phi_S(u)=\frac12\|u-y\|^2,
\]
and that the kernels
\[
    K_{SS}^M(t)=K,
    \qquad
    K_{ES}^M(t)=K_{ES},
\]
are constant on \([s,T]\). Writing \(\Delta=T-s\) and
noting that \(B=I\), we obtain
\begin{equation}
\label{eq:frozen-propagator-W}
    \Pg(t,s)=e^{-(t-s)K},
    \qquad
    \W(s,T)=\frac12\bigl(I-e^{-2\Delta K}\bigr),
\end{equation}
the second equality following from
\[
    \W(s,T)
    =\int_0^\Delta Ke^{-2tK}\,\dd t
    =\frac12\bigl(I-e^{-2\Delta K}\bigr).
\]
Similarly, the transfer operator is given by
\begin{equation}
\label{eq:frozen-transfer-check}
    \G(T,s)=K_{ES}\psi_\Delta(K),
    \qquad
    \psi_\Delta(x)=
    \begin{cases}
        (1-e^{-\Delta x})/x,&x>0,\\
        \Delta,&x=0.
    \end{cases}
\end{equation}
Both formulas follow from the spectral calculus and
remain valid when \(K\) is singular.

If \(h\) is an eigenvector of \(K\) corresponding to
the eigenvalue \(\kappa\), then
\[
    \Pg(t,s)h=e^{-(t-s)\kappa}h,
    \qquad
    \W(s,T)h
    =\frac12(1-e^{-2\Delta\kappa})h.
\]
Since
\[
    \frac12(1-e^{-2\Delta\kappa})
    \sim\Delta\kappa,
    \qquad
    \Delta\kappa\longrightarrow 0,
\]
a direction corresponding to a small positive eigenvalue
of \(K\) may have small cumulative dissipation on a
finite interval without belonging to the reservoir.
The distinction between small and zero eigenvalues will
be of importance in the following sections.

\begin{remark}[Exact and effective reservoirs]
\label{rem:exact-effective-reservoir}
The exact reservoir is \(\Ker\W(s,T)\), whereas by an
effective reservoir we mean the spectral subspace of
\(\W(s,T)\) associated with eigenvalues below a prescribed
positive threshold.

To see that these two notions need not coincide, suppose
that
\[
    K_{SS}^M(t)\succeq k_0I,
    \qquad k_0>0,
\]
on a subinterval of positive length. Since the fundamental
solution \(\Pg(t,s)\) is invertible,
\eqref{eq:W-integrated-energy} implies that
\(\W(s,T)\) is positive definite.

In this case the exact reservoir is zero, although
\(\W(s,T)\) may have small positive eigenvalues.
The corresponding spectral directions are those for
which the resolvent filter introduced in
\cref{sec:resolvent} is particularly useful.
\end{remark}

\subsection{Concluding remarks}

The identities \eqref{eq:finite-window-training-dissipation}
and \eqref{eq:window-output-displacement} give two
descriptions of the same training trajectory, the former
in terms of the dissipation of empirical loss and the
latter in terms of the displacement of evaluation outputs.
In \cref{sec:reservoir}, we shall establish a quantitative
relation between the operators \(\W\) and \(\G\), while
in \cref{sec:risk} we shall examine the conditions under
which the resulting displacement decreases population risk.

The preceding identities are conditional on the given
trajectory and do not determine its future kernels,
identify a task subspace from unlabeled data, or yield
a grokking time from the empirical loss alone. The study
of these questions requires additional information on
the evolution of the operators, which will be introduced
as needed in the following sections.

\clearpage

\section{Reservoir invisibility and quantitative transfer}
\label{sec:reservoir}

We study the relation between cumulative training
dissipation and the transfer of training directions
to evaluation outputs. We first extend the reservoir
invisibility theorem of Litman and Guo to the setting
of preconditioned gradient flows and general evaluation
laws. We then establish a quantitative bound on the
magnitude of the transferred signal in terms of its
cumulative dissipation. This bound leads to a
factorization that distinguishes two mechanisms of
limited visibility, namely insufficient mobility
along the training trajectory and weak transfer
to the evaluation space. The distinction between
these mechanisms will be used in the subsequent
study of delayed and selective generalization.

\begin{theorem}[Kernel of an integrated square]
\label{lem:integrated-square}
Let \(\mathcal X\) and \(\mathcal Y\) be Hilbert spaces,
and let \(A(t):\mathcal X\to\mathcal Y\) be strongly
measurable and square integrable in operator norm.
If
\[
    W=\int_s^T A(t)^*A(t)\,\dd t,
\]
then \(W\) is self-adjoint and positive semidefinite,
and one has
\[
    h\in\Ker W
    \quad\Longleftrightarrow\quad
    A(t)h=0
    \quad\text{for almost every }t\in[s,T].
\]
\end{theorem}

\begin{proof}
Since each operator \(A(t)^*A(t)\) is self-adjoint
and positive semidefinite, the same is true of \(W\).
For every \(h\in\mathcal X\), one has
\[
    \ip{h}{Wh}_{\mathcal X}
    =
    \int_s^T\norm{A(t)h}_{\mathcal Y}^2\,\dd t.
\]
Since \(W\) is positive semidefinite, the condition
\(Wh=0\) is equivalent to
\(\ip{h}{Wh}_{\mathcal X}=0\). By the preceding
identity, this is equivalent to \(A(t)h=0\) for
almost every \(t\in[s,T]\), as asserted.
\end{proof}

\begin{theorem}[Reservoir invisibility]
\label{thm:reservoir}
Under \cref{ass:regularity}, one has
\[
    \Ker\W(s,T)\subseteq\Ker\G(T,s).
\]
Let
\[
    P_0=\1_{\{0\}}(\W(s,T)),
    \qquad
    P_+=\1_{(0,\infty)}(\W(s,T)),
\]
be the spectral projectors associated with the zero
and positive parts of the spectrum of \(\W(s,T)\),
respectively. Then
\[
    \G(T,s)P_0=0,
    \qquad
    \G(T,s)=\G(T,s)P_+.
\]
The same identities hold for the weighted transfer
operator \(C_\mu^{1/2}\G(T,s)\) associated with any
evaluation law \(\mu\) satisfying the assumptions of
\cref{sec:setting}.
\end{theorem}

\begin{proof}
Put
\[
    A(\tau)
    =M(\tau)^{1/2}J_S(\tau)^*\Pg(\tau,s),
\]
so that
\[
    \W(s,T)=\int_s^T A(\tau)^*A(\tau)\,\dd\tau.
\]
If \(h\in\Ker\W(s,T)\), it follows from
\cref{lem:integrated-square} that \(A(\tau)h=0\)
for almost every \(\tau\in[s,T]\). Hence
\[
    K_{ES}^M(\tau)\Pg(\tau,s)h
    =
    J_E(\tau)M(\tau)^{1/2}A(\tau)h
    =0,
\]
almost everywhere, and integration gives
\(\G(T,s)h=0\).

The assertions concerning the spectral projectors
follow from the orthogonal decomposition
\[
    \HS=\Ker\W\oplus\overline{\ran\W},
\]
while those concerning the weighted transfer operators
follow by composition with \(C_\mu^{1/2}\).
\end{proof}

The preceding theorem concerns the directions on which
\(\W\) vanishes. We now examine the extent to which a
direction associated with a small positive eigenvalue
of \(\W\) can affect the evaluation outputs.

\begin{theorem}[Dissipation domination]
\label{thm:domination}
Let \(\mu\) be an evaluation law and put
\begin{equation}
\label{eq:beta}
    \beta_\mu(s,T)
    =
    \int_s^T
    \norm{C_\mu^{1/2}J_E(\tau)
    M(\tau)^{1/2}}_{\op}^2\,\dd\tau.
\end{equation}
Then
\begin{equation}
\label{eq:domination}
    \G(T,s)^*C_\mu\G(T,s)
    \preceq
    \beta_\mu(s,T)\,\W(s,T),
\end{equation}
or, equivalently, for every \(h\in\HS\),
\[
    \norm{C_\mu^{1/2}\G(T,s)h}_{\HE}^2
    \leq
    \beta_\mu(s,T)\ip{h}{\W(s,T)h}_{\HS}.
\]
\end{theorem}

\begin{proof}
With the notation of the preceding proof, one has
\[
    C_\mu^{1/2}\G h
    =
    \int_s^T
    C_\mu^{1/2}J_E(\tau)M(\tau)^{1/2}
    A(\tau)h\,\dd\tau.
\]
By the triangle inequality and the Cauchy--Schwarz
inequality,
\begin{align*}
    \norm{C_\mu^{1/2}\G h}_{\HE}
    &\leq
    \int_s^T
    \norm{C_\mu^{1/2}J_E(\tau)
    M(\tau)^{1/2}}_{\op}
    \norm{A(\tau)h}_{\HP}\,\dd\tau\\
    &\leq
    \beta_\mu(s,T)^{1/2}
    \left(
    \int_s^T
    \norm{A(\tau)h}_{\HP}^2\,\dd\tau
    \right)^{1/2}.
\end{align*}
Since
\[
    \int_s^T\norm{A(\tau)h}_{\HP}^2\,\dd\tau
    =
    \ip{h}{\W h}_{\HS},
\]
the desired inequality follows by squaring both sides.
\end{proof}

The preceding inequality allows us to factor the
weighted transfer operator through the positive
square root of \(\W\). This is an instance of the
factorization theorem of Douglas~\cite{douglas1966},
for which we give a direct construction.

\begin{corollary}[Mobility and transfer efficiency]
\label{cor:factorization}
There exists a unique bounded operator
\[
    Q_\mu(s,T):
    \overline{\ran\W(s,T)^{1/2}}\longrightarrow\HE,
\]
such that
\begin{equation}
\label{eq:douglas-factor}
    C_\mu^{1/2}\G(T,s)
    =
    Q_\mu(s,T)\W(s,T)^{1/2},
    \qquad
    \norm{Q_\mu(s,T)}_{\op}^2
    \leq\beta_\mu(s,T).
\end{equation}

For a nonzero vector \(h\in\HS\), we define its
mobility by
\begin{equation}
\label{eq:mobility}
    \Mob(h)
    =
    \frac{\ip{h}{\W h}}{\norm h^2},
\end{equation}
and, provided that \(h\notin\Ker\W\), its transfer
efficiency under \(\mu\) by
\begin{equation}
\label{eq:efficiency}
    \Eff_\mu(h)
    =
    \frac{\norm{C_\mu^{1/2}\G h}^2}
         {\ip{h}{\W h}}.
\end{equation}
For every \(h\notin\Ker\W\), one has
\begin{equation}
\label{eq:two-gate}
    \frac{\norm{C_\mu^{1/2}\G h}^2}{\norm h^2}
    =
    \Mob(h)\Eff_\mu(h),
    \qquad
    0\leq\Eff_\mu(h)\leq\beta_\mu.
\end{equation}
\end{corollary}

\begin{proof}
We define \(Q_\mu\) on \(\ran\W^{1/2}\) by
\[
    Q_\mu(\W^{1/2}h)=C_\mu^{1/2}\G h.
\]
This definition is independent of the choice of \(h\),
since \(\Ker\W^{1/2}=\Ker\W\) and, by
\cref{thm:reservoir}, \(\Ker\W\subseteq\Ker\G\).

It follows from \cref{thm:domination} that
\[
    \norm{Q_\mu(\W^{1/2}h)}_{\HE}
    \leq
    \sqrt{\beta_\mu}\norm{\W^{1/2}h}_{\HS},
\]
so that \(Q_\mu\) extends uniquely to a bounded
operator on \(\overline{\ran\W^{1/2}}\), with norm
at most \(\sqrt{\beta_\mu}\).

The equality in \eqref{eq:two-gate} follows from
the definitions \eqref{eq:mobility} and
\eqref{eq:efficiency}, and the inequality follows
from \eqref{eq:domination}.
\end{proof}

\begin{remark}
\label{rem:two-gates}
The quantities \(\Mob(h)\) and \(\Eff_\mu(h)\)
describe two different properties of the direction
\(h\). The former depends only on the cumulative
dissipation along the training trajectory, whereas
the latter also depends on the evaluation law.

A direction may have small visibility because its
mobility is small, or because its transfer efficiency
is small despite substantial mobility. In particular,
two directions having the same mobility need not
have the same visibility under a given evaluation
law, and a direction having large mobility may be
visible under one law and nearly invisible under
another.

The distinction between these two quantities will
be used in our study of delayed and selective
generalization.
\end{remark}

\clearpage

\section{A stable signal filter}
\label{sec:resolvent}

We introduce a continuous approximation of the spectral
projector onto the signal channel in order to avoid
the discontinuities associated with spectral thresholding.
We establish the convergence, Lipschitz continuity,
and differentiability properties of the resulting
resolvent filter. Using the dissipation bound obtained
in the preceding section, we derive a quantitative
estimate for the difference between filtered and
unfiltered evaluation outputs, without requiring
a spectral gap at zero. This construction provides
a stable framework for studying the visibility of
training directions and its evolution over time.

To motivate the construction, consider the projector
\(P_+=\1_{(0,\infty)}(\W)\) onto the signal channel.
Its dependence on \(\W\) is discontinuous at operators
with nontrivial kernel. The same difficulty arises
for a spectral projector defined by a positive
threshold \(\alpha\), whenever an eigenvalue crosses
that threshold. Moreover, the inclusion of additional
spectral directions need not increase the energy
transferred to the evaluation space, since their
images under \(\G\) may partially cancel. We shall
therefore replace these projectors by a continuous
function of \(\W\), while distinguishing the evolution
of the filter from that of the resulting evaluation
visibility.

\begin{definition}[Resolvent signal filter]
\label{def:resolvent}
Let \(W\) be a bounded positive semidefinite operator.
For \(\alpha>0\), put
\begin{equation}
\label{eq:signal-filter}
    \Salpha(W)
    =
    W(W+\alpha\Id)^{-1}
    =
    \Id-\alpha(W+\alpha\Id)^{-1}.
\end{equation}
We call \(\Salpha(W)\) the resolvent signal filter.
The corresponding visibility operator associated with
an evaluation law \(\mu\) is
\begin{equation}
\label{eq:soft-H}
    H_{\mu,\alpha}
    =
    \Salpha(\W)\G^*C_\mu\G\Salpha(\W).
\end{equation}
\end{definition}

We recall some elementary properties of the resolvent
which will be used in the study of the time dependence
of these operators.

\begin{proposition}
\label{prop:resolvent-calculus}
Let \(W\) and \(V\) be bounded self-adjoint positive
semidefinite operators on a Hilbert space, and let
\(\alpha>0\).

\begin{enumerate}[label=(\roman*)]

\item One has
\[
    0\preceq\Salpha(W)\preceq P_+(W)\preceq\Id,
\]
where \(P_+(W)=\1_{(0,\infty)}(W)\). Moreover,
\(\Salpha(W)\) converges strongly to \(P_+(W)\)
as \(\alpha\) tends to zero.

\item The dependence of \(\Salpha(W)\) on \(W\)
is Lipschitz continuous in operator norm, with
\[
    \norm{\Salpha(W)-\Salpha(V)}_{\op}
    \leq
    \frac{1}{\alpha}\norm{W-V}_{\op}.
\]

\item Suppose that \(W=W(T)\) is differentiable in
operator norm, and put
\[
    Q(T)=\dot W(T),
    \qquad
    \Ralpha(T)=(W(T)+\alpha\Id)^{-1}.
\]
Then
\begin{equation}
\label{eq:filter-derivative}
    \frac{\dd}{\dd T}\Salpha(W(T))
    =
    \alpha\Ralpha(T)Q(T)\Ralpha(T).
\end{equation}
In particular, if \(\dot W(T)\succeq0\), then
\[
    \frac{\dd}{\dd T}\Salpha(W(T))\succeq0.
\]

\end{enumerate}
\end{proposition}

\begin{proof}
For \(x\geq0\), the function
\[
    f_\alpha(x)=\frac{x}{x+\alpha},
\]
satisfies
\[
    0\leq f_\alpha(x)\leq\1_{(0,\infty)}(x),
\]
and converges pointwise to \(\1_{(0,\infty)}(x)\)
as \(\alpha\) tends to zero. Assertion (i) follows
from the spectral theorem and dominated convergence.

By the resolvent identity, one has
\[
    \Salpha(W)-\Salpha(V)
    =
    \alpha(W+\alpha\Id)^{-1}
    (W-V)(V+\alpha\Id)^{-1}.
\]
Since \(W\) and \(V\) are positive semidefinite,
the norms of the two resolvents are at most
\(1/\alpha\), which proves (ii).

Finally, differentiating the identity
\[
    \Salpha(W(T))
    =
    \Id-\alpha\Ralpha(T),
\]
and using
\[
    \dot\Ralpha(T)
    =
    -\Ralpha(T)\dot W(T)\Ralpha(T),
\]
gives \eqref{eq:filter-derivative}. If \(Q(T)\) is
positive semidefinite, the same is true of
\(\Ralpha(T)Q(T)\Ralpha(T)\), which proves the
last assertion.
\end{proof}

Although the convergence of \(\Salpha(\W)\) to
\(P_+\) in operator norm generally depends on the
positive spectrum of \(\W\), the corresponding
evaluation outputs satisfy a bound which does not
require a spectral gap at zero.

\begin{theorem}[Approximation of evaluation outputs]
\label{thm:resolvent}
Let \(P_+=\1_{(0,\infty)}(\W)\), and let
\(\beta_\mu\) be the constant defined in \eqref{eq:beta}.
For every \(\alpha>0\) and every \(h\in\HS\), one has
\begin{equation}
\label{eq:soft-error}
    \norm{
    C_\mu^{1/2}\G
    \bigl(P_+-\Salpha(\W)\bigr)h
    }_{\HE}^2
    \leq
    \frac{\beta_\mu\alpha}{4}
    \norm h_{\HS}^2.
\end{equation}
Since \(\G=\G P_+\) by \cref{thm:reservoir},
the left-hand side of \eqref{eq:soft-error} is also
equal to
\[
    \norm{
    C_\mu^{1/2}
    \bigl(\G h-\G\Salpha(\W)h\bigr)
    }_{\HE}^2.
\]
The factor \(1/4\) is optimal in the scalar estimate
used in the proof.
\end{theorem}

\begin{proof}
Applying \cref{thm:domination} to the vector
\((P_+-\Salpha(\W))h\), we obtain
\begin{align*}
    &\norm{
    C_\mu^{1/2}\G
    (P_+-\Salpha(\W))h
    }_{\HE}^2\\
    &\qquad\leq
    \beta_\mu
    \ip{(P_+-\Salpha(\W))h}
       {\W(P_+-\Salpha(\W))h}_{\HS}.
\end{align*}
Since \(P_+\) and \(\Salpha(\W)\) are functions
of \(\W\), they commute with it. The quadratic
form on the right is therefore associated with
the spectral function
\[
    x\longmapsto
    x\left(\frac{\alpha}{x+\alpha}\right)^2,
    \qquad x\geq0.
\]
For every \(x\geq0\), one has
\[
    x\left(\frac{\alpha}{x+\alpha}\right)^2
    \leq\frac{\alpha}{4},
\]
as follows from the inequality
\[
    (x+\alpha)^2\geq4\alpha x,
\]
with equality if and only if \(x=\alpha\).
The spectral calculus now gives
\[
    \ip{(P_+-\Salpha(\W))h}
       {\W(P_+-\Salpha(\W))h}_{\HS}
    \leq
    \frac{\alpha}{4}\norm h_{\HS}^2,
\]
which proves \eqref{eq:soft-error}.
\end{proof}

\begin{remark}
The preceding theorem shows that the convergence
\[
    C_\mu^{1/2}\G\Salpha(\W)
    \longrightarrow
    C_\mu^{1/2}\G,
\]
takes place in operator norm as \(\alpha\) tends
to zero, with
\[
    \norm{
    C_\mu^{1/2}\G
    -
    C_\mu^{1/2}\G\Salpha(\W)
    }_{\op}
    \leq
    \frac12\sqrt{\beta_\mu\alpha}.
\]
This bound is independent of the smallest positive
eigenvalue of \(\W\). Thus, although the signal
projector may be sensitive to perturbations of
\(\W\) near zero, its effect on evaluation outputs
admits a stable approximation without any assumption
on the separation of the positive spectrum from zero.
\end{remark}

\clearpage

\section{Task-subspace visibility}
\label{sec:task}

We extend the study of visibility from individual
training directions to subspaces associated with
a given task. By decomposing the visibility operator
relative to the task subspace and its orthogonal
complement, we obtain two-sided bounds that account
for task alignment, visibility within each subspace,
and the coupling between them. These bounds give
sufficient conditions for an increase in visibility
and determine the alignment required to reach a
prescribed visibility threshold. We also establish
a lower bound on the dimension of the subspace of
visible task directions in terms of average task
visibility. Finally, we study the proximity of the
task subspace to a spectral subspace of the visibility
operator and establish the stability of task visibility
under resolvent filtering.

\begin{definition}
\label{def:task-visibility}
Let \(\mathsf F\) be one of the operators
\[
    P_{>\alpha}=\1_{(\alpha,\infty)}(W),
    \qquad
    P_+=\1_{(0,\infty)}(W),
    \qquad
    S_\alpha(W),
\]
and let \(\mu\) be an evaluation law. We associate with
\(\mathsf F\) the positive semidefinite operator
\[
    H_{\mu,\mathsf F}
    =
    \mathsf F G^*C_\mu G\mathsf F,
\]
and, for every nonzero vector \(h\in\HS\), put
\[
    \Geo_{\mu,\mathsf F}(h)
    =
    \frac{\langle h,H_{\mu,\mathsf F}h\rangle}{\|h\|^2}
    =
    \frac{\|C_\mu^{1/2}G\mathsf Fh\|^2}{\|h\|^2}.
\]

By a \emph{task subspace} we mean a nonzero closed
subspace \(\Task\) of \(\HS\), chosen according to the
target rule and independently of the evaluation law.
We denote by \(\Pi=\Pi_\Task\) the orthogonal projector
onto \(\Task\), and put \(\Pi^\perp=I-\Pi\).

For a positive semidefinite operator \(H\), we use the
notation introduced in \cref{sec:background},
\[
    \lambda_{\min}^{\Task}(H)
    =
    \inf_{\substack{v\in\Task\\\|v\|=1}}
    \langle v,Hv\rangle,
\]
\[
    \lambda_{\max}^{\Task}(H)
    =
    \sup_{\substack{v\in\Task\\\|v\|=1}}
    \langle v,Hv\rangle.
\]
The alignment of a nonzero vector \(h\) with \(\Task\)
is defined by
\[
    \rho_\Task(h)
    =
    \frac{\|\Pi h\|^2}{\|h\|^2}.
\]
If \(\Task\) is finite-dimensional, we also define its
average visibility by
\[
    \bar\lambda_\Task(H)
    =
    \frac{\tr(\Pi H\Pi)}{\rank\Pi}.
\]
\end{definition}

For an algebraic learning problem, the subspace
\(\Task\) may be generated by the Fourier characters
associated with the target rule. In a teacher--student
model, it may be taken to be the span of output gradients
generated by the teacher, while in a statistical setting
it may be constructed from features common to independent
samples.

In contrast with the spectral subspaces of \(W\), which
are determined by the training dynamics, the subspace
\(\Task\) is specified by the task under consideration.
The restriction of the quadratic form of
\(H_{\mu,\mathsf F}\) to \(\Task\) then describes the
visibility of that task under the law \(\mu\).

\subsection{The block decomposition}

\begin{theorem}[Block visibility bounds]
\label{thm:block-task}
Let \(H=H_{\mu,\mathsf F}\), and suppose that
\(\Task\) is a proper nonzero subspace of \(\HS\).
Relative to the orthogonal decomposition
\[
    \HS=\Task\oplus\Task^\perp,
\]
write
\[
    H=
    \begin{pmatrix}
        A&B\\
        B^*&D
    \end{pmatrix},
\]
where
\[
    A=\Pi H\Pi|_\Task,
    \qquad
    D=\Pi^\perp H\Pi^\perp|_{\Task^\perp},
    \qquad
    B=\Pi H|_{\Task^\perp}.
\]
Put
\[
    b=\lambda_{\min}(A),
    \qquad
    a=\lambda_{\max}(A),
    \qquad
    d=\lambda_{\min}(D),
    \qquad
    c=\lambda_{\max}(D),
\]
and let \(\kappa=\|B\|_{\op}\).

For every nonzero vector \(h\in\HS\), with
\(\rho=\rho_\Task(h)\), one has
\begin{align}
    \rho b+(1-\rho)d
    -2\kappa\sqrt{\rho(1-\rho)}
    &\leq
    \Geo_{\mu,\mathsf F}(h),
    \label{eq:block-lower}\\
    \Geo_{\mu,\mathsf F}(h)
    &\leq
    \rho a+(1-\rho)c
    +2\kappa\sqrt{\rho(1-\rho)}.
    \label{eq:block-upper}
\end{align}
Equality may occur when the extremal directions of
the diagonal blocks are compatible with the extremal
directions of \(B\).
\end{theorem}

\begin{proof}
Suppose first that \(0<\rho<1\), and write
\[
    \frac{h}{\|h\|}
    =
    \sqrt{\rho}\,v+\sqrt{1-\rho}\,w,
\]
where \(v\in\Task\) and \(w\in\Task^\perp\) are unit
vectors. The block decomposition of \(H\) gives
\[
\begin{aligned}
    \Geo_{\mu,\mathsf F}(h)
    &=
    \rho\langle v,Av\rangle
    +(1-\rho)\langle w,Dw\rangle\\
    &\quad+
    2\sqrt{\rho(1-\rho)}
    \,\Re\langle v,Bw\rangle.
\end{aligned}
\]
Since
\[
    b\leq\langle v,Av\rangle\leq a,
    \qquad
    d\leq\langle w,Dw\rangle\leq c,
\]
and
\[
    |\langle v,Bw\rangle|
    \leq\kappa,
\]
the two inequalities follow immediately. The cases
\(\rho=0\) and \(\rho=1\) follow from the definitions
of the extremal eigenvalues of \(D\) and \(A\),
respectively.

Equality in the lower bound is attained if the
compression of \(H\) to the plane generated by \(v\)
and \(w\) is given by
\[
    \begin{pmatrix}
        b&-\kappa\\
        -\kappa&d
    \end{pmatrix},
\]
provided that such a compression is compatible with
the positivity of \(H\). The corresponding assertion
for the upper bound is obtained by replacing \(b,d\)
and \(-\kappa\) by \(a,c\) and \(\kappa\), respectively.
\end{proof}

\begin{remark}
The two diagonal blocks describe visibility within
\(\Task\) and its orthogonal complement, whereas the
off-diagonal block describes the interaction between
these two subspaces. Since this interaction may have
either sign in the quadratic form, an increase in
task alignment does not, in general, imply an increase
in visibility.

If \(\Task=\HS\), the orthogonal complement is absent,
and the preceding theorem reduces to the usual bounds
for the Rayleigh quotient of \(H\).
\end{remark}

\begin{corollary}[Task-subspace visibility gap]
\label{cor:task-gap}
Suppose that
\[
    \lambda_{\min}^{\Task}(H)\geq b>0,
    \qquad
    \|H\|_{\op}\leq M.
\]
For every nonzero vector \(h\in\HS\), with
\(\rho=\rho_\Task(h)\), one has
\[
    \Geo_{\mu,\mathsf F}(h)
    \geq
    \rho b-2M\sqrt{\rho(1-\rho)}.
\]
If \(\Task\) is invariant under \(H\), this inequality
may be replaced by
\[
    \Geo_{\mu,\mathsf F}(h)\geq\rho b.
\]
Moreover, if \(h\in\Task\) and
\(\lambda_{\max}^{\Task}(H)\leq a\), then
\[
    \Geo_{\mu,\mathsf F}(h)\leq a.
\]
\end{corollary}

\begin{proof}
Since \(H\) is positive semidefinite, one has \(d\geq0\),
while the definition of \(B\) gives
\[
    \kappa=\|B\|_{\op}\leq\|H\|_{\op}\leq M.
\]
The first assertion therefore follows from
\eqref{eq:block-lower}. If \(\Task\) is invariant under
\(H\), then its orthogonal complement is also invariant,
so that \(B=0\). The second assertion follows by taking
\(\kappa=0\), and the last by taking \(\rho=1\) in
\eqref{eq:block-upper}.
\end{proof}

\subsection{Changes in task visibility}

We now compare the visibility operators associated with
two training intervals. The following criterion depends
only on the visibility of the task on the first interval
and on the block parameters and task alignment on the
second.

\begin{corollary}
\label{cor:task-transition}
Let \(H_0\) and \(H_1\) be the filtered visibility
operators associated with two training intervals,
and let \(\Task_0\) and \(\Task_1\) be the corresponding
task subspaces. Suppose that \(\Task_1\) is a proper
nonzero subspace of \(\HS\).

Suppose that \(h_0\in\Task_0\) is nonzero and that
\[
    \lambda_{\max}^{\Task_0}(H_0)\leq a_0.
\]
Let \(h_1\ne0\), put
\[
    \rho_1=\rho_{\Task_1}(h_1),
\]
and denote by \(b_1,d_1,\kappa_1\) the block parameters
of \(H_1\) relative to \(\Task_1\oplus\Task_1^\perp\).

If
\[
    \rho_1b_1+(1-\rho_1)d_1
    -2\kappa_1\sqrt{\rho_1(1-\rho_1)}
    >a_0,
\]
then
\[
    \frac{\langle h_1,H_1h_1\rangle}{\|h_1\|^2}
    >
    \frac{\langle h_0,H_0h_0\rangle}{\|h_0\|^2}.
\]

In particular, if \(\|H_1\|_{\op}\leq M_1\), it
suffices that
\[
    \rho_1b_1-2M_1\sqrt{\rho_1(1-\rho_1)}>a_0.
\]
If only a lower bound \(\rho_1\geq\rho_0\) is known,
where \(0\leq\rho_0\leq1\), a sufficient condition is
\[
    \min_{r\in[\rho_0,1]}
    \left(
        rb_1+(1-r)d_1
        -2\kappa_1\sqrt{r(1-r)}
    \right)
    >a_0.
\]
\end{corollary}

\begin{proof}
Since \(h_0\) belongs to \(\Task_0\), its Rayleigh
quotient is at most \(a_0\). By
\cref{thm:block-task}, the Rayleigh quotient of \(h_1\)
is at least
\[
    \rho_1b_1+(1-\rho_1)d_1
    -2\kappa_1\sqrt{\rho_1(1-\rho_1)},
\]
which proves the first assertion. The remaining
assertions follow by using \(d_1\geq0\) and
\(\kappa_1\leq M_1\), or by taking the minimum over
the possible values of \(\rho_1\), respectively.
\end{proof}

The lower bound of \cref{thm:block-task} also gives
an explicit condition on the alignment required to
reach a prescribed visibility threshold.

\begin{corollary}[Critical alignment]
\label{cor:critical-alignment}
Suppose that \(d=0\), and let \(\vartheta\) be such
that \(0<\vartheta<b\). Put
\[
    t_\vartheta
    =
    \frac{
        \sqrt{\kappa^2+\vartheta(b-\vartheta)}
        -\kappa
    }{\vartheta},
    \qquad
    \rho_\vartheta
    =
    \frac{1}{1+t_\vartheta^2}.
\]
If \(h\ne0\) and \(\rho_\Task(h)\geq\rho_\vartheta\),
then
\[
    \Geo_{\mu,\mathsf F}(h)\geq\vartheta.
\]
When \(\Task\) is invariant under \(H\), so that
\(\kappa=0\), one has
\[
    \rho_\vartheta=\frac{\vartheta}{b}.
\]
\end{corollary}

\begin{proof}
For \(0<\rho\leq1\), put
\[
    t=\sqrt{\frac{1-\rho}{\rho}},
    \qquad
    \rho=\frac{1}{1+t^2}.
\]
The lower bound in \eqref{eq:block-lower} is then
given by
\[
    \rho b-2\kappa\sqrt{\rho(1-\rho)}
    =
    \frac{b-2\kappa t}{1+t^2}.
\]
This quantity is at least \(\vartheta\) if and only if
\[
    \vartheta t^2+2\kappa t+\vartheta-b\leq0.
\]
Since \(0<\vartheta<b\), the positive root of the
corresponding quadratic polynomial is \(t_\vartheta\),
and the inequality holds precisely when
\(0\leq t\leq t_\vartheta\). As \(t\) is a decreasing
function of \(\rho\), this condition is equivalent to
\[
    \rho\geq\frac{1}{1+t_\vartheta^2}.
\]
The last assertion follows by setting \(\kappa=0\).
\end{proof}

\subsection{The number of visible task directions}

The smallest eigenvalue of the compression of \(H\)
to \(\Task\) gives a uniform bound for the visibility
of all directions in \(\Task\). When this eigenvalue
is small, the trace of the compression may nevertheless
give information about the dimension of a subspace
on which visibility is large.

\begin{proposition}
\label{prop:visible-dimension}
Let \(m=\dim\Task<\infty\), and suppose that the
eigenvalues of
\[
    A=\Pi H\Pi|_\Task,
\]
belong to the interval \([0,M]\). Put
\[
    \bar\lambda_\Task=\frac{\tr(A)}{m}.
\]
For \(0\leq\vartheta<M\), let \(N_\vartheta\) denote
the number of eigenvalues of \(A\), counted with
multiplicity, which are at least \(\vartheta\).
Then
\begin{equation}
\label{eq:visible-dimension}
    \frac{N_\vartheta}{m}
    \geq
    \max\left\{
        0,\,
        \frac{\bar\lambda_\Task-\vartheta}
             {M-\vartheta}
    \right\}.
\end{equation}
\end{proposition}

\begin{proof}
Let \(\lambda_1,\ldots,\lambda_m\) be the eigenvalues
of \(A\). By the definition of \(N_\vartheta\),
one has
\[
    \sum_{j=1}^m\lambda_j
    \leq
    (m-N_\vartheta)\vartheta+N_\vartheta M.
\]
Dividing by \(m\) gives
\[
    \bar\lambda_\Task
    \leq
    \vartheta+
    \frac{N_\vartheta}{m}(M-\vartheta),
\]
from which \eqref{eq:visible-dimension} follows,
since \(N_\vartheta\geq0\).
\end{proof}

\begin{remark}
Let \(E_\vartheta\subseteq\Task\) be the spectral
subspace of \(A\) corresponding to eigenvalues in
\([\vartheta,M]\). Then
\[
    \dim E_\vartheta=N_\vartheta,
\]
and every nonzero vector \(h\in E_\vartheta\) satisfies
\[
    \Geo_{\mu,\mathsf F}(h)\geq\vartheta.
\]
Thus a sufficiently large average task visibility
gives a lower bound for the dimension of a subspace
whose directions have visibility at least \(\vartheta\),
even when \(\lambda_{\min}^{\Task}(H)\) remains small.

This distinction between average and minimal visibility
will be useful in the study of sequential learning
in \cref{cor:sequential}, where different orthogonal
components of a task may become visible at different
times.
\end{remark}

\subsection{Stability of the task subspace}

The subspace \(\Task\) need not be invariant under
\(H\). However, when its visibility is separated from
that of its orthogonal complement and the coupling
between the two subspaces is sufficiently small,
\(\Task\) is close to a spectral subspace of \(H\).

\begin{corollary}
\label{cor:task-block-stability}
With the notation of \cref{thm:block-task}, suppose
that \(b>c\), and put \(r=\dim\Task\).
The \(r\) largest eigenvalues of \(H\) are separated
from the remaining eigenvalues by a gap of size at
least \(b-c\).

Let \(\widehat\Pi\) be the orthogonal projector onto
the spectral subspace associated with these \(r\)
largest eigenvalues. Then
\[
    \norm{(I-\widehat\Pi)\Pi}_{\op}
    \leq
    \frac{\kappa}{b-c}.
\]
\end{corollary}

\begin{proof}
Let \(N=\dim\HS\), and order the eigenvalues of \(H\)
so that
\[
    \lambda_1(H)\geq\cdots\geq\lambda_N(H).
\]
By the Courant--Fischer principle, one has
\[
\begin{aligned}
    \lambda_r(H)
    &\geq
    \inf_{\substack{v\in\Task\\\norm v=1}}
    \ip{v}{Hv}
    =b,\\
    \lambda_{r+1}(H)
    &\leq
    \sup_{\substack{v\in\Task^\perp\\\norm v=1}}
    \ip{v}{Hv}
    =c.
\end{aligned}
\]
Consequently,
\[
    \lambda_r(H)-\lambda_{r+1}(H)
    \geq b-c>0,
\]
which proves the first assertion.

Put \(H_0=A\oplus D\), so that
\[
    H-H_0=
    \begin{pmatrix}
        0&B\\
        B^*&0
    \end{pmatrix},
    \qquad
    \norm{H-H_0}_{\op}=\kappa.
\]
Let
\[
    P=I-\widehat\Pi,
    \qquad
    H_c=H|_{\ran P},
    \qquad
    X=P|_\Task.
\]
By the preceding eigenvalue estimates, one has
\[
    H_c\preceq cI,
    \qquad
    A\succeq bI.
\]

Since \(P\) commutes with \(H\), the operator \(X\)
satisfies the Sylvester equation
\[
    H_cX-XA=R,
\]
where
\[
    R=P(H-H_0)|_\Task,
    \qquad
    \norm R_{\op}\leq\kappa.
\]
Differentiating the function
\(t\mapsto e^{tH_c}Xe^{-tA}\), we obtain
\[
    \frac{\dd}{\dd t}
    \left(e^{tH_c}Xe^{-tA}\right)
    =
    e^{tH_c}Re^{-tA}.
\]
Moreover,
\[
    \norm{e^{tH_c}Xe^{-tA}}_{\op}
    \leq
    e^{-t(b-c)}\norm X_{\op}.
\]
Since \(b>c\), integration over \([0,\infty)\) gives
\[
    X=-\int_0^\infty e^{tH_c}Re^{-tA}\,\dd t.
\]
It follows that
\[
\begin{aligned}
    \norm X_{\op}
    &\leq
    \norm R_{\op}
    \int_0^\infty e^{-t(b-c)}\,\dd t\\
    &\leq
    \frac{\kappa}{b-c}.
\end{aligned}
\]
The assertion follows from the identity
\[
    \norm X_{\op}
    =
    \norm{(I-\widehat\Pi)\Pi}_{\op}.
\]
\end{proof}

\begin{remark}
The constant \(1\) in the preceding estimate is
optimal. Indeed, consider the matrices
\[
    H_\kappa=
    \begin{pmatrix}
        b&\kappa\\
        \kappa&c
    \end{pmatrix},
    \qquad b>c>0,
\]
with \(\Task=\operatorname{span}(e_1)\).
For sufficiently small \(\kappa>0\), the matrix
\(H_\kappa\) is positive definite. The angle
\(\theta_\kappa\) between \(\Task\) and the eigenspace
associated with the largest eigenvalue satisfies
\[
    \theta_\kappa
    =
    \frac12\arctan\left(\frac{2\kappa}{b-c}\right).
\]
Since
\[
    \norm{(I-\widehat\Pi)\Pi}_{\op}
    =
    \sin\theta_\kappa,
\]
it follows that
\[
    \lim_{\kappa\downarrow0}
    \frac{
        \norm{(I-\widehat\Pi)\Pi}_{\op}
    }{
        \kappa/(b-c)
    }
    =1.
\]
\end{remark}

Thus the separation between the task spectrum and
that of its orthogonal complement ensures the
existence of a corresponding spectral subspace
of \(H\), independently of the magnitude of the
off-diagonal block. The proximity of this spectral
subspace to \(\Task\) is controlled by the ratio
\(\kappa/(b-c)\). In particular, when this ratio
is small, the task subspace is close to an invariant
subspace of the visibility operator.

\subsection{Stability under resolvent filtering}

We shall now compare the visibility operator obtained
from the exact signal projector with that obtained
from the resolvent filter introduced in
\cref{sec:resolvent}. The approximation estimate of
\cref{thm:resolvent} gives a bound on the difference
between these two operators and hence on the
eigenvalues of their compressions to any task
subspace.

\begin{theorem}
\label{thm:soft-task-stability}
Let
\[
    H_{\mu,+}=P_+G^*C_\mu GP_+,
    \qquad
    H_{\mu,\alpha}=S_\alpha G^*C_\mu GS_\alpha.
\]
Then
\begin{equation}
\label{eq:soft-H-error}
    \|H_{\mu,+}-H_{\mu,\alpha}\|_{\op}
    \leq
    \beta_\mu\sqrt{\alpha\|W\|_{\op}}.
\end{equation}
Consequently, for every nonzero task subspace
\(\Task\), one has
\[
    \left|
        \lambda_{\min}^{\Task}(H_{\mu,+})
        -
        \lambda_{\min}^{\Task}(H_{\mu,\alpha})
    \right|
    \leq
    \beta_\mu\sqrt{\alpha\|W\|_{\op}},
\]
and
\[
    \left|
        \lambda_{\max}^{\Task}(H_{\mu,+})
        -
        \lambda_{\max}^{\Task}(H_{\mu,\alpha})
    \right|
    \leq
    \beta_\mu\sqrt{\alpha\|W\|_{\op}}.
\]
\end{theorem}

\begin{proof}
Put
\[
    X=C_\mu^{1/2}GP_+,
    \qquad
    Y=C_\mu^{1/2}GS_\alpha,
\]
so that
\[
    H_{\mu,+}=X^*X,
    \qquad
    H_{\mu,\alpha}=Y^*Y.
\]
Since \(P_+\) and \(S_\alpha\) are contractions,
it follows from \cref{thm:domination} that
\[
    \|X\|_{\op},\|Y\|_{\op}
    \leq
    \sqrt{\beta_\mu\|W\|_{\op}}.
\]
On the other hand, \cref{thm:resolvent} gives
\[
    \|X-Y\|_{\op}
    \leq
    \frac12\sqrt{\beta_\mu\alpha}.
\]
Using the identity
\[
    X^*X-Y^*Y
    =
    X^*(X-Y)+(X^*-Y^*)Y,
\]
we obtain
\[
\begin{aligned}
    \|H_{\mu,+}-H_{\mu,\alpha}\|_{\op}
    &\leq
    (\|X\|_{\op}+\|Y\|_{\op})
    \|X-Y\|_{\op}\\
    &\leq
    \beta_\mu\sqrt{\alpha\|W\|_{\op}},
\end{aligned}
\]
which proves \eqref{eq:soft-H-error}.

The assertions concerning the extremal eigenvalues
of the compressions to \(\Task\) follow from
\cref{quoted:weyl}.
\end{proof}

\begin{remark}
Although
\[
    0\preceq S_\alpha\preceq P_+,
\]
there is in general no corresponding inequality
between \(H_{\mu,\alpha}\) and \(H_{\mu,+}\) in the
Loewner order. Indeed, the transfer operator need
not preserve the orthogonality of the spectral
subspaces of \(W\), and contributions from distinct
spectral directions may partially cancel.

The estimate \eqref{eq:soft-H-error} is independent
of the smallest positive eigenvalue of \(W\).
It therefore allows us to compare the spectra of the
two visibility operators without assuming a spectral
gap at zero.

In particular, suppose that \(\Task\) and
\(\Task^\perp\) are nonzero and put
\[
    \gamma
    =
    \lambda_{\min}^{\Task}(H_{\mu,+})
    -
    \lambda_{\max}^{\Task^\perp}(H_{\mu,+}).
\]
If
\[
    \gamma>
    2\beta_\mu\sqrt{\alpha\|W\|_{\op}},
\]
then
\[
    \lambda_{\min}^{\Task}(H_{\mu,\alpha})
    >
    \lambda_{\max}^{\Task^\perp}(H_{\mu,\alpha}).
\]
Thus a sufficiently large separation between task
visibility and complement visibility is preserved
under resolvent filtering.
\end{remark}

\clearpage

\section{Evaluation visibility and population risk}
\label{sec:risk}

We examine the conditions under which the transfer
of training directions to evaluation outputs leads
to generalization. By introducing the target function
into the preceding framework, we derive an exact
identity for the variation of squared population
risk and obtain necessary and sufficient conditions
for its decrease in terms of the magnitude and
orientation of the evaluation displacement.
We then combine this identity with the task-subspace
bounds of the preceding section to obtain quantitative
estimates of population-risk reduction. Finally,
under a positive margin assumption on the target,
we relate population risk to classification error
and derive sufficient conditions for achieving a
prescribed classification accuracy. These results
establish the distinction between the emergence of
a visible signal and its contribution to generalization.

\subsection{Variation of population risk}

Let \(f^\star\in\HE\) be a target function, and denote by
\[
    e(t)=U_E(\theta(t))-f^\star,
\]
the corresponding residual at time \(t\). For an evaluation
law \(\mu\), we define the population squared risk by
\begin{equation}
\label{eq:population-risk}
    \Risk_\mu(t)
    =
    \frac12\norm{e(t)}_{C_\mu}^2,
    \qquad
    \norm v_{C_\mu}^2=\ip{v}{C_\mu v}_{\HE}.
\end{equation}

\begin{theorem}[Variation of squared risk]
\label{thm:risk-balance}
Let \(s<T\), put \(g_s=g(s)\), and denote by
\[
    d_\mu(s,T)=\G(T,s)g_s,
\]
the vector subtracted from the evaluation outputs over
the interval \([s,T]\). One has
\begin{align}
    U_E(\theta(T))-U_E(\theta(s))
    &=-d_\mu(s,T),
    \label{eq:eval-displacement}\\
    \Risk_\mu(s)-\Risk_\mu(T)
    &=
    \ip{e(s)}{C_\mu d_\mu(s,T)}
    -\frac12\norm{d_\mu(s,T)}_{C_\mu}^2.
    \label{eq:signed-progress}
\end{align}
In particular, if we put
\[
    \Prog_\mu(s,T)
    =
    \ip{e(s)}{C_\mu d_\mu(s,T)}
    -\frac12\norm{d_\mu(s,T)}_{C_\mu}^2,
\]
then
\[
    \Risk_\mu(T)
    =
    \Risk_\mu(s)-\Prog_\mu(s,T).
\]
\end{theorem}

\begin{proof}
The identity \eqref{eq:eval-displacement} is precisely
\eqref{eq:window-output-displacement} of
\cref{prop:window-identities}. It follows that
\[
    e(T)=e(s)-d_\mu(s,T),
\]
and hence
\[
\begin{aligned}
    \Risk_\mu(T)
    &=
    \frac12\norm{e(s)-d_\mu(s,T)}_{C_\mu}^2\\
    &=
    \frac12\norm{e(s)}_{C_\mu}^2
    -\ip{e(s)}{C_\mu d_\mu(s,T)}
    +\frac12\norm{d_\mu(s,T)}_{C_\mu}^2,
\end{aligned}
\]
which gives \eqref{eq:signed-progress}.
\end{proof}

Although we retain the notation \(d_\mu(s,T)\), the
vector itself is independent of \(\mu\). Its contribution
to the variation of population risk depends on the
evaluation law through \(C_\mu\).

\begin{corollary}
\label{cor:exact-clock}
For \(0\leq\delta<\Risk_\mu(s)\), let
\[
    t_{\mathrm{grok}}^\mu(\delta;s)
    =
    \inf\{T\geq s:\Risk_\mu(T)\leq\delta\},
\]
be the first time after \(s\) at which the population
risk is at most \(\delta\). Then
\begin{equation}
\label{eq:progress-clock}
    t_{\mathrm{grok}}^\mu(\delta;s)
    =
    \inf\left\{
        T\geq s:
        \Prog_\mu(s,T)\geq\Risk_\mu(s)-\delta
    \right\}.
\end{equation}
\end{corollary}

\begin{proof}
By \cref{thm:risk-balance}, the inequality
\(\Risk_\mu(T)\leq\delta\) is equivalent to
\[
    \Prog_\mu(s,T)\geq\Risk_\mu(s)-\delta,
\]
and the two sets occurring in the definitions of the
infima are therefore equal.
\end{proof}

The time introduced in this corollary concerns squared
risk rather than classification accuracy. The relation
between these two quantities will be considered below.

\subsection{The angle criterion}

The variation of risk in \eqref{eq:signed-progress}
depends on the norm of \(d_\mu(s,T)\) and on its position
relative to the residual \(e(s)\). The following corollary
expresses this dependence in terms of the angle between
the two vectors in the evaluation space.

\begin{corollary}
\label{cor:angle}
Suppose that
\[
    \norm{e(s)}_{C_\mu}>0,
    \qquad
    \norm{d_\mu(s,T)}_{C_\mu}>0,
\]
and define \(\omega_\mu(s,T)\in[0,\pi]\) by
\[
    \cos\omega_\mu(s,T)
    =
    \frac{
        \ip{e(s)}{C_\mu d_\mu(s,T)}
    }{
        \norm{e(s)}_{C_\mu}
        \norm{d_\mu(s,T)}_{C_\mu}
    }.
\]
Then \(\Risk_\mu(T)<\Risk_\mu(s)\) if and only if
\begin{equation}
\label{eq:angle-criterion}
    \cos\omega_\mu(s,T)>0
    \quad\text{and}\quad
    \norm{d_\mu(s,T)}_{C_\mu}
    <
    2\norm{e(s)}_{C_\mu}\cos\omega_\mu(s,T).
\end{equation}
\end{corollary}

\begin{proof}
It follows from \eqref{eq:signed-progress} that
\[
\begin{aligned}
    \Risk_\mu(s)-\Risk_\mu(T)
    &=
    \norm{d_\mu(s,T)}_{C_\mu}\\
    &\quad\times
    \left(
        \norm{e(s)}_{C_\mu}\cos\omega_\mu(s,T)
        -\frac12\norm{d_\mu(s,T)}_{C_\mu}
    \right),
\end{aligned}
\]
which is positive precisely under the conditions of
\eqref{eq:angle-criterion}.
\end{proof}

\begin{remark}
If \(\norm{d_\mu(s,T)}_{C_\mu}=0\), the population risk
is unchanged. If \(\norm{e(s)}_{C_\mu}=0\), one has
\[
    \Risk_\mu(T)
    =
    \frac12\norm{d_\mu(s,T)}_{C_\mu}^2,
\]
so that any nonzero displacement in \(L^2(\mu;\R^p)\)
increases the risk.

Thus neither a positive angle cosine nor a large
evaluation displacement is sufficient by itself to
ensure a decrease in population risk. The magnitude
of the displacement must also satisfy the inequality
in \eqref{eq:angle-criterion}.
\end{remark}

\subsection{Task visibility and risk variation}

We shall now combine the estimates of \cref{sec:task}
with the signed identity of \cref{thm:risk-balance}.
The resulting bound depends on the block parameters
of the visibility operator, the alignment of the initial
output gradient with the task subspace, and the angle
between the evaluation displacement and the residual.

\begin{theorem}
\label{thm:task-signed-progress}
Let \(H_{\mu,+}=P_+G^*C_\mu GP_+\), and suppose that
\(g_s\ne0\). With the notation of \cref{thm:block-task},
put
\[
    \rho=\rho_\Task(g_s),
\]
and define
\[
\begin{aligned}
    L_\mu
    &=
    \rho b+(1-\rho)d
    -2\kappa\sqrt{\rho(1-\rho)},\\
    U_\mu
    &=
    \rho a+(1-\rho)c
    +2\kappa\sqrt{\rho(1-\rho)}.
\end{aligned}
\]
Let
\[
    q_-=\|g_s\|\sqrt{\max\{0,L_\mu\}},
    \qquad
    q_+=\|g_s\|\sqrt{\max\{0,U_\mu\}}.
\]
Suppose that the angle \(\omega_\mu(s,T)\) is defined
and that
\[
    \cos\omega_\mu(s,T)\geq\zeta_\mu>0.
\]
Then
\begin{equation}
\label{eq:task-progress-bound}
    \Prog_\mu(s,T)\geq p_\mu,
\end{equation}
where
\[
    p_\mu
    =
    \min_{\sigma\in\{q_-,q_+\}}
    \left\{
        \zeta_\mu\norm{e(s)}_{C_\mu}\sigma
        -\frac12\sigma^2
    \right\}.
\]
In particular, \(p_\mu>0\) if
\[
    0<q_-\leq q_+
    <2\zeta_\mu\norm{e(s)}_{C_\mu}.
\]
\end{theorem}

\begin{proof}
Since \(GP_+=G\), it follows from the definition of
\(H_{\mu,+}\) that
\[
    \norm{d_\mu(s,T)}_{C_\mu}^2
    =
    \ip{g_s}{H_{\mu,+}g_s}_{\HS}.
\]
The bounds of \cref{thm:block-task} therefore give
\[
    q_-\leq\norm{d_\mu(s,T)}_{C_\mu}\leq q_+.
\]

Put \(q=\norm{d_\mu(s,T)}_{C_\mu}\). By
\cref{thm:risk-balance} and the hypothesis on
\(\omega_\mu(s,T)\), one has
\[
    \Prog_\mu(s,T)
    \geq
    \zeta_\mu\norm{e(s)}_{C_\mu}q
    -\frac12q^2.
\]
The function
\[
    \sigma\longmapsto
    \zeta_\mu\norm{e(s)}_{C_\mu}\sigma
    -\frac12\sigma^2,
\]
is concave, so that its minimum on the interval
\([q_-,q_+]\) is attained at one of the endpoints.
This proves \eqref{eq:task-progress-bound}.

If
\[
    0<q_-\leq q_+
    <2\zeta_\mu\norm{e(s)}_{C_\mu},
\]
the function is positive throughout \([q_-,q_+]\),
which proves the last assertion.
\end{proof}

\begin{example}
\label{ex:unsigned}
Let the evaluation space be \(\R\), with its usual
inner product, and suppose that \(e(s)=1\). Consider
the two vectors
\[
    d_+=1,
    \qquad
    d_-=-1.
\]
Although
\[
    |d_+|^2=|d_-|^2,
\]
their contributions to the variation of population
risk are different, since
\[
    \Prog(d_+)=\frac12,
    \qquad
    \Prog(d_-)=-\frac32.
\]
In the first case the residual vanishes, whereas in
the second its norm is doubled. Thus knowledge of
the displacement energy alone does not determine
even the sign of the variation of population risk.
\end{example}

\subsection{Population risk and classification error}

Suppose now that \(p=K\geq2\), so that
\(F(\theta,z)\in\R^K\) is a vector of class logits,
and let
\[
    y:\mathcal Z\longrightarrow\{1,\ldots,K\},
\]
be a measurable function assigning to each input
\(z\) its true class.

We assume that the target function \(f^\star\) has
a positive margin \(\gamma\) under the evaluation
law \(\mu\), i.e.\
\begin{equation}
\label{eq:target-margin}
    f^\star_{y(z)}(z)
    -
    \max_{j\ne y(z)}f^\star_j(z)
    \geq\gamma>0,
\end{equation}
for \(\mu\)-almost every \(z\). Under this assumption,
the squared distance from the target gives an upper
bound for the classification error.

\begin{theorem}
\label{thm:classification}
Suppose that \eqref{eq:target-margin} holds and that
there exists a constant \(c_\mu>0\) such that
\[
    \ip{v}{C_\mu v}_{\HE}
    \geq
    c_\mu\int_{\mathcal Z}
    \norm{v(z)}_2^2\,\dd\mu(z),
\]
for every \(v\in\HE\). Denote by
\[
    \operatorname{err}_\mu(t)
    =
    \mu\left\{
        z:
        \argmax_j F_j(\theta(t),z)\ne y(z)
    \right\},
\]
the classification error at time \(t\), where ties
are resolved by a fixed measurable rule. The
corresponding classification accuracy is denoted by
\[
    A_\mu(t)=1-\Err_\mu(t).
\]
Then
\begin{equation}
\label{eq:error-risk}
    \operatorname{err}_\mu(t)
    \leq
    \frac{4\Risk_\mu(t)}{c_\mu\gamma^2}.
\end{equation}
In particular, if \(0<\delta<1\) and
\[
    \Prog_\mu(s,T)
    \geq
    \Risk_\mu(s)-\frac{c_\mu\gamma^2\delta}{4},
\]
then
\[
    A_\mu(T)\geq1-\delta.
\]
\end{theorem}

\begin{proof}
Let \(z\) be a point at which the predicted class is
different from \(y=y(z)\) and at which the margin
condition \eqref{eq:target-margin} holds. There exists
an index \(j\ne y\) such that
\[
    F_j(\theta(t),z)\geq F_y(\theta(t),z).
\]
Put
\[
    a=F_y(\theta(t),z)-f^\star_y(z),
    \qquad
    b=F_j(\theta(t),z)-f^\star_j(z).
\]
By the choice of \(j\) and the margin condition,
\[
\begin{aligned}
    b-a
    &=
    F_j(\theta(t),z)-F_y(\theta(t),z)
    +f^\star_y(z)-f^\star_j(z)\\
    &\geq\gamma,
\end{aligned}
\]
and consequently
\[
    \norm{F(\theta(t),z)-f^\star(z)}_2^2
    \geq
    a^2+b^2
    \geq
    \frac12(b-a)^2
    \geq
    \frac{\gamma^2}{2}.
\]
Integrating over the set of misclassified points
gives
\[
    \int_{\mathcal Z}
    \norm{e(t,z)}_2^2\,\dd\mu(z)
    \geq
    \frac{\gamma^2}{2}\operatorname{err}_\mu(t).
\]
It follows from the hypothesis on \(C_\mu\) that
\[
\begin{aligned}
    \Risk_\mu(t)
    &=
    \frac12\norm{e(t)}_{C_\mu}^2\\
    &\geq
    \frac{c_\mu}{2}
    \int_{\mathcal Z}
    \norm{e(t,z)}_2^2\,\dd\mu(z)\\
    &\geq
    \frac{c_\mu\gamma^2}{4}
    \operatorname{err}_\mu(t),
\end{aligned}
\]
which proves \eqref{eq:error-risk}.

For the last assertion, observe that
\cref{thm:risk-balance} gives
\[
    \Risk_\mu(T)
    =
    \Risk_\mu(s)-\Prog_\mu(s,T)
    \leq
    \frac{c_\mu\gamma^2\delta}{4}.
\]
Applying \eqref{eq:error-risk} at time \(T\), we
obtain \(\operatorname{err}_\mu(T)\leq\delta\),
as desired.
\end{proof}

\begin{remark}
For the multiplication operator \(C_\mu\) introduced
in \cref{sec:setting}, one has
\[
    \ip{v}{C_\mu v}_{\HE}
    =
    \int_{\mathcal Z}\norm{v(z)}_2^2\,\dd\mu(z),
\]
so that \(c_\mu=1\) may be taken in the preceding
theorem. We retain the constant \(c_\mu\) in order
to allow for more general positive evaluation metrics
satisfying the stated lower bound.
\end{remark}

Combining \cref{thm:task-signed-progress} and
\cref{thm:classification}, we obtain a sufficient
condition for a prescribed classification accuracy
in terms of the task spectrum and the direction of
the evaluation displacement.

\begin{corollary}
\label{cor:task-accuracy}
Suppose that the hypotheses of
\cref{thm:task-signed-progress,thm:classification}
are satisfied, and denote by \(\gamma_\mu\) the
target margin under \(\mu\).

If the constant \(p_\mu\) of
\eqref{eq:task-progress-bound} satisfies
\[
    \Risk_\mu(s)-p_\mu
    \leq
    \frac{c_\mu\gamma_\mu^2\delta}{4},
\]
then
\[
    A_\mu(T)\geq1-\delta.
\]
\end{corollary}

\begin{proof}
By \cref{thm:task-signed-progress}, one has
\[
    \Risk_\mu(T)
    =
    \Risk_\mu(s)-\Prog_\mu(s,T)
    \leq
    \Risk_\mu(s)-p_\mu,
\]
and the assertion follows from
\eqref{eq:error-risk}.
\end{proof}

\begin{remark}
The preceding results separate the roles of the
visibility operator, the target residual, and the
classification margin. The first determines the
magnitude of the evaluation displacement, the second
determines its contribution to the variation of risk,
and the third relates the resulting risk to
classification error.

The estimate of \cref{thm:classification} differs from
statistical margin bounds involving the complexity of
a class of predictors~\cite{bartlett2017margin}, since
it concerns the relation between two population
quantities for a given predictor. No estimate of
population risk in terms of empirical risk is involved.
\end{remark}

\clearpage

\section{Evolution of evaluation visibility}
\label{sec:kinetics}

We study the evolution of evaluation visibility by
fixing the left endpoint \(s\) and considering the
dependence of the dissipation and transfer operators
\(\W(s,T)\) and \(\G(T,s)\) on \(T\).
By differentiating the filtered evaluation displacement,
we distinguish two contributions to its evolution,
namely direct transfer to the evaluation space and
changes in the effective signal channel. We then
derive bounds for the evolution of the extremal
eigenvalues of the task visibility operator and
obtain sufficient conditions for reaching prescribed
visibility thresholds. These estimates also allow
us to compare the rates at which a task becomes
visible under different evaluation laws. Finally,
we obtain bounds on the crossing times of individual
training directions and distinguish the evolution
of cumulative dissipation, evaluation visibility,
and population risk.

For \(T>s\), put
\begin{align}
    A(T;s)
    &=
    M(T)^{1/2}J_S(T)^*\Pg(T,s),
    \label{eq:A-endpoint}\\
    Q(T;s)
    &=
    A(T;s)^*A(T;s),
    \label{eq:Q-endpoint}\\
    L(T;s)
    &=
    K_{ES}^M(T)\Pg(T,s).
    \label{eq:L-endpoint}
\end{align}
It follows from the definitions of \(\W\) and \(\G\)
that
\begin{equation}
\label{eq:endpoint-derivatives}
    \partial_T\W(s,T)=Q(T;s)\succeq0,
    \qquad
    \partial_T\G(T,s)=L(T;s),
\end{equation}
for almost every \(T\). In particular, \(\W(s,T)\)
is a nondecreasing function of \(T\) in the Loewner
order. No analogous monotonicity property holds in
general for the visibility operators.

\subsection{The variation of filtered visibility}

\begin{theorem}
\label{thm:endpoint-balance}
Let \(h\in\HS\) and \(\alpha>0\) be fixed. For simplicity,
we write
\[
    W(T)=\W(s,T),
    \qquad
    G(T)=\G(T,s),
    \qquad
    S_\alpha(T)=\Salpha(W(T)),
\]
and put
\[
    d_{\mu,\alpha}(T)=G(T)S_\alpha(T)h.
\]
At almost every \(T>s\), one has
\begin{equation}
\label{eq:d-balance}
    \dot d_{\mu,\alpha}(T)
    =
    L(T;s)S_\alpha(T)h
    +
    \alpha G(T)\Ralpha(T)Q(T;s)\Ralpha(T)h,
\end{equation}
where \(\Ralpha(T)=(W(T)+\alpha\Id)^{-1}\).
Consequently,
\begin{align}
    \frac{\dd}{\dd T}
    \norm{d_{\mu,\alpha}(T)}_{C_\mu}^2
    &=
    2\ip{d_{\mu,\alpha}}{C_\mu LS_\alpha h}
    \nonumber\\
    &\quad+
    2\alpha
    \ip{d_{\mu,\alpha}}
       {C_\mu G\Ralpha Q\Ralpha h}.
    \label{eq:energy-balance}
\end{align}

If \(h=g(s)\), the vector \(d_\mu(T)=G(T)h\)
satisfies
\[
    e(T)=e(s)-d_\mu(T),
\]
and the corresponding variation of population risk
is given by
\begin{equation}
\label{eq:risk-rate}
    \frac{\dd}{\dd T}\Risk_\mu(T)
    =
    -\ip{e(T)}{C_\mu L(T;s)h}.
\end{equation}
\end{theorem}

\begin{proof}
Differentiating the product \(G(T)S_\alpha(T)h\),
we obtain
\[
    \dot d_{\mu,\alpha}(T)
    =
    \dot G(T)S_\alpha(T)h
    +
    G(T)\dot S_\alpha(T)h.
\]
By \eqref{eq:endpoint-derivatives} and
\eqref{eq:filter-derivative}, one has
\[
    \dot G(T)=L(T;s),
    \qquad
    \dot S_\alpha(T)
    =
    \alpha\Ralpha(T)Q(T;s)\Ralpha(T),
\]
which gives \eqref{eq:d-balance}. The identity
\eqref{eq:energy-balance} follows by substituting
this expression into
\[
    \frac{\dd}{\dd T}
    \norm{d_{\mu,\alpha}(T)}_{C_\mu}^2
    =
    2\ip{d_{\mu,\alpha}(T)}
       {C_\mu\dot d_{\mu,\alpha}(T)}.
\]

For \(h=g(s)\), it follows from
\cref{prop:window-identities} that
\[
    e(T)=e(s)-G(T)h,
    \qquad
    \dot e(T)=-L(T;s)h,
\]
and hence
\[
    \frac{\dd}{\dd T}\Risk_\mu(T)
    =
    \ip{e(T)}{C_\mu\dot e(T)}
    =
    -\ip{e(T)}{C_\mu L(T;s)h}.
\]
This proves \eqref{eq:risk-rate}.
\end{proof}

\begin{remark}
The two terms in \eqref{eq:d-balance} have different
origins. The first is due to the variation of the
transfer operator \(G\), with the signal filter held
fixed, whereas the second is due to the variation
of the filter itself.

In particular, a variation of \(W\) may change the
filtered evaluation displacement even when \(G\)
remains constant. We shall refer to these two
contributions as direct transfer and channel opening,
respectively.

Although
\[
    \dot W(T)\succeq0,
    \qquad
    \dot S_\alpha(T)\succeq0,
\]
neither of the two terms on the right-hand side of
\eqref{eq:energy-balance} is necessarily positive.
The monotonicity of the dissipation operator and
that of the signal filter do not imply the monotonicity
of evaluation visibility.
\end{remark}

\subsection{Variation of the task spectrum}

We now consider a fixed task subspace \(\Task\) and
study the dependence on \(T\) of the extremal eigenvalues
of the compression of \(H_{\mu,\alpha}(T)\) to \(\Task\).
Rather than differentiating these eigenvalues directly,
we obtain bounds by integrating the derivatives of
the corresponding Rayleigh quotients.

\begin{theorem}
\label{thm:task-clock}
Let \(\Task\) be a fixed nonzero task subspace on
an interval \([T_0,T_1]\), with \(s\leq T_0<T_1\).
For \(v\in\Task\), put
\[
    d_v(T)=G(T)S_\alpha(T)v,
\]
so that, by \cref{thm:endpoint-balance},
\[
    \dot d_v(T)
    =
    L(T;s)S_\alpha(T)v
    +
    \alpha G(T)\Ralpha(T)Q(T;s)\Ralpha(T)v,
\]
for almost every \(T\in[T_0,T_1]\).

Define
\[
\begin{aligned}
    \chi^-_{\mu,\Task}(T)
    &=
    2\inf_{\substack{v\in\Task\\\|v\|=1}}
    \Re\langle d_v(T),C_\mu\dot d_v(T)\rangle,\\
    \chi^+_{\mu,\Task}(T)
    &=
    2\sup_{\substack{v\in\Task\\\|v\|=1}}
    \Re\langle d_v(T),C_\mu\dot d_v(T)\rangle.
\end{aligned}
\]
Then, for every \(T\in[T_0,T_1]\), one has
\begin{align}
    \lambda_{\min}^{\Task}(H_{\mu,\alpha}(T))
    &\geq
    \lambda_{\min}^{\Task}(H_{\mu,\alpha}(T_0))
    +
    \int_{T_0}^{T}
    \chi^-_{\mu,\Task}(\tau)\,\dd\tau,
    \label{eq:task-clock-lower}\\
    \lambda_{\max}^{\Task}(H_{\mu,\alpha}(T))
    &\leq
    \lambda_{\max}^{\Task}(H_{\mu,\alpha}(T_0))
    +
    \int_{T_0}^{T}
    \chi^+_{\mu,\Task}(\tau)\,\dd\tau.
    \label{eq:task-clock-upper}
\end{align}
\end{theorem}

\begin{proof}
For a fixed unit vector \(v\in\Task\), the definition
of \(H_{\mu,\alpha}(T)\) gives
\[
    \langle v,H_{\mu,\alpha}(T)v\rangle
    =
    \norm{d_v(T)}_{C_\mu}^2.
\]
Differentiating and applying
\cref{thm:endpoint-balance}, we obtain
\[
    \frac{\dd}{\dd T}
    \langle v,H_{\mu,\alpha}(T)v\rangle
    =
    2\Re\langle d_v(T),C_\mu\dot d_v(T)\rangle.
\]
It follows from the definitions of
\(\chi^-_{\mu,\Task}\) and \(\chi^+_{\mu,\Task}\)
that
\[
    \chi^-_{\mu,\Task}(T)
    \leq
    \frac{\dd}{\dd T}
    \langle v,H_{\mu,\alpha}(T)v\rangle
    \leq
    \chi^+_{\mu,\Task}(T),
\]
for almost every \(T\).

Integrating the first inequality over \([T_0,T]\)
gives
\[
    \langle v,H_{\mu,\alpha}(T)v\rangle
    \geq
    \langle v,H_{\mu,\alpha}(T_0)v\rangle
    +
    \int_{T_0}^{T}
    \chi^-_{\mu,\Task}(\tau)\,\dd\tau.
\]
Since this inequality holds for every unit vector
\(v\in\Task\), taking the infimum over such vectors
gives \eqref{eq:task-clock-lower}. The proof of
\eqref{eq:task-clock-upper} is analogous, with the
infimum replaced by the supremum.
\end{proof}

\begin{remark}
The quantities \(\chi^-_{\mu,\Task}\) and
\(\chi^+_{\mu,\Task}\) admit a spectral interpretation.
Indeed, if we put
\[
    X_\alpha(T)
    =
    C_\mu^{1/2}G(T)S_\alpha(T),
\]
then
\[
    H_{\mu,\alpha}(T)
    =
    X_\alpha(T)^*X_\alpha(T),
\]
and consequently
\[
    \dot H_{\mu,\alpha}(T)
    =
    \dot X_\alpha(T)^*X_\alpha(T)
    +
    X_\alpha(T)^*\dot X_\alpha(T).
\]
Thus, at every point where these derivatives exist,
\[
    \chi^-_{\mu,\Task}(T)
    =
    \lambda_{\min}^{\Task}(\dot H_{\mu,\alpha}(T)),
    \qquad
    \chi^+_{\mu,\Task}(T)
    =
    \lambda_{\max}^{\Task}(\dot H_{\mu,\alpha}(T)).
\]
The inequalities of \cref{thm:task-clock} may therefore
be regarded as integrated min--max estimates for the
variation of the task spectrum.
\end{remark}

An immediate consequence of \cref{thm:task-clock}
is an upper bound for the first time at which the
normalized visibility of every nonzero direction
in \(\Task\) is at least a prescribed level.

Let \(\vartheta>0\) and put
\[
    T_\vartheta
    =
    \inf\left\{
        T\in[T_0,T_1]:
        \lambda_{\min}^{\Task}(H_{\mu,\alpha}(T))
        \geq\vartheta
    \right\}.
\]
By \eqref{eq:task-clock-lower}, one has
\[
    T_\vartheta
    \leq
    \inf\left\{
        T\in[T_0,T_1]:
        \lambda_{\min}^{\Task}(H_{\mu,\alpha}(T_0))
        +
        \int_{T_0}^{T}
        \chi^-_{\mu,\Task}(\tau)\,\dd\tau
        \geq\vartheta
    \right\},
\]
with the convention that the infimum of the empty
set is \(+\infty\). This estimate does not require
the task spectrum to be monotone.

\subsection{Comparison of two evaluation laws}

The preceding estimates may be applied simultaneously
to several evaluation laws. Since the operators \(W\),
\(G\), and \(S_\alpha\) are independent of the law,
the difference between their respective visibility
rates arises from the operators \(C_\mu\).

\begin{corollary}
\label{cor:kinetic-selective}
Let \(\mu\) and \(\nu\) be two evaluation laws,
and suppose that, at some time \(T\in[T_0,T_1]\),
\[
    \lambda_{\min}^{\Task}(H_{\mu,\alpha}(T_0))
    +
    \int_{T_0}^{T}
    \chi^-_{\mu,\Task}(\tau)\,\dd\tau
    \geq b_\mu,
\]
and
\[
    \lambda_{\max}^{\Task}(H_{\nu,\alpha}(T_0))
    +
    \int_{T_0}^{T}
    \chi^+_{\nu,\Task}(\tau)\,\dd\tau
    \leq a_\nu.
\]
Then
\[
    \lambda_{\min}^{\Task}(H_{\mu,\alpha}(T))
    \geq b_\mu,
    \qquad
    \lambda_{\max}^{\Task}(H_{\nu,\alpha}(T))
    \leq a_\nu.
\]

In particular, if
\[
    b_\mu\geq\vartheta_\mu^+,
    \qquad
    a_\nu\leq\vartheta_\nu^-,
\]
the spectral hypotheses of
\cref{thm:abstract-mu-not-nu} are satisfied for
the corresponding filtered visibility operators.

If the additional orientation and margin hypotheses
of \cref{thm:explicit-mu-not-nu} are satisfied,
with the required estimates for the exact-channel
operators, then
\[
    A_\mu(T)\geq1-\delta,
    \qquad
    A_\nu(T)<1-\delta.
\]
\end{corollary}

\begin{proof}
The first two inequalities follow by applying
\cref{thm:task-clock} to \(\mu\) and \(\nu\),
respectively. The remaining assertions follow from
the corresponding results on selective generalization.
When passing from filtered to exact-channel operators,
one may use the perturbation estimate of
\cref{thm:soft-task-stability}.
\end{proof}

Thus the estimates on the variation of the task
spectrum give conditions under which a task may
become visible under one evaluation law while
remaining poorly visible under another. Whether
this difference results in different classification
accuracies depends on the signed progress and
margin conditions of \cref{sec:risk}.

\subsection{Crossing times for individual directions}

We return to the case of a fixed direction \(h\in\HS\).
The preceding arguments concern the squared norm
of the filtered evaluation displacement, but it
will also be useful to consider the derivative
of its norm.

\begin{corollary}
\label{cor:kinetic-crossing}
Put
\[
    q_\alpha(T)
    =
    \norm{d_{\mu,\alpha}(T)}_{C_\mu},
\]
and, at points where \(q_\alpha(T)>0\) and the
derivative exists, define
\begin{equation}
\label{eq:coherent-speed}
    \kappa_{\mu,\alpha,h}(T)
    =
    \frac{
        \ip{d_{\mu,\alpha}}
        {C_\mu\bigl(
            LS_\alpha h+
            \alpha G\Ralpha Q\Ralpha h
        \bigr)}
    }{q_\alpha(T)}.
\end{equation}
Then
\[
    q_\alpha'(T)=\kappa_{\mu,\alpha,h}(T).
\]

Suppose that \(q_\alpha(T)>0\) for almost every
\(T\in[T_0,T_1]\) and that there exists a constant
\(v_0>0\) such that
\[
    \kappa_{\mu,\alpha,h}(T)\geq v_0,
\]
for almost every \(T\in[T_0,T_1]\). Then
\[
    q_\alpha(T)
    \geq
    q_\alpha(T_0)+v_0(T-T_0),
\]
for every \(T\in[T_0,T_1]\).

In particular, let \(\Theta>0\) and put
\[
    T_\Theta
    =
    \inf\left\{
        T\in[T_0,T_1]:
        \norm{d_{\mu,\alpha}(T)}_{C_\mu}^2
        \geq\Theta
    \right\}.
\]
One has
\[
    T_\Theta
    \leq
    T_0+
    \frac{
        \bigl(\sqrt{\Theta}-q_\alpha(T_0)\bigr)_+
    }{v_0},
\]
provided that the right-hand side belongs to
\([T_0,T_1]\).
\end{corollary}

\begin{proof}
The function \(q_\alpha\) is absolutely continuous,
since \(d_{\mu,\alpha}\) is absolutely continuous
and \(C_\mu^{1/2}\) is bounded. At every point where
\(q_\alpha(T)>0\) and \(d_{\mu,\alpha}\) is
differentiable, one has
\[
    q_\alpha'(T)
    =
    \frac{
        \ip{d_{\mu,\alpha}(T)}
        {C_\mu\dot d_{\mu,\alpha}(T)}
    }{q_\alpha(T)},
\]
which, together with \eqref{eq:d-balance}, gives
\eqref{eq:coherent-speed}.

Under the stated hypotheses, integration yields
\[
\begin{aligned}
    q_\alpha(T)-q_\alpha(T_0)
    &=
    \int_{T_0}^{T}
    \kappa_{\mu,\alpha,h}(\tau)\,\dd\tau\\
    &\geq v_0(T-T_0).
\end{aligned}
\]
If \(q_\alpha(T_0)\geq\sqrt{\Theta}\), the assertion
is immediate. Otherwise, the last inequality gives
\(q_\alpha(T)\geq\sqrt{\Theta}\) whenever
\[
    T-T_0
    \geq
    \frac{\sqrt{\Theta}-q_\alpha(T_0)}{v_0},
\]
from which the claimed bound follows.
\end{proof}

\begin{remark}
The hypothesis on \(\kappa_{\mu,\alpha,h}\) is a
condition on the direction of the derivative of the
filtered displacement, rather than on its norm alone.
Indeed, a large value of
\[
    \norm{\dot d_{\mu,\alpha}(T)}_{C_\mu},
\]
need not imply a large increase in
\(\norm{d_{\mu,\alpha}(T)}_{C_\mu}\), since the
derivative may be nearly orthogonal to the
displacement or may point in the opposite direction.

The distinction is analogous to that encountered
in \cref{sec:risk}. In the present case, the
variation of visibility is determined by the
position of \(\dot d_{\mu,\alpha}(T)\) relative
to \(d_{\mu,\alpha}(T)\), whereas the variation
of population risk is determined by the position
of \(L(T;s)h\) relative to the residual \(e(T)\).

Thus the increase of cumulative dissipation,
the increase of evaluation visibility, and the
decrease of population risk are three distinct
properties of the training trajectory.
\end{remark}

\clearpage

\section{Spectral transport and reservoir escape}
\label{sec:spectral-transport}

We study the evolution of the spectral subspaces of
the cumulative dissipation operator and identify two
mechanisms by which a training direction may enter
the effective signal channel. The first is spectral
activation, whereby eigenvalues cross a prescribed
threshold, and the second is spectral transport,
whereby the corresponding eigenspaces change their
orientation. We establish an evolution formula for
the spectral projector without restrictions on
eigenvalue multiplicities and characterize the
conditions under which spectral transport vanishes.
We also distinguish the evolution of the exact
signal channel from that of the effective signal
channel and examine how these spectral changes
relate to evaluation visibility and population risk.

Throughout this section, we continue to regard the
left endpoint \(s\) as fixed and write
\(W(T)=\W(s,T)\). Although \(W(T)\) is nondecreasing
in the Loewner order, its spectral subspaces need
not remain fixed. In particular, the subspace
associated with eigenvalues above a positive
threshold may change even when its dimension
remains constant. To describe this phenomenon,
we shall study the dependence of the corresponding
spectral projector on the cumulative dissipation
operator \(W(T)\) and its increment \(Q(T;s)\).

\subsection{Monotonicity of the dissipation operator}

\begin{proposition}
\label{prop:nested-reservoir}
For \(s<T_1<T_2\), one has
\[
    \W(s,T_1)\preceq\W(s,T_2),
    \qquad
    \Ker\W(s,T_2)\subseteq\Ker\W(s,T_1).
\]
Moreover, the eigenvalues of \(\W(s,T)\), arranged
in increasing order and counted with multiplicity,
are nondecreasing functions of \(T\).
\end{proposition}

\begin{proof}
It follows from \eqref{eq:endpoint-derivatives} that
\[
    \W(s,T_2)-\W(s,T_1)
    =
    \int_{T_1}^{T_2}Q(\tau;s)\,\dd\tau
    \succeq0.
\]
If \(h\in\Ker\W(s,T_2)\), then
\[
    0
    =
    \ip{h}{\W(s,T_2)h}
    \geq
    \ip{h}{\W(s,T_1)h}
    \geq0,
\]
and the positivity of \(\W(s,T_1)\) implies that
\(h\in\Ker\W(s,T_1)\). The assertion concerning the
eigenvalues follows from the min--max principle.
\end{proof}

In particular, the exact reservoirs form a decreasing
family of subspaces, while the corresponding signal
channels form an increasing family. On an interval
where the rank of \(W(T)\) is constant, both subspaces
are therefore constant.

The situation is different for a signal channel defined
by a positive spectral threshold. Its dimension is
nondecreasing, but the subspace itself may vary even
when its dimension remains constant. This is possible
because the operators \(W(T)\) and \(Q(T;s)\) need
not commute.

\subsection{Variation of the spectral projector}

Let \(\alpha>0\) and denote by
\[
    P_\alpha(T)=\1_{(\alpha,\infty)}(W(T)),
\]
the orthogonal projector onto the sum of the eigenspaces
of \(W(T)\) corresponding to eigenvalues greater than
\(\alpha\). We first study the variation of \(P_\alpha(T)\)
on an interval on which its rank is constant.

For each distinct eigenvalue \(\lambda\) of \(W(T)\),
let \(E_\lambda(T)\) denote the orthogonal projector
onto the corresponding eigenspace, so that
\[
    W(T)
    =
    \sum_{\lambda\in\sigma(W(T))}
    \lambda E_\lambda(T),
    \qquad
    P_\alpha(T)
    =
    \sum_{\lambda>\alpha}E_\lambda(T).
\]
The use of spectral projectors rather than individual
eigenvectors allows us to treat eigenvalues of arbitrary
multiplicity.

\begin{theorem}[Spectral transport]
\label{thm:projector-transport}
Suppose that \(W(T)\) is continuously differentiable
on an open interval on which \(P_\alpha(T)\) has
constant rank. Put \(Q=\dot W(T)\), which agrees with
\(Q(T;s)\) almost everywhere by
\eqref{eq:endpoint-derivatives}.

\begin{enumerate}[label=(\roman*)]

\item Let \(\Gamma\) be a positively oriented contour
enclosing the eigenvalues greater than \(\alpha\)
and no others, and chosen locally independently of
\(T\). Then
\begin{equation}
\label{eq:riesz-derivative}
    \dot P_\alpha(T)
    =
    \frac{1}{2\pi i}
    \oint_\Gamma
    (z\Id-W(T))^{-1}
    Q
    (z\Id-W(T))^{-1}\,\dd z.
\end{equation}

\item With the notation introduced above, one has
\begin{equation}
\label{eq:projector-block-transport}
    \dot P_\alpha
    =
    \sum_{\substack{
        \lambda\in\sigma(W),\ \lambda>\alpha\\
        \eta\in\sigma(W),\ \eta\leq\alpha
    }}
    \frac{
        E_\eta Q E_\lambda+
        E_\lambda Q E_\eta
    }{\lambda-\eta}.
\end{equation}
This formula holds without any assumption on the
multiplicities of the eigenvalues.

\item For a fixed vector \(q\in\HS\), one has
\begin{equation}
\label{eq:task-mass-block-transport}
    \frac{\dd}{\dd T}\norm{P_\alpha(T)q}^2
    =
    2
    \sum_{\substack{
        \lambda\in\sigma(W),\ \lambda>\alpha\\
        \eta\in\sigma(W),\ \eta\leq\alpha
    }}
    \frac{
        \ip{E_\eta q}{Q E_\lambda q}
    }{\lambda-\eta}.
\end{equation}

\item The following conditions are equivalent.
\begin{align}
    \dot P_\alpha&=0,
    \label{eq:stationary-projector}\\
    (\Id-P_\alpha)QP_\alpha&=0,
    \label{eq:no-cross-channel-transport}\\
    [Q,P_\alpha]&=0.
    \label{eq:projector-commutation}
\end{align}
These conditions are satisfied, in particular, if
\([Q,W]=0\).

\end{enumerate}
\end{theorem}

\begin{proof}
By the Riesz representation of the spectral projector,
one has
\[
    P_\alpha(T)
    =
    \frac{1}{2\pi i}
    \oint_\Gamma(z\Id-W(T))^{-1}\,\dd z,
\]
and differentiation under the integral sign gives
\eqref{eq:riesz-derivative}, as in
\cref{quoted:dk-kato}.

To prove (ii), we use the spectral decomposition
\[
    (z\Id-W)^{-1}
    =
    \sum_{\lambda\in\sigma(W)}
    \frac{E_\lambda}{z-\lambda}.
\]
Substituting this expression into
\eqref{eq:riesz-derivative}, we obtain
\[
    \dot P_\alpha
    =
    \sum_{\lambda,\eta\in\sigma(W)}
    E_\eta Q E_\lambda
    \frac{1}{2\pi i}
    \oint_\Gamma
    \frac{\dd z}{(z-\eta)(z-\lambda)}.
\]

If both eigenvalues lie inside \(\Gamma\), or if
both lie outside it, the corresponding integral
vanishes. In particular, when \(\lambda=\eta\),
one has
\[
    \oint_\Gamma
    \frac{\dd z}{(z-\lambda)^2}=0.
\]
If \(\lambda>\alpha\) and \(\eta\leq\alpha\), the
residue theorem gives
\[
    \frac{1}{2\pi i}
    \oint_\Gamma
    \frac{\dd z}{(z-\eta)(z-\lambda)}
    =
    \frac{1}{\lambda-\eta}.
\]
Grouping the terms corresponding to the two possible
orders of \(\lambda\) and \(\eta\) gives
\eqref{eq:projector-block-transport}.

Since \(P_\alpha(T)\) is an orthogonal projector,
one has
\[
    \norm{P_\alpha(T)q}^2
    =
    \ip{q}{P_\alpha(T)q},
\]
and hence
\[
    \frac{\dd}{\dd T}\norm{P_\alpha(T)q}^2
    =
    \ip{q}{\dot P_\alpha(T)q}.
\]
Substituting \eqref{eq:projector-block-transport}
and using the self-adjointness of \(Q\) and of
the spectral projectors, we obtain
\eqref{eq:task-mass-block-transport}.

It remains to prove (iv). For \(\lambda>\alpha\)
and \(\eta\leq\alpha\), it follows from
\eqref{eq:projector-block-transport} that
\[
    E_\eta\dot P_\alpha E_\lambda
    =
    \frac{E_\eta Q E_\lambda}{\lambda-\eta}.
\]
Since \(\lambda-\eta>0\), the equality
\(\dot P_\alpha=0\) implies
\[
    E_\eta Q E_\lambda=0,
\]
for every such pair of eigenvalues. Summing over
\(\lambda\) and \(\eta\), we obtain
\[
    (\Id-P_\alpha)QP_\alpha
    =
    \sum_{\substack{\lambda>\alpha\\\eta\leq\alpha}}
    E_\eta Q E_\lambda
    =0.
\]

Conversely, suppose that
\((\Id-P_\alpha)QP_\alpha=0\). Multiplication on
the left by \(E_\eta\) and on the right by
\(E_\lambda\) gives
\[
    E_\eta Q E_\lambda=0,
\]
for every \(\lambda>\alpha\) and \(\eta\leq\alpha\).
Since \(Q\) is self-adjoint, one also has
\[
    E_\lambda Q E_\eta=0,
\]
and \eqref{eq:projector-block-transport} therefore
implies that \(\dot P_\alpha=0\).

Finally, by self-adjointness of \(Q\), the vanishing
of \((\Id-P_\alpha)QP_\alpha\) is equivalent to the
vanishing of both off-diagonal blocks of \(Q\)
relative to the decomposition
\[
    \HS=\ran P_\alpha\oplus\Ker P_\alpha.
\]
This is equivalent to \([Q,P_\alpha]=0\). If \(Q\)
commutes with \(W\), it commutes with all spectral
projectors of \(W\), and in particular with
\(P_\alpha\).
\end{proof}

When the eigenvalues are simple, the preceding
formulas may be expressed in terms of the corresponding
orthonormal eigenvectors.

\begin{corollary}
\label{cor:projector-eigenvectors}
Suppose that the eigenvalues \(\lambda_i(T)\) of
\(W(T)\) are simple, with corresponding orthonormal
eigenvectors \(u_i(T)\), and put
\[
    I_\alpha(T)=\{i:\lambda_i(T)>\alpha\}.
\]
Then
\begin{equation}
\label{eq:projector-eigen}
    \dot P_\alpha
    =
    \sum_{\substack{i\in I_\alpha\\j\notin I_\alpha}}
    \frac{\ip{u_j}{Q u_i}}
         {\lambda_i-\lambda_j}
    \left(u_j u_i^*+u_i u_j^*\right),
\end{equation}
and, for every fixed vector \(q\in\HS\),
\begin{equation}
\label{eq:task-mass-transport}
    \frac{\dd}{\dd T}\norm{P_\alpha(T)q}^2
    =
    2
    \sum_{\substack{i\in I_\alpha\\j\notin I_\alpha}}
    \frac{\ip{u_j}{Q u_i}}
         {\lambda_i-\lambda_j}
    \ip{q}{u_i}\ip{q}{u_j}.
\end{equation}
\end{corollary}

\begin{proof}
It suffices to substitute
\[
    E_{\lambda_i}=u_i u_i^*,
\]
into \eqref{eq:projector-block-transport} and
\eqref{eq:task-mass-block-transport}.
\end{proof}

\begin{remark}
The condition \([Q,W]=0\) is stronger than the
condition \([Q,P_\alpha]=0\). Indeed, \(Q\) may mix
distinct eigenspaces of \(W\) lying on the same side
of the threshold without changing the signal channel.

Relative to the orthogonal decomposition
\[
    \HS=\ran P_\alpha\oplus\Ker P_\alpha,
\]
write
\[
    W=
    \begin{pmatrix}
        W_+&0\\
        0&W_-
    \end{pmatrix},
    \qquad
    Q=
    \begin{pmatrix}
        Q_{++}&Q_{+-}\\
        Q_{-+}&Q_{--}
    \end{pmatrix}.
\]
By \cref{thm:projector-transport}, the spectral
projector is stationary at \(T\) if and only if
\[
    Q_{-+}=0,
\]
or, equivalently, \(Q_{+-}=0\). No commutation
condition between \(W_+\) and \(Q_{++}\), or between
\(W_-\) and \(Q_{--}\), is required.

Thus it is the off-diagonal part of the increment
\(Q\), relative to the signal channel and its
orthogonal complement, that governs the motion of
the spectral projector. The denominators in
\eqref{eq:projector-block-transport} describe the
dependence of this motion on the separation between
the two parts of the spectrum.

Note that stationarity of the projector implies
stationarity of the signal-channel mass of every
vector \(q\), whereas the converse for a single
vector \(q\) need not hold.
\end{remark}

\subsection{Activation and transport}

The preceding results give two different descriptions
of the variation of the signal channel, according to
whether or not its dimension changes.

\begin{corollary}
\label{cor:two-mechanisms}
Suppose that \(W(T)\) is continuously differentiable,
and let \(q\in\HS\) be fixed. The variation of
\[
    \norm{P_\alpha(T)q}^2,
\]
may be attributed to the following two phenomena.

\begin{enumerate}[label=(\roman*)]

\item \emph{Spectral activation}, when an eigenvalue
of \(W(T)\) passes from a value at most \(\alpha\)
to a value greater than \(\alpha\), thereby increasing
the dimension of the signal channel.

\item \emph{Spectral transport}, when the signal
channel varies while its dimension remains constant.
On intervals where the rank of \(P_\alpha(T)\) is
constant, this variation is governed by
\eqref{eq:projector-block-transport}, and its effect
on the vector \(q\) is given by
\eqref{eq:task-mass-block-transport}.

\end{enumerate}

If
\[
    [Q(T;s),P_\alpha(T)]=0,
\]
for almost every \(T\), the spectral projector is
constant on each interval on which its rank remains
constant. In this case, only spectral activation
can change the signal-channel mass of \(q\).

In particular, the same conclusion holds if
\[
    [Q(T;s),W(T)]=0,
\]
for almost every \(T\).
\end{corollary}

\begin{proof}
By \cref{prop:nested-reservoir}, the eigenvalues of
\(W(T)\) are nondecreasing, and consequently so is
\[
    \rank P_\alpha(T)
    =
    \#\{i:\lambda_i(T)>\alpha\},
\]
where the eigenvalues are counted with multiplicity.
A change in this rank can therefore occur only when
one or more eigenvalues pass above \(\alpha\).

On any open interval where the rank remains constant,
the spectral projector is locally given by a Riesz
integral with a contour separating the corresponding
spectral subspace from its complement. Its variation
is thus described by
\eqref{eq:projector-block-transport}.

If \([Q(T;s),P_\alpha(T)]=0\) almost everywhere,
it follows from assertion (iv) of
\cref{thm:projector-transport} that
\[
    \dot P_\alpha(T)=0,
\]
almost everywhere on each such interval. Since
\(P_\alpha(T)\) is continuously differentiable
there, it is constant on that interval.

The last assertion follows from the fact that an
operator commuting with \(W(T)\) also commutes
with its spectral projectors.
\end{proof}

\begin{remark}
An increase in the dimension of the signal channel
need not imply that a given vector acquires additional
signal-channel mass. Conversely, the mass of a vector
may increase without any change in the dimension of
the channel, through the motion of the corresponding
spectral subspace.

The latter phenomenon is specific to a positive
spectral threshold. For the exact signal channel,
the inclusion of subspaces established in
\cref{prop:nested-reservoir} implies that
\[
    T\longmapsto
    \norm{P_+(W(T))q}^2,
\]
is nondecreasing and is constant on every interval
where \(W(T)\) has constant rank.

Thus, whereas the exact signal channel can change
only through an increase in its dimension, an
effective signal channel may also change through
the motion of a subspace of fixed dimension.
\end{remark}

\subsection{Relation to evaluation visibility}

The spectral motion considered here is related to
the study of spectral transitions in update Gram
matrices~\cite{xu2026spectral}. The operator \(W(T)\),
however, is a cumulative dissipation operator in which
the instantaneous training kernels are transported
to a common initial time by the propagator \(\Pg\).

Neither spectral activation nor spectral transport
determines, by itself, the effect of a direction on
the evaluation outputs. This effect also depends on
the transfer operator \(G(T)\) and on the evaluation
law, as in the visibility operators of \cref{sec:task}.

The evolution of the corresponding filtered outputs
is described by \cref{thm:endpoint-balance}, while
their contribution to population risk is given by
\cref{thm:risk-balance}. Thus a change in the signal
channel, a change in evaluation visibility, and a
decrease in population risk remain distinct properties
of the training trajectory.

\clearpage

\section{Residual transport in a moving kernel basis}
\label{sec:moving-basis}

We study how the evolution of the instantaneous training
kernel affects the distribution of the residual among
its spectral modes. By expressing the residual in a
moving eigenbasis, we derive an evolution equation
that separates dissipation within individual modes
from the transfer of residual energy between them.
We show that this transfer is conservative and obtain
a quantitative bound in terms of the variation of
the kernel and the separation of its spectral blocks.
We also characterize the conditions under which
transport between spectral blocks vanishes. Finally,
we give an intrinsic formulation in terms of spectral
projectors, which remains valid for eigenvalues of
arbitrary multiplicity. These results distinguish
slow learning due to small kernel eigenvalues from
the redistribution of residual energy induced by
the evolution of the kernel.

While \cref{sec:spectral-transport} concerns the
spectral subspaces of the cumulative dissipation
operator \(W\), we shall now consider those of the
instantaneous training kernel. In contrast with the
constant-kernel case, where each spectral component
of the residual evolves independently, the motion
of the eigenbasis introduces a coupling between
these components.

Suppose that the training loss is given by
\[
    \Phi_S(u)=\frac12\norm{u-y}^2,
\]
and put \(r(t)=u(t)-y=g(t)\). By \eqref{eq:output-flow},
the residual satisfies
\[
    \dot r(t)=-K(t)r(t),
\]
where \(K(t)=K_{SS}^M(t)\).

Suppose that \(K(t)\) is continuously differentiable
and has simple eigenvalues on the interval under
consideration. Put \(N=\dim\HS=np\), and write
\[
    K(t)=U(t)\Lambda(t)U(t)^*,
\]
where
\[
    \Lambda(t)
    =\operatorname{diag}(\lambda_1(t),\ldots,\lambda_N(t)),
\]
and the columns \(u_i(t)\), \(1\leq i\leq N\), of
\(U(t)\) form a continuously differentiable orthonormal
eigenbasis of \(K(t)\).

We denote the coordinates of the residual in this
basis and the matrix describing its variation by
\[
    a(t)=U(t)^*r(t),
    \qquad
    \Omega(t)=U(t)^*\dot U(t),
\]
respectively. Differentiating the identity
\(U(t)^*U(t)=\Id\), we obtain
\[
    \Omega(t)^*=-\Omega(t),
\]
so that \(\Omega(t)\) is skew-adjoint.

\begin{theorem}
\label{thm:moving-basis}
With the notation introduced above, one has
\begin{align}
    \dot a(t)
    &=
    -\Lambda(t)a(t)-\Omega(t)a(t),
    \label{eq:moving-residual}\\
    \Omega_{ij}(t)
    &=
    \frac{\ip{u_i(t)}{\dot K(t)u_j(t)}}
         {\lambda_j(t)-\lambda_i(t)},
    \qquad i\ne j.
    \label{eq:connection}
\end{align}
In particular,
\begin{equation}
\label{eq:total-residual-dissipation}
    \frac{\dd}{\dd t}\norm{r(t)}^2
    =
        -2\sum_{i=1}^N\lambda_i(t)\abs{a_i(t)}^2.
\end{equation}

Let \(I\subseteq\{1,\ldots,N\}\) be a fixed set of
spectral indices and put \(J=I^c\). If \(a_I\) and
\(a_J\) denote the corresponding coordinate vectors,
then
\begin{equation}
\label{eq:block-flux}
    \frac{\dd}{\dd t}\norm{a_I(t)}^2
    =
    -2\sum_{i\in I}\lambda_i(t)\abs{a_i(t)}^2
    -
    2\operatorname{Re}
    \ip{a_I(t)}{\Omega_{IJ}(t)a_J(t)}.
\end{equation}
The first term on the right-hand side describes
the dissipation within the block \(I\), while the
second describes the transfer of residual energy
between the two blocks.
\end{theorem}

\begin{proof}
Since \(r=Ua\) and \(\dot r=-Kr\), one has
\[
\begin{aligned}
    \dot a
    &=
    \dot U^*r+U^*\dot r\\
    &=
    \dot U^*Ua-U^*KU a\\
    &=
    -\Omega a-\Lambda a,
\end{aligned}
\]
which proves \eqref{eq:moving-residual}.

Differentiating the eigenvalue equation
\[
    Ku_j=\lambda_j u_j,
\]
and taking the inner product with \(u_i\), for
\(i\ne j\), gives
\[
    \ip{u_i}{\dot K u_j}
    +
    \lambda_i\ip{u_i}{\dot u_j}
    =
    \lambda_j\ip{u_i}{\dot u_j},
\]
from which \eqref{eq:connection} follows.

Since \(U\) is orthogonal and \(\Omega\) is
skew-adjoint, one has
\[
    \norm r^2=\norm a^2,
    \qquad
    \operatorname{Re}\ip{a}{\Omega a}=0.
\]
Consequently,
\[
\begin{aligned}
    \frac{\dd}{\dd t}\norm r^2
    &=
    2\operatorname{Re}\ip{a}{\dot a}\\
    &=
    -2\ip{a}{\Lambda a}\\
    &=
    -2\sum_i\lambda_i\abs{a_i}^2,
\end{aligned}
\]
which gives \eqref{eq:total-residual-dissipation}.

To prove the last assertion, write
\[
    \dot a_I
    =
    -\Lambda_Ia_I-\Omega_{II}a_I
    -\Omega_{IJ}a_J.
\]
Since \(\Omega_{II}\) is skew-adjoint, its contribution
to the derivative of \(\norm{a_I}^2\) vanishes, and
we obtain
\[
\begin{aligned}
    \frac{\dd}{\dd t}\norm{a_I}^2
    &=
    2\operatorname{Re}\ip{a_I}{\dot a_I}\\
    &=
    -2\ip{a_I}{\Lambda_Ia_I}
    -2\operatorname{Re}\ip{a_I}{\Omega_{IJ}a_J},
\end{aligned}
\]
as asserted.
\end{proof}

\begin{remark}
The transport term in \eqref{eq:block-flux} is
conservative. Indeed, since
\[
    \Omega_{JI}=-\Omega_{IJ}^*,
\]
one has
\[
    \operatorname{Re}\ip{a_I}{\Omega_{IJ}a_J}
    +
    \operatorname{Re}\ip{a_J}{\Omega_{JI}a_I}
    =0.
\]
Thus the contribution of the motion of the eigenbasis
to the energy of one block is the negative of its
contribution to the energy of the other. The transport
terms disappear when the two block identities are
added, leaving only the dissipation term in
\eqref{eq:total-residual-dissipation}.

In particular, the residual energy of an individual
spectral block may increase even though the total
residual energy is nonincreasing.
\end{remark}

The magnitude of the transport term depends on the
off-diagonal coefficients of \(\dot K\) and on the
separation between the spectra of the two blocks.
The following estimate makes this dependence explicit.

\begin{corollary}
\label{cor:gap-flux}
Let \(I\) and \(J=I^c\) be nonempty sets of spectral
indices, and suppose that the corresponding spectra
are separated by a positive gap
\[
    \Delta_I(t)
    =
    \min_{\substack{i\in I\\j\in J}}
    \abs{\lambda_i(t)-\lambda_j(t)}
    >0.
\]
Denote the corresponding orthogonal projectors by
\[
    P_I=\sum_{i\in I}u_i u_i^*,
    \qquad
    P_J=\Id-P_I.
\]
Then
\begin{equation}
\label{eq:flux-bound}
    \abs{
        2\operatorname{Re}\ip{a_I}{\Omega_{IJ}a_J}
    }
    \leq
    \frac{2\norm{P_I\dot K P_J}_{\Frob}}{\Delta_I}
    \norm{a_I}\norm{a_J}.
\end{equation}

Moreover, the following conditions are equivalent,
\[
    \Omega_{IJ}=0,
    \qquad
    P_I\dot K P_J=0,
    \qquad
    [\dot K,P_I]=0.
\]
They are satisfied, in particular, if
\([K,\dot K]=0\).
\end{corollary}

\begin{proof}
By \eqref{eq:connection}, one has
\[
\begin{aligned}
    \norm{\Omega_{IJ}}_{\Frob}^2
    &=
    \sum_{\substack{i\in I\\j\in J}}
    \frac{\abs{\ip{u_i}{\dot K u_j}}^2}
         {\abs{\lambda_j-\lambda_i}^2}\\
    &\leq
    \frac{1}{\Delta_I^2}
    \sum_{\substack{i\in I\\j\in J}}
    \abs{\ip{u_i}{\dot K u_j}}^2\\
    &=
    \frac{\norm{P_I\dot K P_J}_{\Frob}^2}
         {\Delta_I^2}.
\end{aligned}
\]
Since the operator norm is bounded by the Frobenius
norm, it follows that
\[
    \norm{\Omega_{IJ}}_{\op}
    \leq
    \frac{\norm{P_I\dot K P_J}_{\Frob}}{\Delta_I},
\]
and \eqref{eq:flux-bound} follows from the
Cauchy--Schwarz inequality.

The equivalence between the first two conditions
follows from \eqref{eq:connection}, since the
denominators are nonzero. Their equivalence with
the third follows from the self-adjointness of
\(\dot K\). Finally, if \(\dot K\) commutes with
\(K\), it commutes with its spectral projectors,
which proves the last assertion.
\end{proof}

\subsection{An intrinsic formulation}

The preceding identities may also be expressed without
choosing an eigenbasis. This formulation is useful when
the eigenvalues within a spectral block are not simple.

Let \(P(t)\) be the spectral projector of \(K(t)\)
associated with a cluster separated from the remainder
of its spectrum. Since \(P(t)\) commutes with \(K(t)\),
one has
\[
    P(t)K(t)=K(t)P(t).
\]
The residual energy associated with this cluster
satisfies
\begin{equation}
\label{eq:intrinsic-residual-flux}
    \frac{\dd}{\dd t}\norm{P(t)r(t)}^2
    =
    -2\ip{P(t)r(t)}{K(t)P(t)r(t)}
    +
    \ip{r(t)}{\dot P(t)r(t)}.
\end{equation}
Indeed, differentiating the identity
\[
    \norm{P(t)r(t)}^2=\ip{r(t)}{P(t)r(t)},
\]
and using \(\dot r=-Kr\) gives
\eqref{eq:intrinsic-residual-flux}.

The last term in \eqref{eq:intrinsic-residual-flux}
describes the transfer of residual energy due to the
variation of the spectral subspace. For the complementary
projector \(P^\perp=\Id-P\), the corresponding term is
\[
    \ip{r}{\dot P^\perp r}
    =
    -\ip{r}{\dot P r},
\]
which gives another proof of the conservation property
established above.

When \(P=P_I\), comparison of
\eqref{eq:block-flux} and
\eqref{eq:intrinsic-residual-flux} gives
\[
    \ip{r}{\dot P_I r}
    =
    -2\operatorname{Re}\ip{a_I}{\Omega_{IJ}a_J}.
\]
Thus the transport term in the moving-basis equation
is precisely the contribution of the derivative of
the spectral projector to the residual energy.

Moreover, the spectral projector formula of
\cref{thm:projector-transport}, applied to \(K\)
rather than to \(W\), expresses \(\dot P\) in terms
of \(\dot K\) and the gaps between the two spectral
blocks. In particular, the intrinsic identity
\eqref{eq:intrinsic-residual-flux} remains valid
for eigenvalues of arbitrary multiplicity.

\begin{remark}[Dissipation and transport]
\label{rem:two-delays}
When the kernel is constant, one may choose a fixed
orthonormal eigenbasis, in which case \(\Omega=0\)
and the residual coordinates satisfy
\[
    a_i(t)=e^{-(t-s)\lambda_i}a_i(s).
\]
The decay of each component is then governed entirely
by the corresponding eigenvalue. Small positive
eigenvalues give rise to long dissipation times,
as in the model of \cref{thm:two-mode}.

When the kernel varies, there is an additional
mechanism, namely the transfer of residual energy
between spectral subspaces. A component initially
associated with small eigenvalues may be transferred
to a subspace on which dissipation is stronger,
or conversely.

The estimate \eqref{eq:flux-bound} shows that small
spectral gaps permit large transport rates, provided
that the corresponding off-diagonal coefficients of
\(\dot K\) do not vanish. Neither the magnitude nor
the sign of the transport term is determined by the
spectral gap alone.

As in \cref{sec:spectral-transport}, the resulting
change in the training residual must be distinguished
from its effect on evaluation outputs and from its
contribution to the decrease of population risk.
\end{remark}

\clearpage

\section{An exact spectral-delay model}
\label{sec:two-mode}

We study the separation between the time required to
achieve low training loss and that required to achieve
low population risk. We construct an explicit model
with two spectral modes for which these times can be
estimated directly, and prove that the delay between
them can be arbitrarily large even when the features
of the model remain fixed. We also extend this
construction to several spectral modes and obtain
explicit expressions for the times at which different
components of a task reach prescribed levels of
population risk. These results show that delayed
generalization and sequential learning can arise
solely from differences in spectral learning rates
and their contributions to evaluation outputs.

In the example that follows, the training kernel
is constant. The separation between the training
and evaluation times results from the difference
between its two eigenvalues and from the dependence
of the evaluation output on the corresponding
spectral directions. In particular, neither the
activation of new spectral directions nor the
rotation of the kernel eigenspaces is required
to produce such a delay.

Let \(\HP=\HS=\R^2\), and consider the linear output map
\[
    U_S(w)=J_Sw,
    \qquad
    J_S=
    \begin{pmatrix}
        \sqrt{\kappa_f}&0\\
        0&\sqrt{\kappa_s}
    \end{pmatrix},
\]
where \(0<\kappa_s<\kappa_f\). The training loss is
given by
\[
    \Phi_S(U)=\frac12\norm{U-y}^2,
    \qquad
    y=(a,b)^\top.
\]
We take \(w(0)=0\) and \(M=\Id\), so that the training
kernel is constant and equal to
\[
    K_{SS}
    =
    \begin{pmatrix}
        \kappa_f&0\\
        0&\kappa_s
    \end{pmatrix}.
\]

We take the evaluation space to be \(\R\), with its
usual norm, and define the evaluation output and its
target by
\[
    U_E(w)=j_Ew,
    \qquad
    j_E=(0,c\sqrt{\kappa_s}),
    \qquad
    f^\star=cb.
\]
Thus the evaluation output depends only on the
component associated with the smaller eigenvalue
of the training kernel. We put
\[
    \Risk_E(t)
    =
    \frac12\abs{U_E(w(t))-f^\star}^2.
\]

\begin{theorem}[Spectral delay]
\label{thm:two-mode}
For every \(t\geq0\), one has
\begin{align}
    \Loss_S(t)
    &=
    \frac12\left(
        a^2e^{-2\kappa_f t}
        +
        b^2e^{-2\kappa_s t}
    \right),
    \label{eq:two-mode-train}\\
    \Risk_E(t)
    &=
    \frac12c^2b^2e^{-2\kappa_s t}.
    \label{eq:two-mode-risk}
\end{align}

Let \(\eps\) and \(\delta\) satisfy
\[
    b^2<2\eps<a^2+b^2,
    \qquad
    0<\delta<\frac12c^2b^2,
\]
and denote the corresponding first passage times by
\[
    t_{\mathrm{mem}}(\eps)
    =
    \inf\{t\geq0:\Loss_S(t)\leq\eps\},
\]
and
\[
    t_{\mathrm{grok}}(\delta)
    =
    \inf\{t\geq0:\Risk_E(t)\leq\delta\}.
\]
Then
\begin{align}
    t_{\mathrm{mem}}(\eps)
    &\leq
    \frac{1}{2\kappa_f}
    \log\left(
        \frac{a^2}{2\eps-b^2}
    \right),
    \label{eq:mem-upper}\\
    t_{\mathrm{grok}}(\delta)
    &=
    \frac{1}{2\kappa_s}
    \log\left(
        \frac{c^2b^2}{2\delta}
    \right).
    \label{eq:grok-exact}
\end{align}
Consequently,
\begin{equation}
\label{eq:delay-lower}
    t_{\mathrm{grok}}(\delta)
    -
    t_{\mathrm{mem}}(\eps)
    \geq
    \frac{1}{2\kappa_s}
    \log\left(
        \frac{c^2b^2}{2\delta}
    \right)
    -
    \frac{1}{2\kappa_f}
    \log\left(
        \frac{a^2}{2\eps-b^2}
    \right).
\end{equation}
If \(a,b,c,\kappa_f,\eps,\delta\) are fixed, the
right-hand side tends to \(+\infty\) as
\(\kappa_s\) tends to zero.
\end{theorem}

\begin{proof}
The training residual
\[
    r(t)=U_S(w(t))-y,
\]
satisfies
\[
    \dot r(t)=-K_{SS}r(t),
    \qquad
    r(0)=(-a,-b)^\top.
\]
Since \(K_{SS}\) is diagonal, one obtains
\[
    r(t)
    =
    \begin{pmatrix}
        -ae^{-\kappa_ft}\\
        -be^{-\kappa_st}
    \end{pmatrix},
\]
from which \eqref{eq:two-mode-train} follows.

Moreover, since \(U_S(w(t))=y+r(t)\), one has
\[
    w(t)
    =
    \begin{pmatrix}
        \displaystyle
        \frac{a}{\sqrt{\kappa_f}}
        \bigl(1-e^{-\kappa_ft}\bigr)\\[6pt]
        \displaystyle
        \frac{b}{\sqrt{\kappa_s}}
        \bigl(1-e^{-\kappa_st}\bigr)
    \end{pmatrix}.
\]
It follows from the definition of \(j_E\) that
\[
    U_E(w(t))
    =
    cb\bigl(1-e^{-\kappa_st}\bigr),
\]
and hence
\[
    U_E(w(t))-f^\star
    =
    -cb\,e^{-\kappa_st}.
\]
This proves \eqref{eq:two-mode-risk}.

To estimate the first passage time for the training
loss, put
\[
    t_f
    =
    \frac{1}{2\kappa_f}
    \log\left(
        \frac{a^2}{2\eps-b^2}
    \right).
\]
The hypotheses on \(\eps\) imply that \(t_f>0\).
At this time, one has
\[
    a^2e^{-2\kappa_ft_f}=2\eps-b^2,
\]
and, since \(e^{-2\kappa_st_f}\leq1\),
\[
\begin{aligned}
    2\Loss_S(t_f)
    &=
    2\eps-b^2+b^2e^{-2\kappa_st_f}\\
    &\leq2\eps.
\end{aligned}
\]
Thus \(t_{\mathrm{mem}}(\eps)\leq t_f\), which proves
\eqref{eq:mem-upper}.

On the other hand, by \eqref{eq:two-mode-risk},
the inequality \(\Risk_E(t)\leq\delta\) is equivalent
to
\[
    e^{-2\kappa_st}
    \leq
    \frac{2\delta}{c^2b^2},
\]
or
\[
    t\geq
    \frac{1}{2\kappa_s}
    \log\left(
        \frac{c^2b^2}{2\delta}
    \right).
\]
This gives \eqref{eq:grok-exact}, and subtraction
of the upper bound for \(t_{\mathrm{mem}}(\eps)\)
gives \eqref{eq:delay-lower}.

Finally, the hypothesis on \(\delta\) implies that
\[
    \log\left(
        \frac{c^2b^2}{2\delta}
    \right)>0,
\]
while the second term on the right-hand side of
\eqref{eq:delay-lower} is independent of \(\kappa_s\).
The last assertion follows.
\end{proof}

\begin{remark}
One can describe the limiting behaviour of the two
passage times more precisely. Since the training loss
is strictly decreasing, \(t_{\mathrm{mem}}(\eps)\)
is the unique solution of
\[
    a^2e^{-2\kappa_f t}
    +
    b^2e^{-2\kappa_s t}
    =
    2\eps.
\]
As \(\kappa_s\) tends to zero, the left-hand side
converges uniformly on compact intervals to
\[
    a^2e^{-2\kappa_f t}+b^2,
\]
whose value is \(2\eps\) precisely at \(t=t_f\).
It follows that
\[
    \lim_{\kappa_s\downarrow0}
    t_{\mathrm{mem}}(\eps)
    =
    \frac{1}{2\kappa_f}
    \log\left(
        \frac{a^2}{2\eps-b^2}
    \right).
\]
Consequently,
\[
\begin{aligned}
    t_{\mathrm{grok}}(\delta)
    -
    t_{\mathrm{mem}}(\eps)
    &=
    \frac{1}{2\kappa_s}
    \log\left(
        \frac{c^2b^2}{2\delta}
    \right)\\
    &\quad-
    \frac{1}{2\kappa_f}
    \log\left(
        \frac{a^2}{2\eps-b^2}
    \right)
    +o(1),
\end{aligned}
\]
as \(\kappa_s\downarrow0\). Thus the delay is
asymptotically proportional to \(1/\kappa_s\).
\end{remark}

\begin{remark}
The hypothesis \(b^2<2\eps\) allows the training
loss to reach the prescribed tolerance before the
second component of the residual has decreased
substantially. The evaluation risk, however, depends
entirely on this component, and its decrease therefore
takes place on the time scale \(1/\kappa_s\).

The example shows that an arbitrarily large separation
between training and evaluation times may occur even
when the kernel is constant. The motion of the spectral
subspaces studied in \cref{sec:spectral-transport,sec:moving-basis}
constitutes a different mechanism, which is absent
from the present example.

It should be noted that \cref{thm:two-mode} concerns
squared population risk. The attainment of the training
tolerance does not imply exact interpolation, nor does
the separation of the two passage times by itself
imply a rapid increase in classification accuracy.
In \cref{sec:abrupt}, we retain the same training
dynamics and construct an evaluation map for which
an arbitrarily long delay is accompanied by a
concentration of the classification margin crossing
times on an interval of vanishing length.
\end{remark}

\subsection{Several spectral scales}

The preceding construction extends to an arbitrary
finite number of orthogonal modes. Each mode has its
own dissipation time, and its contribution to the
evaluation risk depends also on its amplitude in
the evaluation space.

\begin{corollary}
\label{cor:sequential}
Let \(\kappa_1,\ldots,\kappa_m>0\), and consider
\(m\) independent training modes with target
coefficients \(b_1,\ldots,b_m\). Suppose that their
evaluation outputs are mutually orthogonal, with
target amplitudes \(c_1b_1,\ldots,c_mb_m\).

Then the population risk is given by
\[
    \Risk_E(t)
    =
    \frac12
    \sum_{j=1}^m
    c_j^2b_j^2e^{-2\kappa_jt}.
\]
For each \(j\), let \(\delta_j\) satisfy
\[
    0<\delta_j<\frac12c_j^2b_j^2.
\]
The first time \(t_j\) at which the contribution
of the \(j\)-th mode to the population risk is at
most \(\delta_j\) is
\begin{equation}
\label{eq:subrule-time}
    t_j
    =
    \frac{1}{2\kappa_j}
    \log\left(
        \frac{c_j^2b_j^2}{2\delta_j}
    \right).
\end{equation}
\end{corollary}

\begin{proof}
By the calculation in \cref{thm:two-mode}, the
evaluation residual associated with the \(j\)-th
mode has amplitude
\[
    -c_jb_j e^{-\kappa_jt}.
\]
Since these residuals are mutually orthogonal in
the evaluation space, their squared norms add.
Consequently,
\[
    \Risk_E(t)
    =
    \sum_{j=1}^m\Risk_{E,j}(t),
    \qquad
    \Risk_{E,j}(t)
    =
    \frac12c_j^2b_j^2e^{-2\kappa_jt}.
\]
Solving the inequality
\(\Risk_{E,j}(t)\leq\delta_j\) gives
\eqref{eq:subrule-time}.
\end{proof}

\begin{remark}
The order in which the individual components reach
their prescribed tolerances is not determined by
the order of the eigenvalues \(\kappa_j\) alone,
but by the quantities in \eqref{eq:subrule-time}.

If, for instance, the tolerances are chosen to be
a fixed fraction of the respective initial risks,
so that
\[
    \delta_j
    =
    \frac{\eta}{2}c_j^2b_j^2,
    \qquad
    0<\eta<1,
\]
then
\[
    t_j
    =
    \frac{\log(1/\eta)}{2\kappa_j},
\]
and the order of the passage times is the reverse
of that of the eigenvalues.

For general tolerances, the evaluation amplitudes
also enter into the determination of these times.
Thus even in the absence of spectral transport,
different orthogonal components of a task may
reach their respective evaluation tolerances at
widely separated times.
\end{remark}

\clearpage

\section{Margin crossings and the emergence of generalization}
\label{sec:abrupt}

We study how a smooth evolution of the model outputs
can give rise to a rapid increase in classification
accuracy. We first relate accuracy to the distribution
of the times at which individual classification margins
change sign. We then extend the spectral-delay model of
\cref{sec:two-mode} to a binary classification problem.
In this model, we prove that the time between achieving
low training loss and high evaluation accuracy can be
arbitrarily large, while the interval over which
evaluation accuracy increases from zero to one tends
to zero. Both phenomena arise from the spectral
properties of a gradient flow whose features do not
evolve during training.

\subsection{Classification accuracy and crossing times}

The example of \cref{sec:two-mode} establishes a
separation between the times required to achieve
low training loss and low population risk. However,
the population risk in that example decreases
smoothly from the beginning of training. To understand
how such dynamics may give rise to a rapid improvement
in classification accuracy, we must distinguish the
evolution of the model outputs from that of the
associated classification rule.

In a binary classification problem, correct
classification depends on the sign of the signed
margin rather than on its magnitude. Consequently,
a gradual evolution of the margins may produce
a rapid increase in accuracy when a large proportion
of the evaluation examples cross the decision boundary
within a short time interval.

\begin{proposition}
\label{prop:accuracy-cdf}
Consider a binary classification problem in which
the signed margin at \(z\) is of the form
\begin{equation}
\label{eq:margin-mode}
    m(t,z)
    =
    m_\infty(z)-d(z)e^{-\kappa t},
\end{equation}
where \(\kappa>0\), and \(m_\infty\) and \(d\) are
positive measurable functions. We assume that ties
are resolved by a fixed measurable rule.

For every \(z\), put
\[
    \tau(z)
    =
    \max\left\{
        0,\,
        \frac{1}{\kappa}
        \log\left(\frac{d(z)}{m_\infty(z)}\right)
    \right\}.
\]
At every time \(t\geq0\) for which the set of examples
with zero margin has measure zero under \(\mu\), the
evaluation accuracy is given by
\begin{equation}
\label{eq:accuracy-cdf}
    A_\mu(t)
    =
    \mu\{z:\tau(z)\leq t\}.
\end{equation}
More generally, irrespective of the convention adopted
for zero margins, one has
\[
    A_\mu(t+)
    =
    \mu\{z:\tau(z)\leq t\},
\]
where \(A_\mu(t+)\) denotes the right-hand limit of
the accuracy at \(t\).

Let \(t_0\geq0\), \(w>0\), and \(0\leq\eta<1/2\).
If
\[
    \mu\{\tau<t_0\}\leq\eta,
    \qquad
    \mu\{\tau>t_0+w\}\leq\eta,
\]
then
\[
    A_\mu(t)\leq\eta
    \qquad (0\leq t<t_0),
\]
and
\[
    A_\mu(t)\geq1-\eta
    \qquad (t>t_0+w).
\]
\end{proposition}

\begin{proof}
For every \(z\), the function \(t\mapsto m(t,z)\)
is strictly increasing, since
\[
    \frac{\partial m}{\partial t}(t,z)
    =
    \kappa d(z)e^{-\kappa t}>0.
\]
If \(d(z)>m_\infty(z)\), its unique zero on
\([0,\infty)\) occurs at
\[
    t=
    \frac{1}{\kappa}
    \log\left(\frac{d(z)}{m_\infty(z)}\right).
\]
If \(d(z)<m_\infty(z)\), the margin is positive
for every \(t\geq0\), whereas if
\(d(z)=m_\infty(z)\), it vanishes at \(t=0\)
and is positive thereafter.

It follows that, for every \(t\geq0\),
\[
    \mu\{\tau<t\}
    \leq A_\mu(t)
    \leq \mu\{\tau\leq t\}.
\]
When the set of examples with zero margin at time
\(t\) has measure zero, correct classification is
equivalent almost everywhere to \(\tau(z)\leq t\).
This proves \eqref{eq:accuracy-cdf}.

For every \(z\) and all sufficiently small \(h>0\),
the indicator of correct classification at time
\(t+h\) agrees with
\[
    \1_{\{\tau(z)\leq t\}}.
\]
The assertion concerning the right-hand limit
therefore follows from dominated convergence.

Finally, if \(t<t_0\), then
\[
    A_\mu(t)
    \leq
    \mu\{\tau\leq t\}
    \leq
    \mu\{\tau<t_0\}
    \leq\eta,
\]
whereas if \(t>t_0+w\), then
\[
    A_\mu(t)
    \geq
    \mu\{\tau<t\}
    \geq
    \mu\{\tau\leq t_0+w\}
    \geq1-\eta.
\]
This proves the remaining assertions.
\end{proof}

\begin{remark}
Let \(\nu_\mu=\tau_*\mu\) be the image measure of
\(\mu\) under \(\tau\). By the preceding proposition,
the right-continuous version of the accuracy is the
distribution function of \(\nu_\mu\), namely
\[
    A_\mu(t+)=\nu_\mu([0,t]).
\]
For \(t>0\), one has
\[
    A_\mu(t+)-A_\mu(t-)
    =
    \nu_\mu(\{t\}).
\]
Thus a discontinuity of accuracy occurs precisely
when a set of examples of positive measure has the
same crossing time.

If \(\nu_\mu\) is absolutely continuous, with density
\(p_\mu\), then
\[
    \frac{\dd}{\dd t}A_\mu(t)=p_\mu(t),
\]
for almost every \(t>0\). In this case, a rapid increase
in accuracy corresponds to a concentration of the
crossing-time distribution on a short interval,
rather than to a discontinuity.
\end{remark}

\begin{remark}
For examples with \(d(z)>m_\infty(z)\), the crossing
time is determined by the ratio
\[
    \frac{d(z)}{m_\infty(z)}.
\]
Suppose that there exist constants \(1<r_-\leq r_+\)
such that
\[
    \mu\left\{
        z:
        r_-\leq
        \frac{d(z)}{m_\infty(z)}
        \leq r_+
    \right\}
    \geq1-2\eta.
\]
Then a set of examples of measure at least \(1-2\eta\)
has crossing times in the interval
\[
    \left[
        \frac{\log r_-}{\kappa},
        \frac{\log r_+}{\kappa}
    \right],
\]
whose length is
\[
    w=
    \frac{1}{\kappa}
    \log\left(\frac{r_+}{r_-}\right).
\]
The location of this interval and its length depend
on different properties of the crossing-time
distribution. A late emergence of generalization
may therefore occur over a short interval if the
ratios \(d(z)/m_\infty(z)\) are sufficiently close
to one another.
\end{remark}

\subsection{A two-mode classification model}

The preceding proposition describes the relation
between classification accuracy and the distribution
of margin crossing times. It does not, however,
determine how such a distribution arises from the
training dynamics. We now construct an example in
which the concentration of crossing times follows
directly from the spectral structure of a gradient
flow.

We retain the training problem of \cref{sec:two-mode},
with \(a,b>0\), and write
\[
    U_S(w)
    =
    \begin{pmatrix}
        u_1(w)\\
        u_2(w)
    \end{pmatrix}
    =
    \begin{pmatrix}
        \sqrt{\kappa_f}\,w_1\\
        \sqrt{\kappa_s}\,w_2
    \end{pmatrix},
\]
where \(0<\kappa_s<\kappa_f\). The training loss
and the initialization are given by
\[
    \Loss_S(w)
    =
    \frac12\left(
        (u_1(w)-a)^2+(u_2(w)-b)^2
    \right),
    \qquad
    w(0)=0.
\]
Under gradient flow, the training outputs satisfy
\[
    u_1(w(t))
    =
    a(1-e^{-\kappa_f t}),
    \qquad
    u_2(w(t))
    =
    b(1-e^{-\kappa_s t}).
\]

We now replace the evaluation map of
\cref{sec:two-mode} by a scalar classification score.
Let
\[
    \mathcal Z_E=[-2,-1]\cup[1,2],
\]
and let \(\mu\) be the normalized Lebesgue measure
on \(\mathcal Z_E\). The true class of an evaluation
example \(z\) is given by
\[
    y(z)=\operatorname{sgn}(z).
\]
We define the evaluation score by
\begin{equation}
\label{eq:two-mode-classification-readout}
    F(w,z)
    =
    z\left(\frac{2u_2(w)}{b}-1\right)
    +
    \frac14\left(1-\frac{u_1(w)}{a}\right).
\end{equation}
The predicted class is determined by the sign of
\(F(w,z)\), with a fixed measurable convention
for zero scores.

The training and evaluation inputs may be regarded
as disjoint subsets of a common input space. On the
two training inputs, the model is defined by the
outputs \(u_1(w)\) and \(u_2(w)\), respectively,
and on \(\mathcal Z_E\) it is defined by
\eqref{eq:two-mode-classification-readout}.
The resulting model is affine in the parameters,
and its Jacobians are constant along each training
trajectory.

Substituting the expressions for the training outputs
into \eqref{eq:two-mode-classification-readout}, we
obtain
\begin{equation}
\label{eq:two-mode-classification-score}
    F(w(t),z)
    =
    z(1-2e^{-\kappa_s t})
    +
    \frac14e^{-\kappa_f t}.
\end{equation}
Consequently, the signed classification margin is
given by
\begin{equation}
\label{eq:two-mode-classification-margin}
    m(t,z)
    =
    |z|(1-2e^{-\kappa_s t})
    +
    \frac{\operatorname{sgn}(z)}{4}
    e^{-\kappa_f t}.
\end{equation}

The two spectral modes play different roles in this
expression. The slow mode determines the eventual
change of sign of the margins, while the fast mode
introduces a difference between the crossing times
of individual examples. At the time
\[
    t_\star=\frac{\log 2}{\kappa_s},
\]
the contribution of the slow mode to the score
vanishes, whereas the contribution of the fast
mode is exponentially small in
\(\kappa_f/\kappa_s\). This observation leads to
the following result.

\begin{theorem}[Delayed generalization with concentrated
crossing times]
\label{thm:two-mode-classification}
Consider the classification model defined by
\eqref{eq:two-mode-classification-readout}.
Fix \(a,b,\kappa_f>0\) and a training-loss tolerance
\(\varepsilon\) satisfying
\[
    b^2<2\varepsilon<a^2+b^2.
\]
For \(0<\kappa_s<\kappa_f\), put
\[
    t_\star=\frac{\log 2}{\kappa_s},
    \qquad
    q_s=\frac14e^{-\kappa_f t_\star/2},
\]
and define
\begin{equation}
\label{eq:two-mode-crossing-interval}
    t_-=
    \frac{1}{\kappa_s}
    \log\left(\frac{2}{1+q_s}\right),
    \qquad
    t_+=
    \frac{1}{\kappa_s}
    \log\left(\frac{2}{1-q_s}\right).
\end{equation}
Then the following assertions hold.

\begin{enumerate}[label=(\roman*)]

\item The training-loss passage time satisfies
\[
    t_{\rm mem}(\varepsilon)
    \leq
    \frac{1}{2\kappa_f}
    \log\left(\frac{a^2}{2\varepsilon-b^2}\right).
\]
In particular, it remains bounded as
\(\kappa_s\downarrow0\).

\item Every evaluation example has a unique margin
crossing time in \([t_-,t_+]\). The evaluation
accuracy is continuous and nondecreasing and
satisfies
\[
    A_\mu(t)=0
    \qquad (0\leq t\leq t_-),
\]
and
\[
    A_\mu(t)=1
    \qquad (t\geq t_+).
\]

\item The length of the interval containing all
margin crossing times satisfies
\begin{equation}
\label{eq:two-mode-crossing-width}
    w_s:=t_+-t_-
    =
    \frac{1}{\kappa_s}
    \log\left(\frac{1+q_s}{1-q_s}\right),
\end{equation}
and
\[
    w_s
    \sim
    \frac{1}{2\kappa_s}
    \exp\left(
        -\frac{\kappa_f\log 2}{2\kappa_s}
    \right)
    \longrightarrow0,
\]
as \(\kappa_s\downarrow0\).

\item For every fixed accuracy tolerance
\(0<\delta<1\), the classification grokking time
of \cref{def:grokking} satisfies
\begin{equation}
\label{eq:two-mode-classification-grokking-time}
    t_{\rm grok}^{\mu}(\varepsilon,\delta)
    =
    \frac{\log 2}{\kappa_s}+o(1),
\end{equation}
as \(\kappa_s\downarrow0\). In particular,
\[
    t_{\rm grok}^{\mu}(\varepsilon,\delta)
    -
    t_{\rm mem}(\varepsilon)
    \longrightarrow+\infty.
\]

\end{enumerate}
\end{theorem}

\begin{proof}
The first assertion follows from
\cref{thm:two-mode}, since the training problem
has not been modified.

To prove the second assertion, put
\[
    v(t)=1-2e^{-\kappa_s t},
    \qquad
    q(t)=\frac14e^{-\kappa_f t}.
\]
The signed margin is then given by
\[
    m(t,z)
    =
    |z|v(t)+\operatorname{sgn}(z)q(t).
\]

For \(0\leq t\leq t_\star/2\), one has
\[
    v(t)
    \leq
    1-\sqrt{2}.
\]
Since \(|z|\geq1\) and \(q(t)\leq1/4\), it follows
that
\[
    m(t,z)
    \leq
    1-\sqrt{2}+\frac14<0.
\]
Thus every evaluation example is incorrectly
classified throughout this interval.

For \(t\geq t_\star/2\), one has
\[
    0<q(t)\leq q_s.
\]
By the definitions of \(t_-\) and \(t_+\),
\[
    v(t_-)=-q_s,
    \qquad
    v(t_+)=q_s.
\]
Moreover, since
\[
    q_s<\frac14<\sqrt{2}-1,
\]
the definitions of \(t_-\) and \(t_+\) give
\[
    \frac{t_\star}{2}<t_-<t_\star<t_+.
\]

The function \(v\) is strictly increasing.
For \(t_\star/2\leq t<t_-\), we therefore obtain
\[
    m(t,z)
    \leq
    v(t)+q_s<0.
\]
Similarly, for \(t>t_+\), one has
\[
    m(t,z)
    \geq
    v(t)-q_s>0.
\]
Consequently,
\[
    A_\mu(t)=0
    \qquad (0\leq t<t_-),
\]
and
\[
    A_\mu(t)=1
    \qquad (t>t_+).
\]

We next establish the uniqueness of the margin
crossing times. If \(z<0\), then
\[
    m(t,z)
    =
    |z|(1-2e^{-\kappa_s t})
    -
    \frac14e^{-\kappa_f t},
\]
and hence
\[
    \frac{\partial m}{\partial t}(t,z)
    =
    2\kappa_s|z|e^{-\kappa_s t}
    +
    \frac{\kappa_f}{4}e^{-\kappa_f t}>0.
\]
The margin therefore has a unique zero, which
lies in \([t_-,t_+]\).

Suppose now that \(z>0\). For \(t\geq t_\star\),
the margin is strictly positive. On
\([0,t_\star]\), the equation \(m(t,z)=0\)
is equivalent to
\[
    h(t)=\frac{1}{4z},
\]
where
\[
    h(t)
    =
    e^{\kappa_f t}(2e^{-\kappa_s t}-1).
\]
One has
\[
    h(0)=1,
    \qquad
    h(t_\star)=0,
\]
and
\[
    h'(t)
    =
    e^{\kappa_f t}
    \left(
        2(\kappa_f-\kappa_s)e^{-\kappa_s t}
        -
        \kappa_f
    \right).
\]
The expression in parentheses is strictly decreasing.
Thus \(h\) is either strictly decreasing throughout
the interval, or initially increasing and subsequently
strictly decreasing. Since
\[
    0<\frac{1}{4z}\leq\frac14<1,
\]
the equation has exactly one solution. The margin
is negative before this solution and positive
thereafter.

It follows that every evaluation example has a
unique crossing time \(\tau(z)\), with
\[
    t_-\leq\tau(z)\leq t_+.
\]
At each fixed time \(t\), the score
\eqref{eq:two-mode-classification-score} is an
affine function of \(z\). If \(t\ne t_\star\),
it has at most one zero, whereas if \(t=t_\star\),
it is a strictly positive constant. Since \(\mu\)
has no atoms, the set of examples with zero margin
has measure zero at every time.

Consequently,
\[
    A_\mu(t)=\mu\{z:\tau(z)\leq t\}.
\]
The crossing-time distribution has no atoms, since
\[
    \mu\{\tau=t\}
    =
    \mu\{z:m(t,z)=0\}
    =0.
\]
Thus the accuracy is continuous and nondecreasing. By continuity of \(A_\mu\), the preceding accuracy
bounds also hold at the endpoints \(t_-\) and \(t_+\).
This proves the second assertion.

The expression for \(w_s\) follows directly from
\eqref{eq:two-mode-crossing-interval}. Since
\(q_s\to0\) and
\[
    \log\left(\frac{1+q_s}{1-q_s}\right)
    =
    2q_s+O(q_s^3),
\]
we obtain
\[
    w_s
    \sim
    \frac{2q_s}{\kappa_s}
    =
    \frac{1}{2\kappa_s}
    \exp\left(
        -\frac{\kappa_f\log 2}{2\kappa_s}
    \right).
\]
The right-hand side tends to zero as
\(\kappa_s\downarrow0\), which proves the third
assertion.

Finally, by
\eqref{eq:two-mode-crossing-interval},
\[
    t_\pm
    =
    t_\star+O\left(\frac{q_s}{\kappa_s}\right)
    =
    t_\star+o(1).
\]
In particular, \(t_-\to+\infty\), whereas
\(t_{\rm mem}(\varepsilon)\) remains bounded.
For sufficiently small \(\kappa_s\), one has
\[
    t_{\rm mem}(\varepsilon)<t_-.
\]
The accuracy bounds established above then imply
\[
    t_-
    \leq
    t_{\rm grok}^{\mu}(\varepsilon,\delta)
    \leq
    t_+.
\]
Since both endpoints are equal to
\(t_\star+o(1)\), we obtain
\[
    t_{\rm grok}^{\mu}(\varepsilon,\delta)
    =
    \frac{\log 2}{\kappa_s}+o(1).
\]
Together with the first assertion, this proves
the last assertion.
\end{proof}

\subsection{Population risk and classification accuracy}

The distinction between population risk and
classification accuracy can also be seen directly
in the preceding model. To this end, take the
evaluation target to be
\[
    f^\star(z)=z,
\]
whose sign agrees with the true class of every
evaluation example. The evaluation residual is
given by
\[
    F(w(t),z)-f^\star(z)
    =
    -2ze^{-\kappa_s t}
    +
    \frac14e^{-\kappa_f t}.
\]

\begin{corollary}
\label{cor:two-mode-classification-risk}
For the model of
\cref{thm:two-mode-classification}, the squared
population risk is given by
\begin{equation}
\label{eq:two-mode-classification-risk}
    \Risk_\mu(t)
    =
    \frac{14}{3}e^{-2\kappa_s t}
    +
    \frac{1}{32}e^{-2\kappa_f t}.
\end{equation}
In particular, the population risk is strictly
decreasing and analytic in time.

At the time \(t_\star=(\log 2)/\kappa_s\), one has
\[
    \Risk_\mu(t_\star)
    =
    \frac76
    +
    \frac{1}{32}
    \exp\left(
        -\frac{2\kappa_f\log 2}{\kappa_s}
    \right).
\]
Moreover,
\[
    \Risk_\mu(t_-)-\Risk_\mu(t_+)
    \longrightarrow0,
\]
as \(\kappa_s\downarrow0\).
\end{corollary}

\begin{proof}
By symmetry of the evaluation measure, one has
\[
    \int_{\mathcal Z_E}z\,\dd\mu(z)=0,
    \qquad
    \int_{\mathcal Z_E}z^2\,\dd\mu(z)
    =
    \int_1^2z^2\,\dd z
    =
    \frac73.
\]
It follows that
\[
\begin{aligned}
    \Risk_\mu(t)
    &=
    \frac12
    \int_{\mathcal Z_E}
    \left(
        -2ze^{-\kappa_s t}
        +
        \frac14e^{-\kappa_f t}
    \right)^2
    \dd\mu(z)\\
    &=
    2e^{-2\kappa_s t}
    \int_{\mathcal Z_E}z^2\,\dd\mu(z)
    +
    \frac{1}{32}e^{-2\kappa_f t}\\
    &=
    \frac{14}{3}e^{-2\kappa_s t}
    +
    \frac{1}{32}e^{-2\kappa_f t}.
\end{aligned}
\]
The assertions concerning monotonicity, analyticity,
and the value at \(t_\star\) follow immediately.

Differentiating
\eqref{eq:two-mode-classification-risk}, we obtain
\[
    \Risk_\mu'(t)
    =
    -\frac{28}{3}\kappa_s e^{-2\kappa_s t}
    -
    \frac{\kappa_f}{16}e^{-2\kappa_f t}.
\]
In particular,
\[
    \sup_{t\geq0}|\Risk_\mu'(t)|
    \leq
    \frac{28}{3}\kappa_s+\frac{\kappa_f}{16}.
\]
Consequently,
\[
\begin{aligned}
    0
    &\leq
    \Risk_\mu(t_-)-\Risk_\mu(t_+)\\
    &\leq
    \left(
        \frac{28}{3}\kappa_s
        +
        \frac{\kappa_f}{16}
    \right)w_s.
\end{aligned}
\]
The last expression tends to zero by
\cref{thm:two-mode-classification}.
\end{proof}

Thus, in the limit \(\kappa_s\downarrow0\), the
classification accuracy increases from zero to one
over an interval whose length tends to zero, while
the variation of population risk over the same
interval also tends to zero. A substantial improvement
in classification accuracy need not therefore be
accompanied by a substantial decrease in squared
population risk over the corresponding interval.

\begin{remark}
The preceding construction distinguishes two
properties of the learning dynamics. The time at
which generalization emerges is determined by the
slow spectral mode and is of order
\(\kappa_s^{-1}\). The concentration of crossing
times is governed by the remaining contribution
of the fast mode near that time. Since this
contribution is exponentially small in the ratio
\(\kappa_f/\kappa_s\), the interval containing
the crossing times becomes shorter as the
generalization time increases.

The two properties are therefore compatible
within a single gradient flow, even in the
absence of feature evolution.
\end{remark}

\begin{remark}
The evaluation readout in
\eqref{eq:two-mode-classification-readout} is
chosen so that the slow mode controls the
change of sign of the margins, while the fast
mode determines their dispersion near the
crossing time. The result is an existence
statement for a family of learning problems
and does not assert that such an organization
of the evaluation readout necessarily arises
in general neural network training.

Although the model varies with \(\kappa_s\),
its feature maps remain constant in time
for each member of the family.
\end{remark}

The results of this section distinguish the
regularity of the learning dynamics from that
of classification accuracy. The spectral
separation between training directions may
produce an arbitrarily long delay before
generalization, while the concentration of
margin crossing times may cause generalization
to emerge over an arbitrarily short interval.
Neither phenomenon requires a discontinuity
in the model outputs or an evolution of the
feature representation.

The distinction between these two time scales
also depends on the evaluation law and on the
manner in which the learned training directions
are transferred to evaluation outputs. We shall
examine this dependence in the following section.

\clearpage

\section{Distribution shift and identifiability}
\label{sec:shift}

We study the dependence of generalization on the
evaluation law while keeping the training trajectory
fixed. We first establish quantitative bounds on
the variation of evaluation visibility, population
risk, and signed risk reduction under changes of
the evaluation distribution. These bounds also give
conditions under which the separation between task
visibility and that of its orthogonal complement
is preserved. We then derive sufficient conditions
for selective generalization, whereby a model
achieves high classification accuracy under one
evaluation law but not under another, and examine
the stability of these conditions under perturbations.
Finally, through explicit linear constructions,
we show that neither evaluation visibility nor
the time required to achieve a prescribed
classification accuracy can, in general, be
determined from the training dynamics alone.

Until now, the evaluation law has generally been
regarded as fixed. We shall now compare the
quantities associated with different evaluation
laws on a common reference space, while keeping
the training trajectory and the target function
unchanged. In this setting, the cumulative
dissipation and transfer operators are common to
the evaluation laws, whereas the visibility
operators and population risks depend on the
respective evaluation metrics. This distinction
allows us to isolate the role of the evaluation
distribution in the emergence of generalization,
independently of changes in the training dynamics.

\subsection{Dependence on the evaluation law}

Let \(\mu\) and \(\nu\) be two probability measures
satisfying the assumptions of \cref{sec:setting}, and
let \(C_\mu\) and \(C_\nu\) be the corresponding
evaluation operators on \(\HE\). Since the training
trajectory is fixed, the operators \(W\), \(G\), and
\(S_\alpha\) are common to the two laws.

\begin{theorem}
\label{thm:shift}
With the notation introduced above, one has
\begin{align}
    \norm{H_{\mu,\alpha}-H_{\nu,\alpha}}_{\op}
    &\leq
    \norm G_{\op}^2
    \norm{C_\mu-C_\nu}_{\op},
    \label{eq:H-shift}\\
    \abs{\Risk_\mu(t)-\Risk_\nu(t)}
    &\leq
    \frac12
    \norm{C_\mu-C_\nu}_{\op}
    \norm{e(t)}_{\HE}^2,
    \label{eq:risk-shift}\\
    \abs{\Prog_\mu(s,T)-\Prog_\nu(s,T)}
    &\leq
    \norm{C_\mu-C_\nu}_{\op}
    \left(
        \norm{e(s)}_{\HE}\norm{d(s,T)}_{\HE}
        +\frac12\norm{d(s,T)}_{\HE}^2
    \right),
    \label{eq:progress-shift}
\end{align}
where
\[
    H_{\xi,\alpha}
    =
    S_\alpha G^*C_\xi G S_\alpha,
    \qquad
    d(s,T)=G(T,s)g(s).
\]
In the last two inequalities, the target function
\(f^\star\), and hence the residual \(e\), is assumed
to be common to the two evaluation laws.
\end{theorem}

\begin{proof}
By the definition of the visibility operators,
\[
    H_{\mu,\alpha}-H_{\nu,\alpha}
    =
    S_\alpha G^*(C_\mu-C_\nu)G S_\alpha.
\]
Since \(\norm{S_\alpha}_{\op}\leq1\), it follows
that
\[
\begin{aligned}
    \norm{H_{\mu,\alpha}-H_{\nu,\alpha}}_{\op}
    &\leq
    \norm{GS_\alpha}_{\op}^2
    \norm{C_\mu-C_\nu}_{\op}\\
    &\leq
    \norm G_{\op}^2
    \norm{C_\mu-C_\nu}_{\op},
\end{aligned}
\]
which proves \eqref{eq:H-shift}.

The second inequality follows from the identity
\[
    \Risk_\mu(t)-\Risk_\nu(t)
    =
    \frac12
    \ip{e(t)}{(C_\mu-C_\nu)e(t)}_{\HE},
\]
and the Cauchy--Schwarz inequality.

For the last assertion, put \(d=d(s,T)\). Subtracting
the two identities of \cref{thm:risk-balance}, we
obtain
\[
\begin{aligned}
    \Prog_\mu(s,T)-\Prog_\nu(s,T)
    &=
    \ip{e(s)}{(C_\mu-C_\nu)d}_{\HE}\\
    &\quad-
    \frac12\ip{d}{(C_\mu-C_\nu)d}_{\HE}.
\end{aligned}
\]
Applying the triangle inequality and the
Cauchy--Schwarz inequality gives
\eqref{eq:progress-shift}.
\end{proof}

\begin{remark}
For the multiplication operators considered in
\cref{sec:setting}, one has
\[
    \norm{C_\mu-C_\nu}_{\op}
    =
    \norm{r_\mu-r_\nu}_{L^\infty(\lambda)},
\]
where \(r_\mu\) and \(r_\nu\) are the densities
of \(\mu\) and \(\nu\) with respect to \(\lambda\).

The estimates of \cref{thm:shift} therefore concern
perturbations of the evaluation densities in the
\(L^\infty\)-norm. Under additional boundedness
assumptions on the residual and the evaluation
transfer, analogous estimates may be obtained
in weaker norms.

For instance, if \(e(t)\in L^\infty(\lambda;\R^p)\),
then
\[
    \abs{\Risk_\mu(t)-\Risk_\nu(t)}
    \leq
    \frac12
    \norm{e(t)}_{L^\infty(\lambda)}^2
    \norm{r_\mu-r_\nu}_{L^1(\lambda)}.
\]
Indeed, this follows directly by integrating
\(\norm{e(t,z)}_2^2(r_\mu(z)-r_\nu(z))\) with
respect to \(\lambda\).
\end{remark}

The preceding operator estimate also controls the
variation of the task spectrum.

\begin{corollary}
\label{cor:task-shift}
Let \(\mathsf F\) be a common signal filter for the
two evaluation laws, and put
\[
    d_{\mathrm{vis},\mathsf F}(\mu,\nu)
    =
    \norm{
        H_{\mu,\mathsf F}-H_{\nu,\mathsf F}
    }_{\op}.
\]
For every nonzero task subspace \(\Task\), one has
\[
    \abs{
        \lambda_{\min}^{\Task}(H_{\mu,\mathsf F})
        -
        \lambda_{\min}^{\Task}(H_{\nu,\mathsf F})
    }
    \leq
    d_{\mathrm{vis},\mathsf F}(\mu,\nu),
\]
and
\[
    \abs{
        \lambda_{\max}^{\Task}(H_{\mu,\mathsf F})
        -
        \lambda_{\max}^{\Task}(H_{\nu,\mathsf F})
    }
    \leq
    d_{\mathrm{vis},\mathsf F}(\mu,\nu).
\]

In particular, if \(\Task^\perp\ne0\) and
\[
    m
    =
    \lambda_{\min}^{\Task}(H_{\mu,\mathsf F})
    -
    \lambda_{\max}^{\Task^\perp}(H_{\mu,\mathsf F})
    >0,
\]
then
\[
\begin{aligned}
    &\lambda_{\min}^{\Task}(H_{\nu,\mathsf F})
    -
    \lambda_{\max}^{\Task^\perp}(H_{\nu,\mathsf F})\\
    &\qquad\geq
    m-2d_{\mathrm{vis},\mathsf F}(\mu,\nu).
\end{aligned}
\]
Thus the separation between the visibility of
\(\Task\) and that of its orthogonal complement
is preserved whenever
\[
    d_{\mathrm{vis},\mathsf F}(\mu,\nu)<\frac m2.
\]
\end{corollary}

\begin{proof}
The first two inequalities follow from the min--max
estimate of \cref{quoted:weyl}, applied to the
compressions of the two visibility operators
to \(\Task\).

Applying the same estimate to \(\Task^\perp\)
gives
\[
    \lambda_{\max}^{\Task^\perp}(H_{\nu,\mathsf F})
    \leq
    \lambda_{\max}^{\Task^\perp}(H_{\mu,\mathsf F})
    +
    d_{\mathrm{vis},\mathsf F}(\mu,\nu).
\]
Subtracting this inequality from the corresponding
lower bound for
\(\lambda_{\min}^{\Task}(H_{\nu,\mathsf F})\)
gives the desired result.
\end{proof}

\subsection{Selective generalization}
\label{sec:selective}

Let \(\mu\) and \(\nu\) be two evaluation laws, and
write \(A_\xi(t)=1-\Err_\xi(t)\) for the accuracy
associated with the law \(\xi\).

We shall first consider the case where the relation
between visibility and accuracy is given in terms
of prescribed thresholds. We shall then derive a
sufficient condition for selective generalization
directly from the signed risk identity and the
classification margins.

\begin{assumption}
\label{ass:two-sided-link}
Let \(0<\delta<1\), let \([s,T]\) be a training
interval, and let \(\mathsf F\) be a signal filter.
Suppose that \(g_s=g(s)\ne0\).

For each \(\xi\in\{\mu,\nu\}\), suppose there
exist constants
\[
    \vartheta_\xi^+>\vartheta_\xi^-\geq0,
\]
such that
\[
    \Geo_{\xi,\mathsf F}(g_s)
    \geq\vartheta_\xi^+
    \quad\Longrightarrow\quad
    A_\xi(T)\geq1-\delta,
\]
and
\[
    \Geo_{\xi,\mathsf F}(g_s)
    \leq\vartheta_\xi^-
    \quad\Longrightarrow\quad
    A_\xi(T)<1-\delta.
\]
\end{assumption}

\begin{theorem}
\label{thm:abstract-mu-not-nu}
Fix a training trajectory, an interval \([s,T]\),
a signal filter \(\mathsf F\), and a task subspace
\(\Task\). Suppose that \(g_s\ne0\), that \(g_s\in\Task\),
and that
\[
    \lambda_{\min}^{\Task}(H_{\mu,\mathsf F})
    \geq b_\mu,
    \qquad
    \lambda_{\max}^{\Task}(H_{\nu,\mathsf F})
    \leq a_\nu.
\]
If \cref{ass:two-sided-link} is satisfied and
\[
    b_\mu\geq\vartheta_\mu^+,
    \qquad
    a_\nu\leq\vartheta_\nu^-,
\]
then
\[
    A_\mu(T)\geq1-\delta,
    \qquad
    A_\nu(T)<1-\delta.
\]
\end{theorem}

\begin{proof}
Since \(g_s\) belongs to \(\Task\), its Rayleigh
quotients satisfy
\[
    \Geo_{\mu,\mathsf F}(g_s)
    \geq
    \lambda_{\min}^{\Task}(H_{\mu,\mathsf F})
    \geq b_\mu,
\]
and
\[
    \Geo_{\nu,\mathsf F}(g_s)
    \leq
    \lambda_{\max}^{\Task}(H_{\nu,\mathsf F})
    \leq a_\nu.
\]
The assertion follows by applying the two implications
of \cref{ass:two-sided-link}.
\end{proof}

\begin{corollary}
\label{cor:constructive-selective}
Suppose that the compressions of the two visibility
operators to \(\Task\) satisfy
\[
    \Pi H_{\mu,\mathsf F}\Pi|_\Task
    =
    bI_\Task,
    \qquad
    \Pi H_{\nu,\mathsf F}\Pi|_\Task
    =
    0,
\]
where \(b\geq\vartheta_\mu^+\). Then the conclusion
of \cref{thm:abstract-mu-not-nu} holds for every
nonzero \(g_s\in\Task\) for which
\cref{ass:two-sided-link} is satisfied.
\end{corollary}

\begin{proof}
The hypotheses of \cref{thm:abstract-mu-not-nu}
are satisfied with \(b_\mu=b\) and \(a_\nu=0\),
since \(\vartheta_\nu^-\geq0\).
\end{proof}

The preceding theorem is conditional upon the
relation between visibility and accuracy specified
in \cref{ass:two-sided-link}. We now replace this
assumption by conditions involving the residual
and the classification margins.

For the law \(\mu\), we shall use the upper bound
on classification error obtained from the signed
variation of population risk. For the law \(\nu\),
we shall estimate the measure of the set of
initially misclassified examples which can become
correctly classified under a displacement of
prescribed norm.

\begin{theorem}[Selective generalization]
\label{thm:explicit-mu-not-nu}
Let \(0<\delta<1\), and suppose that \(g_s\ne0\)
belongs to \(\Task\). Consider the exact-channel
visibility operators \(H_{\mu,+}\) and \(H_{\nu,+}\),
and suppose that
\[
    0<b_\mu
    \leq
    \lambda_{\min}^{\Task}(H_{\mu,+})
    \leq
    \lambda_{\max}^{\Task}(H_{\mu,+})
    \leq B_\mu,
\]
and
\[
    \lambda_{\max}^{\Task}(H_{\nu,+})
    \leq a_\nu.
\]

Put \(r_\mu=\norm{e(s)}_{C_\mu}\), and suppose that
the angle defined in \cref{cor:angle} satisfies
\[
    \cos\omega_\mu(s,T)\geq\zeta_\mu>0.
\]
Define
\[
    p_\mu
    =
    \min_{\substack{
        \sigma\in
        \{\norm{g_s}\sqrt{b_\mu},
        \norm{g_s}\sqrt{B_\mu}\}
    }}
    \left(
        \zeta_\mu r_\mu\sigma-\frac12\sigma^2
    \right).
\]

Suppose that the target function has margin
\(\gamma_\mu>0\) under \(\mu\), as in
\eqref{eq:target-margin}, and that the hypotheses
of \cref{thm:classification} are satisfied.

Suppose further that there exists a measurable
set \(B_\nu\) of measure
\[
    \nu(B_\nu)=m_\nu,
\]
such that, for \(\nu\)-almost every \(z\in B_\nu\),
there exists a class \(j(z)\ne y(z)\) satisfying
\[
    F_{j(z)}(\theta(s),z)
    -
    F_{y(z)}(\theta(s),z)
    \geq\gamma_\nu,
\]
where \(\gamma_\nu>0\).

If
\begin{equation}
\label{eq:selective-conditions}
\begin{aligned}
    \Risk_\mu(s)-p_\mu
    &\leq
    \frac{c_\mu\gamma_\mu^2\delta}{4},\\
    m_\nu
    -
    \frac{2a_\nu\norm{g_s}^2}{\gamma_\nu^2}
    &>\delta,
\end{aligned}
\end{equation}
then
\[
    A_\mu(T)\geq1-\delta,
    \qquad
    A_\nu(T)<1-\delta.
\]
\end{theorem}

\begin{proof}
We first consider the law \(\mu\). Since \(g_s\)
belongs to \(\Task\), one has
\[
    b_\mu\norm{g_s}^2
    \leq
    \ip{g_s}{H_{\mu,+}g_s}
    \leq
    B_\mu\norm{g_s}^2.
\]
By the exact-channel identity of
\cref{thm:reservoir}, this is equivalent to
\[
    \norm{g_s}\sqrt{b_\mu}
    \leq
    \norm{d(s,T)}_{C_\mu}
    \leq
    \norm{g_s}\sqrt{B_\mu}.
\]
The angle hypothesis and
\cref{thm:task-signed-progress} therefore give
\[
    \Prog_\mu(s,T)\geq p_\mu.
\]
It follows from \cref{thm:risk-balance} and the
first inequality of \eqref{eq:selective-conditions}
that
\[
\begin{aligned}
    \Risk_\mu(T)
    &=
    \Risk_\mu(s)-\Prog_\mu(s,T)\\
    &\leq
    \Risk_\mu(s)-p_\mu\\
    &\leq
    \frac{c_\mu\gamma_\mu^2\delta}{4},
\end{aligned}
\]
and hence \(A_\mu(T)\geq1-\delta\), by
\cref{thm:classification}.

We next consider the law \(\nu\). Put
\(d=d(s,T)=G(T,s)g_s\). Since \(GP_+=G\), one has
\[
\begin{aligned}
    \int_{\mathcal Z}\norm{d(z)}_2^2\,\dd\nu(z)
    &=
    \norm d_{C_\nu}^2\\
    &=
    \ip{g_s}{H_{\nu,+}g_s}\\
    &\leq
    a_\nu\norm{g_s}^2.
\end{aligned}
\]

Let \(B_\nu'\) be the subset of \(B_\nu\) consisting
of the examples which are correctly classified
at time \(T\). For \(z\in B_\nu'\), one has
\[
    F_{y(z)}(\theta(T),z)
    \geq
    F_{j(z)}(\theta(T),z).
\]
Since
\[
    F(\theta(T),z)
    =
    F(\theta(s),z)-d(z),
\]
it follows that
\[
\begin{aligned}
    d_{j(z)}(z)-d_{y(z)}(z)
    &\geq
    F_{j(z)}(\theta(s),z)
    -
    F_{y(z)}(\theta(s),z)\\
    &\geq\gamma_\nu.
\end{aligned}
\]
Consequently,
\[
\begin{aligned}
    \norm{d(z)}_2^2
    &\geq
    d_{j(z)}(z)^2+d_{y(z)}(z)^2\\
    &\geq
    \frac12
    \bigl(d_{j(z)}(z)-d_{y(z)}(z)\bigr)^2\\
    &\geq
    \frac{\gamma_\nu^2}{2}.
\end{aligned}
\]
Integrating over \(B_\nu'\), we obtain
\[
\begin{aligned}
    \frac{\gamma_\nu^2}{2}\nu(B_\nu')
    &\leq
    \int_{B_\nu'}\norm{d(z)}_2^2\,\dd\nu(z)\\
    &\leq
    a_\nu\norm{g_s}^2,
\end{aligned}
\]
so that
\[
    \nu(B_\nu')
    \leq
    \frac{2a_\nu\norm{g_s}^2}{\gamma_\nu^2}.
\]

The examples in \(B_\nu\setminus B_\nu'\) remain
misclassified at time \(T\). Thus
\[
\begin{aligned}
    \Err_\nu(T)
    &\geq
    \nu(B_\nu\setminus B_\nu')\\
    &\geq
    m_\nu
    -
    \frac{2a_\nu\norm{g_s}^2}{\gamma_\nu^2}\\
    &>\delta,
\end{aligned}
\]
by the second inequality of
\eqref{eq:selective-conditions}. Hence
\(A_\nu(T)<1-\delta\), as asserted.
\end{proof}

\begin{remark}
The two conclusions of \cref{thm:explicit-mu-not-nu}
are obtained in different ways. The first uses a lower
bound for the signed decrease of population risk,
whereas the second uses an upper bound for the measure
of the set of examples whose initial classification
errors can be corrected.

In particular, small visibility under \(\nu\) is
not sufficient to imply low accuracy. The initial
classifier must also be incorrect on a sufficiently
large set, with margins which cannot all be corrected
by the available evaluation displacement.

This is the role of the constants \(m_\nu\)
and \(\gamma_\nu\) in the second inequality of
\eqref{eq:selective-conditions}.
\end{remark}

\subsection{An example of selective generalization}

We give a one-dimensional example in which the
training trajectory is common to two evaluation
laws, although the evaluation output is learned
under one law and remains unchanged under the other.

\begin{proposition}
\label{prop:minimal-selective}
For every \(b>0\), there exist a one-dimensional
task subspace \(\Task\), a training trajectory,
and two point-mass evaluation laws \(\mu\) and
\(\nu\), such that
\[
    H_{\mu,+}|_\Task=bI_\Task,
    \qquad
    H_{\nu,+}|_\Task=0.
\]
Moreover, the population squared risk under \(\mu\)
vanishes at the right endpoint of the training
interval, whereas the example supporting \(\nu\)
remains misclassified throughout the interval.
\end{proposition}

\begin{proof}
Let \(\HP=\HS=\Task=\R\), and consider the training
output and loss
\[
    U_S(w)=w,
    \qquad
    \Loss_S(w)=\frac12(w-1)^2.
\]
Starting from \(w(0)=0\), the gradient flow is
given by
\[
    \dot w(t)=1-w(t),
    \qquad
    w(t)=1-e^{-t}.
\]
Thus
\[
    g(t)=-e^{-t},
    \qquad
    K_{SS}=1,
    \qquad
    \Pg(t,0)=e^{-t}.
\]
Fix \(T>0\). The cumulative dissipation operator
on \([0,T]\) is multiplication by
\[
    W(0,T)
    =
    \int_0^T e^{-2t}\,\dd t
    =
    \frac12(1-e^{-2T})>0,
\]
so that \(P_+=I\).

Let the evaluation space consist of two points
\(z_\mu\) and \(z_\nu\), with reference measure
\[
    \lambda
    =
    \frac12(\delta_{z_\mu}+\delta_{z_\nu}).
\]
We identify \(\HE\) with \(\R^2\), equipped with
the inner product
\[
    \ip{x}{y}_{\HE}
    =
    \frac12(x_1y_1+x_2y_2).
\]
The point-mass laws
\[
    \mu=\delta_{z_\mu},
    \qquad
    \nu=\delta_{z_\nu},
\]
give the multiplication operators
\[
    C_\mu=
    \begin{pmatrix}
        2&0\\
        0&0
    \end{pmatrix},
    \qquad
    C_\nu=
    \begin{pmatrix}
        0&0\\
        0&2
    \end{pmatrix}.
\]

Define the evaluation outputs by
\[
    F(w,z_\mu)
    =
    -\frac{\sqrt b}{2}
    +
    \frac{\sqrt b}{1-e^{-T}}w,
    \qquad
    F(w,z_\nu)=-1.
\]
Their derivative with respect to \(w\) is
\[
    J_E
    =
    \begin{pmatrix}
        \displaystyle\frac{\sqrt b}{1-e^{-T}}\\
        0
    \end{pmatrix}.
\]
It follows that the evaluation transfer operator
on \([0,T]\) is given by
\[
\begin{aligned}
    Gh
    &=
    \int_0^T J_Ee^{-t}h\,\dd t\\
    &=
    \begin{pmatrix}
        \sqrt b\,h\\
        0
    \end{pmatrix}.
\end{aligned}
\]
Consequently,
\[
    \norm{Gh}_{C_\mu}^2=bh^2,
    \qquad
    \norm{Gh}_{C_\nu}^2=0,
\]
which proves that
\[
    H_{\mu,+}=bI,
    \qquad
    H_{\nu,+}=0.
\]

We interpret the evaluation outputs as binary
classification scores, with positive true labels
at both points, and take the target function
to satisfy
\[
    f^\star(z_\mu)=\frac{\sqrt b}{2},
    \qquad
    f^\star(z_\nu)=1.
\]
At the initial time, both scores are negative,
whereas at time \(T\), one has
\[
    F(w(T),z_\mu)
    =
    \frac{\sqrt b}{2}
    =
    f^\star(z_\mu),
    \qquad
    F(w(T),z_\nu)=-1.
\]
It follows that
\[
    \Risk_\mu(T)=0,
    \qquad
    A_\mu(T)=1,
    \qquad
    A_\nu(T)=0.
\]
Since the score at \(z_\nu\) is constant throughout
the trajectory, this point remains misclassified
for every \(t\in[0,T]\).
\end{proof}

\begin{remark}
The two evaluation laws in the preceding construction
are supported on disjoint parts of the evaluation
space. In particular, the corresponding seminorms
have different kernels.

The training loss, the cumulative dissipation
operator, the training gradient, and the parameter
trajectory are identical under the two laws.
Even the evaluation transfer operator \(G\) is
common to both. The difference between the two
outcomes arises solely from the choice of the
evaluation law.

The example establishes selective generalization
at a prescribed time. A separation between the
training and evaluation times requires, in addition,
a condition on the time at which the training
loss reaches its prescribed tolerance.
\end{remark}

\subsection{Stability of selective generalization}

We now examine the effect of perturbing the visibility
operators appearing in \cref{thm:explicit-mu-not-nu}.
The estimates of \cref{cor:task-shift} apply directly
to perturbations of the evaluation laws. More
generally, we may consider perturbations of the
training trajectory and of the resulting transfer
operators.

\begin{corollary}
\label{cor:selective-window}
Suppose that the hypotheses of
\cref{thm:explicit-mu-not-nu} are satisfied, with
strict inequalities in \eqref{eq:selective-conditions}.

Let \(\widetilde H_{\mu,+}\) and
\(\widetilde H_{\nu,+}\) be perturbed visibility
operators on the same training-output space, and
suppose that
\[
    \norm{
        \widetilde H_{\xi,+}-H_{\xi,+}
    }_{\op}
    \leq\varepsilon_\xi,
    \qquad
    \xi\in\{\mu,\nu\}.
\]
Put
\[
    \widetilde b_\mu=b_\mu-\varepsilon_\mu,
    \qquad
    \widetilde B_\mu=B_\mu+\varepsilon_\mu,
    \qquad
    \widetilde a_\nu=a_\nu+\varepsilon_\nu.
\]
Suppose that \(\widetilde b_\mu>0\), and that
the remaining hypotheses of
\cref{thm:explicit-mu-not-nu}, including the
task alignment and the angle and margin conditions,
hold for the perturbed trajectory and evaluation
laws.

Define
\[
    \widetilde p_\mu
    =
    \min_{\substack{
        \sigma\in
        \{
            \norm{\widetilde g_s}\sqrt{\widetilde b_\mu},
            \norm{\widetilde g_s}\sqrt{\widetilde B_\mu}
        \}
    }}
    \left(
        \widetilde\zeta_\mu
        \widetilde r_\mu\sigma
        -
        \frac12\sigma^2
    \right).
\]
If
\[
    \widetilde\Risk_\mu(s)-\widetilde p_\mu
    \leq
    \frac{
        \widetilde c_\mu
        \widetilde\gamma_\mu^2
        \delta
    }{4},
\]
and
\[
    \widetilde m_\nu
    -
    \frac{
        2\widetilde a_\nu\norm{\widetilde g_s}^2
    }{
        \widetilde\gamma_\nu^2
    }
    >\delta,
\]
then
\[
    \widetilde A_\mu(T)\geq1-\delta,
    \qquad
    \widetilde A_\nu(T)<1-\delta.
\]
\end{corollary}

\begin{proof}
By the min--max estimate of \cref{quoted:weyl},
one has
\[
    \lambda_{\min}^{\Task}(\widetilde H_{\mu,+})
    \geq
    b_\mu-\varepsilon_\mu,
\]
and
\[
    \lambda_{\max}^{\Task}(\widetilde H_{\mu,+})
    \leq
    B_\mu+\varepsilon_\mu.
\]
Similarly,
\[
    \lambda_{\max}^{\Task}(\widetilde H_{\nu,+})
    \leq
    a_\nu+\varepsilon_\nu.
\]
The assertion therefore follows by applying
\cref{thm:explicit-mu-not-nu} to the perturbed
operators and the corresponding quantities.
\end{proof}

\begin{remark}
If the quantities appearing in
\cref{cor:selective-window} depend continuously on
the perturbation, the strict inequalities in
\eqref{eq:selective-conditions} persist for all
sufficiently small perturbations for which the
remaining hypotheses are satisfied.

However, closeness of the visibility operators
alone is not sufficient to preserve selective
generalization. In particular, it does not control
the angle between the evaluation displacement and
the target residual, nor does it control the measure
of examples with initially incorrect margins.
These quantities must be controlled separately.
\end{remark}

The results of this subsection distinguish the
spectral conditions governing evaluation visibility
from the additional conditions governing
classification accuracy. The former concern the
quadratic forms of the visibility operators, whereas
the latter also involve the residual and the
pointwise margins.

\subsection{Limits of training-only information}

The training kernel determines the evolution of the
training outputs, but it does not determine the
transfer of this evolution to an arbitrary evaluation
space. We illustrate this fact by a linear model
in which the training dynamics are fixed and the
evaluation law alone is varied.

\begin{theorem}
\label{thm:no-free}
Let \(g\in\R^2\setminus\{0\}\), let \(T>0\), and
let \(R>0\). There exist a linear model, a fixed
training sample, and evaluation laws \(\mu_1\)
and \(\mu_2\), for which all quantities determined
by the training trajectory are identical, while
\[
    \frac{
        \norm{G(T,0)g}_{C_{\mu_1}}^2
    }{
        \norm{G(T,0)g}_{C_{\mu_2}}^2
    }
    =R.
\]
There also exists an evaluation law \(\mu_0\)
such that
\[
    \norm{G(T,0)g}_{C_{\mu_0}}^2=0,
    \qquad
    \norm{G(T,0)g}_{C_{\mu_1}}^2>0.
\]
\end{theorem}

\begin{proof}
Consider the linear model
\[
    F(w,x)=x^\top w,
    \qquad
    w,x\in\R^2,
\]
and take the training inputs to be the canonical
basis vectors \(e_1,e_2\). The training output map
is then \(U_S(w)=w\), so that
\[
    J_S=I,
    \qquad
    K_{SS}=I.
\]
Choose the training target \(y=-g\), and start
the squared-loss gradient flow from \(w(0)=0\).
The initial training gradient is therefore \(g\),
and one has
\[
    \Pg(t,0)=e^{-t}I,
    \qquad
    W(0,T)=\frac12(1-e^{-2T})I.
\]

For an evaluation point \(x\), the corresponding
transfer operator is
\[
\begin{aligned}
    G_x(T,0)
    &=
    \int_0^T x^\top e^{-t}\,\dd t\\
    &=
    (1-e^{-T})x^\top.
\end{aligned}
\]
Put \(v=g/\norm g\), and choose a unit vector
\(v_\perp\) orthogonal to \(v\). Consider the three
evaluation laws
\[
    \mu_1=\delta_{\sqrt R\,v+v_\perp},
    \qquad
    \mu_2=\delta_v,
    \qquad
    \mu_0=\delta_{v_\perp}.
\]
These laws may be regarded as measures on a common
reference space by taking
\[
    \lambda=\frac13(\mu_0+\mu_1+\mu_2).
\]

Since \(g=\norm g\,v\), one has
\[
    (\sqrt R\,v+v_\perp)^\top g
    =
    \sqrt R\,\norm g,
    \qquad
    v^\top g=\norm g,
    \qquad
    v_\perp^\top g=0.
\]
Consequently,
\[
    \norm{G(T,0)g}_{C_{\mu_1}}^2
    =
    R(1-e^{-T})^2\norm g^2,
\]
and
\[
    \norm{G(T,0)g}_{C_{\mu_2}}^2
    =
    (1-e^{-T})^2\norm g^2,
\]
whereas
\[
    \norm{G(T,0)g}_{C_{\mu_0}}^2=0.
\]
This proves both assertions.

The training sample and its targets, the initial
condition, the loss, the training kernel, and the
parameter trajectory were fixed independently of
the evaluation laws.
\end{proof}

\begin{corollary}
\label{cor:training-only}
There is no function of the training loss, the
parameter norm, the cumulative dissipation operator,
the training-kernel spectrum, and the training
gradient which determines evaluation visibility
for every evaluation law.

Nor do these quantities determine, in general,
the first time at which a prescribed classification
accuracy is attained after the training loss has
reached a given tolerance.
\end{corollary}

\begin{proof}
The first assertion follows from \cref{thm:no-free},
since all the training-side quantities remain
unchanged while the evaluation visibility varies.

To prove the second assertion, consider again the
linear model used in the proof of \cref{thm:no-free}.
Its parameter trajectory is given by
\[
    w(t)=-(1-e^{-t})g.
\]
Let \(v=g/\norm g\), and consider the evaluation
points
\[
    x_+=-v,
    \qquad
    x_-=v,
\]
both with positive true binary labels. For every
\(t>0\), one has
\[
    F(w(t),x_+)
    =
    (1-e^{-t})\norm g>0,
\]
whereas
\[
    F(w(t),x_-)
    =
    -(1-e^{-t})\norm g<0.
\]
Thus, for the point-mass laws
\[
    \xi_+=\delta_{x_+},
    \qquad
    \xi_-=\delta_{x_-},
\]
the corresponding accuracies satisfy
\[
    A_{\xi_+}(t)=1,
    \qquad
    A_{\xi_-}(t)=0,
\]
for every \(t>0\).

For a training tolerance
\[
    0<\eps<\frac12\norm g^2,
\]
the first passage time of the training loss is
\[
    t_{\mathrm{mem}}(\eps)
    =
    \frac12
    \log\left(
        \frac{\norm g^2}{2\eps}
    \right)>0.
\]
It follows that, for every \(0<\delta<1\),
\[
    t_{\mathrm{grok}}^{\xi_+}(\eps,\delta)
    =
    t_{\mathrm{mem}}(\eps),
    \qquad
    t_{\mathrm{grok}}^{\xi_-}(\eps,\delta)
    =
    +\infty.
\]
The training trajectory and all quantities determined
by it are nevertheless identical under the two
evaluation laws.
\end{proof}

\begin{remark}
The obstruction in \cref{thm:no-free} does not
concern the accuracy with which the training dynamics
are observed. Even complete knowledge of these
dynamics does not suffice to recover the evaluation
visibility for an arbitrary evaluation law.

Additional hypotheses relating the training sample
to the evaluation space are therefore necessary
for a training-only characterization of generalization.
In the present framework, this dependence is expressed
by the transfer operator \(G\) and the evaluation
operator \(C_\mu\).
\end{remark}

\clearpage

\section{Stochastic updates and optimizer geometry}
\label{sec:stochastic}

We study the effect of stochastic training updates on
the evolution of population risk. Within a linearized
update model, we derive an exact expression for the
conditional expected variation of squared population
risk, separating the contribution of the mean update
from that of its covariance. This identity gives
necessary and sufficient conditions for a decrease
in expected population risk and allows us to determine
the range of admissible step sizes. We also examine
how the orientation of the noise covariance relative
to the evaluation transfer affects population risk.
In particular, we characterize the covariance
structures that minimize or maximize the contribution
of stochastic fluctuations, as well as those that
remain invisible under a given evaluation law.

We have thus far considered deterministic training
dynamics. The introduction of stochastic updates
requires us to account not only for the direction
and magnitude of the mean update, but also for the
distribution of its fluctuations among training
directions. The relation between stochastic gradient
descent and generalization has been studied by means
of algorithmic stability~\cite{hardt2016stability}.
Our purpose here is different. We consider the
variation of population risk during a single
linearized update, conditional on the current state
of the training process, without making any
assumption on the generalization gap.

Let \((\mathcal F_k)_{k\geq0}\) be a filtration and
consider the linearized update model
\begin{equation}
\label{eq:stochastic-update}
    e_{k+1}
    =
    e_k-\eta A_k(\bar g_k+\xi_k),
\end{equation}
where \(\eta>0\) and \(A_k:\HS\to\HE\) is a bounded
operator. We suppose that \(e_k\), \(A_k\), and
\(\bar g_k\) are \(\mathcal F_k\)-measurable, and
that the random vector \(\xi_k\) satisfies
\[
    \E[\xi_k\mid\mathcal F_k]=0,
    \qquad
    \E[\xi_k\xi_k^*\mid\mathcal F_k]=\Sigma_k.
\]
Thus \(\bar g_k\) is the conditional mean of the
stochastic gradient, and \(\Sigma_k\) is its
conditional covariance operator. We assume that
the random vectors
\[
    e_k,\quad \bar g_k,\quad \xi_k,\quad
    A_k\bar g_k,\quad A_k\xi_k
\]
have finite second moments and, in infinite
dimension, that \(\Sigma_k\) is of trace class.

All identities below are exact for the update model
\eqref{eq:stochastic-update}. For a nonlinear output
map, their application to an actual parameter update
requires control of the remainder in its linearization.

\subsection{Conditional variation of population risk}

\begin{theorem}
\label{thm:drift-variance}
Let
\[
    \Risk_k
    =
    \frac12\norm{e_k}_{C_\mu}^2.
\]
Under the assumptions above, one has
\begin{align}
    \E[\Risk_{k+1}-\Risk_k\mid\mathcal F_k]
    &=
    -\eta\ip{e_k}{C_\mu A_k\bar g_k}
    +\frac{\eta^2}{2}
    \norm{A_k\bar g_k}_{C_\mu}^2
    \notag\\
    &\quad+
    \frac{\eta^2}{2}
    \tr\left(A_k^*C_\mu A_k\Sigma_k\right).
    \label{eq:drift-variance}
\end{align}
Consequently, the conditional expectation of the
population risk decreases strictly if and only if
\begin{equation}
\label{eq:optimizer-safe}
    \ip{e_k}{C_\mu A_k\bar g_k}
    >
    \frac{\eta}{2}
    \left(
        \norm{A_k\bar g_k}_{C_\mu}^2
        +
        \tr(A_k^*C_\mu A_k\Sigma_k)
    \right).
\end{equation}
\end{theorem}

\begin{proof}
Put \(v_k=A_k\bar g_k\) and \(z_k=A_k\xi_k\).
By \eqref{eq:stochastic-update}, one has
\[
    e_{k+1}=e_k-\eta v_k-\eta z_k,
\]
and hence
\[
\begin{aligned}
    \Risk_{k+1}-\Risk_k
    &=
    -\eta\ip{e_k}{C_\mu(v_k+z_k)}
    +\frac{\eta^2}{2}
    \norm{v_k+z_k}_{C_\mu}^2\\
    &=
    -\eta\ip{e_k}{C_\mu v_k}
    -\eta\ip{e_k}{C_\mu z_k}\\
    &\quad+
    \frac{\eta^2}{2}\norm{v_k}_{C_\mu}^2
    +\eta^2\ip{v_k}{C_\mu z_k}
    +\frac{\eta^2}{2}\norm{z_k}_{C_\mu}^2.
\end{aligned}
\]
Since \(A_k\) is \(\mathcal F_k\)-measurable and
\(\E[\xi_k\mid\mathcal F_k]=0\), one has
\[
    \E[z_k\mid\mathcal F_k]=0.
\]
The terms which are linear in \(z_k\) therefore
vanish upon taking the conditional expectation.
For the remaining quadratic term, one has
\[
\begin{aligned}
    \E[\norm{z_k}_{C_\mu}^2\mid\mathcal F_k]
    &=
    \E\left[
        \ip{\xi_k}{A_k^*C_\mu A_k\xi_k}
        \mid\mathcal F_k
    \right]\\
    &=
    \tr(A_k^*C_\mu A_k\Sigma_k).
\end{aligned}
\]
Substitution gives \eqref{eq:drift-variance}.
Since \(\eta>0\), the right-hand side of this
identity is negative precisely when
\eqref{eq:optimizer-safe} holds.
\end{proof}

\begin{remark}
The first two terms on the right-hand side of
\eqref{eq:drift-variance} are the signed variation
of population risk associated with the mean
displacement \(\eta A_k\bar g_k\), as in
\cref{thm:risk-balance}. The last term is due to
the covariance of the stochastic update.

Since both \(A_k^*C_\mu A_k\) and \(\Sigma_k\)
are positive semidefinite, one has
\[
    \tr(A_k^*C_\mu A_k\Sigma_k)
    =
    \norm{C_\mu^{1/2}A_k\Sigma_k^{1/2}}_{\Frob}^2
    \geq0,
\]
where, in infinite dimension, the Frobenius norm
is replaced by the Hilbert--Schmidt norm.

Thus, for fixed \(e_k\), \(A_k\), and \(\bar g_k\),
the conditional expected risk after a stochastic
update is at least as large as that obtained from
the corresponding mean update.
\end{remark}

The preceding theorem also gives an explicit
condition on the admissible step size.

\begin{corollary}
\label{cor:stochastic-step-size}
Put
\[
    D_k=\ip{e_k}{C_\mu A_k\bar g_k},
\]
and
\[
    V_k
    =
    \norm{A_k\bar g_k}_{C_\mu}^2
    +
    \tr(A_k^*C_\mu A_k\Sigma_k).
\]
If \(D_k>0\), then \(V_k>0\), and the conditional
expected population risk decreases strictly if and
only if
\[
    0<\eta<\frac{2D_k}{V_k}.
\]
For fixed \(e_k\), \(A_k\), \(\bar g_k\), and
\(\Sigma_k\), the step size minimizing the
conditional expected risk is
\[
    \eta_k^\star=\frac{D_k}{V_k},
\]
and the corresponding expected decrease is
\[
    \Risk_k
    -
    \E[\Risk_{k+1}\mid\mathcal F_k]
    =
    \frac{D_k^2}{2V_k}.
\]
If \(D_k\leq0\), no positive step size gives a
strict decrease in conditional expected risk.
\end{corollary}

\begin{proof}
By \cref{thm:drift-variance}, one has
\[
    \E[\Risk_{k+1}-\Risk_k\mid\mathcal F_k]
    =
    -\eta D_k+\frac12\eta^2V_k.
\]
If \(D_k>0\), the Cauchy--Schwarz inequality implies
that \(\norm{A_k\bar g_k}_{C_\mu}>0\), and hence
\(V_k>0\).

The polynomial on the right is negative precisely
when \(0<\eta<2D_k/V_k\), and attains its minimum
at \(\eta=D_k/V_k\). Substituting this value gives
\[
    -\eta_k^\star D_k
    +
    \frac12(\eta_k^\star)^2V_k
    =
    -\frac{D_k^2}{2V_k},
\]
which proves the first two assertions.

If \(D_k\leq0\), both terms in the expression for
the conditional expected risk variation are
nonnegative for every \(\eta>0\).
\end{proof}

\subsection{Dependence on the covariance operator}

We now fix the evaluation transfer and examine the
dependence of the variance term on the spectral
distribution of the covariance operator.

\begin{proposition}
\label{prop:noise-placement}
Let \(A:\HS\to\HE\) be a bounded operator, with
\(\HS\) finite-dimensional and nonzero, and put
\[
    B=A^*C_\mu A.
\]
For a prescribed \(\sigma^2>0\), consider the
positive semidefinite operators \(\Sigma\) on
\(\HS\) satisfying \(\tr\Sigma=\sigma^2\).
Then
\begin{align}
    \min_{\substack{
        \Sigma\succeq0\\
        \tr\Sigma=\sigma^2
    }}
    \tr(B\Sigma)
    &=
    \sigma^2\lambda_{\min}(B),
    \label{eq:min-noise}\\
    \max_{\substack{
        \Sigma\succeq0\\
        \tr\Sigma=\sigma^2
    }}
    \tr(B\Sigma)
    &=
    \sigma^2\lambda_{\max}(B).
    \label{eq:max-noise}
\end{align}

The minimum is attained precisely when the range
of \(\Sigma\) is contained in the eigenspace
associated with \(\lambda_{\min}(B)\), and the
maximum is attained precisely when its range is
contained in the eigenspace associated with
\(\lambda_{\max}(B)\).
\end{proposition}

\begin{proof}
Let \(u_1,\ldots,u_n\) be an orthonormal eigenbasis
of \(B\), with corresponding eigenvalues
\(\lambda_1,\ldots,\lambda_n\). One has
\[
    \tr(B\Sigma)
    =
    \sum_{i=1}^n
    \lambda_i\ip{u_i}{\Sigma u_i},
\]
where
\[
    \ip{u_i}{\Sigma u_i}\geq0,
    \qquad
    \sum_{i=1}^n\ip{u_i}{\Sigma u_i}
    =
    \sigma^2.
\]
It follows that
\[
    \sigma^2\lambda_{\min}(B)
    \leq
    \tr(B\Sigma)
    \leq
    \sigma^2\lambda_{\max}(B).
\]
Both bounds are attained. Indeed, if \(u_-\) and
\(u_+\) are unit eigenvectors associated with the
smallest and largest eigenvalues, respectively,
it suffices to take
\[
    \Sigma_-=\sigma^2u_-u_-^*,
    \qquad
    \Sigma_+=\sigma^2u_+u_+^*.
\]

For the equality conditions, observe that
\[
    \tr(B\Sigma)
    -
    \sigma^2\lambda_{\min}(B)
    =
    \sum_{i=1}^n
    \bigl(\lambda_i-\lambda_{\min}(B)\bigr)
    \ip{u_i}{\Sigma u_i}.
\]
Since each term is nonnegative, the sum vanishes
if and only if
\[
    \ip{u_i}{\Sigma u_i}=0,
\]
for every \(i\) such that
\(\lambda_i>\lambda_{\min}(B)\). By positivity
of \(\Sigma\), this is equivalent to
\(\Sigma u_i=0\) for all such \(i\), or,
equivalently, to the inclusion of the range of
\(\Sigma\) in the eigenspace associated with
\(\lambda_{\min}(B)\).

The equality condition for the maximum is proved
in the same way.
\end{proof}

\begin{corollary}
\label{cor:invisible-covariance}
With the notation of \cref{prop:noise-placement},
the following conditions are equivalent,
\[
    \tr(B\Sigma)=0,
\]
\[
    C_\mu^{1/2}A\Sigma^{1/2}=0,
\]
and
\[
    \ran\Sigma
    \subseteq
    \Ker(C_\mu^{1/2}A).
\]
In particular, a nonzero covariance operator with
zero instantaneous contribution to population risk
exists if and only if
\[
    \Ker(C_\mu^{1/2}A)\ne0.
\]
\end{corollary}

\begin{proof}
Since \(B=A^*C_\mu A\), one has
\[
\begin{aligned}
    \tr(B\Sigma)
    &=
    \tr(\Sigma^{1/2}A^*C_\mu A\Sigma^{1/2})\\
    &=
    \norm{C_\mu^{1/2}A\Sigma^{1/2}}_{\Frob}^2,
\end{aligned}
\]
which proves the equivalence of the first two
conditions. The third follows from the identity
\[
    \ran\Sigma^{1/2}=\ran\Sigma,
\]
valid in finite dimension.

If the kernel of \(C_\mu^{1/2}A\) contains a
unit vector \(u\), the operator
\[
    \Sigma=\sigma^2uu^*
\]
has trace \(\sigma^2\) and satisfies all three
conditions. Conversely, a nonzero covariance
satisfying these conditions has nonzero range
contained in the indicated kernel.
\end{proof}

\begin{remark}
The kernel occurring in
\cref{cor:invisible-covariance} is the kernel of
the evaluation transfer weighted by \(C_\mu\).
It may depend on the evaluation law and need
not coincide with the exact reservoir
\(\Ker W\) introduced in \cref{sec:reservoir}.

The result nevertheless gives a stochastic
analogue of evaluation invisibility. A direction
which is invisible under a given evaluation law
may support covariance without contributing to
the instantaneous variance of the evaluation
output under that law.

In particular, the same covariance operator may
have zero instantaneous cost under one evaluation
law and a strictly positive cost under another.
\end{remark}

\subsection{Mean updates and subsequent dynamics}

The two contributions to the expected population
risk have different geometric properties. The
contribution of the mean update depends on its
position relative to the target residual, whereas
that of the covariance depends on its position
relative to the spectral subspaces of
\(A_k^*C_\mu A_k\).

The results above concern a single update with
the current residual, transfer operator, and mean
gradient held fixed. They do not imply that
deterministic training is preferable to stochastic
training over an entire trajectory. The stochastic
update may change the parameter region subsequently
visited by the training process, and hence the
future residuals, transfer operators, and mean
gradients.

Any eventual reduction in population risk due to
such changes must be distinguished from the
instantaneous variance contribution in
\eqref{eq:drift-variance}, which is always
nonnegative. In particular, a beneficial effect
of stochasticity over several updates cannot be
deduced from the one-step variance term alone.

\clearpage

\section{Categorical, geometric, and symmetry structures}
\label{sec:structures}

We examine the algebraic and geometric structures
underlying the preceding constructions. We first
give a categorical formulation of the composition
of training intervals and study the quotient of
the training-output space by the reservoir. We
then relate cumulative dissipation to observability
and describe the evolution of spectral subspaces
through Grassmannian geometry and parallel transport.
Finally, we study the decomposition induced by
symmetries of the training dynamics and describe
changes in the dimension of the signal channel
using determinantal geometry.

Most of the constructions used in this section are
classical. Their interest here lies in their
application to the dissipation and transfer operators
of the present theory. The additional hypotheses
required for the geometric and representation-theoretic
formulations will be specified as they arise.

Throughout this section, \(V=\HS\) is finite-dimensional
and \(E=\HE\). For simplicity, we use the notation
\[
    P_{ts}=\Pg(t,s),
    \qquad
    G_{ts}=\G(t,s),
    \qquad
    W_{st}=\W(s,t).
\]
All the intervals considered below belong to the same
realized training trajectory.

\subsection{Spaces, maps, and quadratic forms}
\label{sec:categorical-objects}

Let \(\mathbf{Hilb}_{\R}\) be the category of real
Hilbert spaces and bounded linear maps, and denote
by \(\mathbf{FinHilb}_{\R}\) its full subcategory of
finite-dimensional spaces.

The spaces occurring in the preceding theory are
the parameter space \(\HP\), the training-output
space \(V\), and the evaluation space \(E\). To
these are associated the inclusions
\[
    \Task\hookrightarrow V,
    \qquad
    \Ker W_{st}\hookrightarrow V,
\]
and the quotient map
\[
    V\longrightarrow V/\Ker W_{st}.
\]
The Jacobians, propagators, and transfer operators
are morphisms between these spaces, while \(M\),
\(W_{st}\), and \(C_\mu\) are positive semidefinite
endomorphisms of the corresponding spaces.

The actual gradient and the target function enter
the theory as distinguished vectors. From the
categorical point of view, a vector \(v\) in a
Hilbert space \(H\) may be regarded as a morphism
\[
    \R\longrightarrow H,
    \qquad
    a\longmapsto av.
\]
Thus the target-dependent theory involves not only
the spaces and their operators, but also distinguished
maps from the scalar field.

Let \(\mathcal P(H)\) denote the partially ordered
set of bounded positive semidefinite operators on
a Hilbert space \(H\), ordered by the Loewner order.
A bounded linear map \(T:H_1\to H_2\) induces a map
\[
    \mathcal P(H_2)\longrightarrow\mathcal P(H_1),
    \qquad
    C\longmapsto T^*CT.
\]
This map preserves the order, and, for composable
maps \(T_1\) and \(T_2\), one has
\[
    (T_2T_1)^*C(T_2T_1)
    =
    T_1^*(T_2^*CT_2)T_1.
\]
Consequently, \(H\mapsto\mathcal P(H)\) defines a
contravariant functor from \(\mathbf{Hilb}_{\R}\)
to the category of partially ordered sets.

In particular, the visibility operator
\begin{equation}
\label{eq:visibility-pullback}
    H_{\mu,\mathsf F}
    =
    (G_{ts}\mathsf F)^*C_\mu(G_{ts}\mathsf F),
\end{equation}
is the pullback of the evaluation form by the
filtered transfer map.

For a fixed trajectory and a fixed filter
\(\mathsf F\), the dependence of this operator
on the evaluation law is affine. Indeed, since
\[
    C_{a\mu+(1-a)\nu}
    =
    aC_\mu+(1-a)C_\nu,
    \qquad 0\leq a\leq1,
\]
one has
\[
    H_{a\mu+(1-a)\nu,\mathsf F}
    =
    aH_{\mu,\mathsf F}
    +(1-a)H_{\nu,\mathsf F}.
\]
This is the positive-form structure underlying
the comparison of evaluation laws in
\cref{sec:shift}.

\subsection{Composition of training intervals}
\label{sec:interval-category}

We regard the ordered set of training times as
a category \(\mathbf{Time}\), with one morphism
\(s\to t\) whenever \(s\leq t\), and no morphism
otherwise.

By \cref{prop:exact-gradient-propagation}, the
assignments
\[
    t\longmapsto V,
    \qquad
    (s\to t)\longmapsto P_{ts},
\]
define a functor from \(\mathbf{Time}\) to
\(\mathbf{FinHilb}_{\R}\). We shall extend this
description to include the transfer and dissipation
operators associated with each interval.

\begin{definition}
\label{def:interval-summary-category}
Fix a real Hilbert space \(E\). Let \(\mathbf D_E\)
be the category whose objects are finite-dimensional
real Hilbert spaces and whose morphisms from \(V_0\)
to \(V_1\) are triples \((P,G,W)\), where
\[
    P:V_0\to V_1,
    \qquad
    G:V_0\to E,
    \qquad
    W:V_0\to V_0,
\]
with \(W=W^*\succeq0\).

Composition is defined by
\begin{equation}
\label{eq:interval-summary-composition}
\begin{aligned}
    &(P_2,G_2,W_2)\circ(P_1,G_1,W_1)\\
    &\qquad=
    \bigl(
        P_2P_1,\,
        G_1+G_2P_1,\,
        W_1+P_1^*W_2P_1
    \bigr),
\end{aligned}
\end{equation}
and the identity of \(V\) is the triple
\((I_V,0,0)\).
\end{definition}

The category \(\mathbf D_E\) is defined independently
of any particular training model. A realized trajectory
will determine a family of morphisms in this category,
although not every morphism need arise in this way.

\begin{proposition}
\label{prop:interval-functor}
The composition defined in
\eqref{eq:interval-summary-composition} is associative
and preserves positive semidefiniteness of the
dissipation component.

For \(s\leq r\leq t\), the operators associated
with the training trajectory satisfy
\begin{align}
    P_{ts}
    &=
    P_{tr}P_{rs},
    \label{eq:categorical-P-composition}\\
    G_{ts}
    &=
    G_{rs}+G_{tr}P_{rs},
    \label{eq:categorical-G-composition}\\
    W_{st}
    &=
    W_{sr}+P_{rs}^*W_{rt}P_{rs}.
    \label{eq:categorical-W-composition}
\end{align}
Consequently, the assignments
\[
    \mathfrak F(t)=V,
    \qquad
    \mathfrak F(s\to t)=(P_{ts},G_{ts},W_{st}),
\]
define a functor
\[
    \mathfrak F:\mathbf{Time}\longrightarrow\mathbf D_E.
\]
\end{proposition}

\begin{proof}
For three composable triples, either association
of the composition gives the transition map
\(P_3P_2P_1\), the transfer map
\[
    G_1+G_2P_1+G_3P_2P_1,
\]
and the quadratic form
\[
    W_1
    +P_1^*W_2P_1
    +(P_2P_1)^*W_3(P_2P_1).
\]
This proves associativity. The identity laws follow
directly from the definition, and positivity follows
from the positivity of each summand of the last
expression.

The first of the three trajectory identities is
the propagator composition law. To prove the
second, split the integral defining \(G_{ts}\)
at \(r\). Since
\[
    \Pg(\tau,s)=\Pg(\tau,r)P_{rs},
    \qquad r\leq\tau\leq t,
\]
one obtains
\[
\begin{aligned}
    G_{ts}
    &=
    \int_s^r K_{ES}^M(\tau)\Pg(\tau,s)\,\dd\tau\\
    &\quad+
    \int_r^t K_{ES}^M(\tau)\Pg(\tau,r)
    P_{rs}\,\dd\tau\\
    &=
    G_{rs}+G_{tr}P_{rs}.
\end{aligned}
\]
Similarly,
\[
\begin{aligned}
    W_{st}
    &=
    W_{sr}
    +
    \int_r^t
    P_{rs}^*\Pg(\tau,r)^*K_{SS}^M(\tau)
    \Pg(\tau,r)P_{rs}\,\dd\tau\\
    &=
    W_{sr}+P_{rs}^*W_{rt}P_{rs},
\end{aligned}
\]
which proves \eqref{eq:categorical-W-composition}.

Finally, the operators associated with an interval
of length zero are \((I_V,0,0)\), so that the
assignments preserve both composition and identities.
\end{proof}

The identity \eqref{eq:categorical-W-composition}
expresses the dissipation over a composite interval
as the sum of the dissipation over the first interval
and the pullback of that over the second.

The composition considered here concerns consecutive
intervals of a fixed trajectory. It is distinct
from the categorical composition of parameterized
maps and learning rules studied in
\cite{fong2019backprop}.

\begin{remark}
\label{rem:target-marked-functor}
The triples \((P_{ts},G_{ts},W_{st})\) do not
contain the target function. Its role may be
recovered by adjoining the evaluation residual
to the gradient state.

Consider the endomorphism of \(V\oplus E\) given by
\[
    \mathcal U_{ts}
    =
    \begin{pmatrix}
        P_{ts}&0\\
        -G_{ts}&I_E
    \end{pmatrix}.
\]
By \eqref{eq:categorical-P-composition} and
\eqref{eq:categorical-G-composition}, one has
\[
    \mathcal U_{ts}
    =
    \mathcal U_{tr}\mathcal U_{rs}.
\]
Moreover, the exact trajectory satisfies
\[
    \begin{pmatrix}
        g(t)\\
        e(t)
    \end{pmatrix}
    =
    \mathcal U_{ts}
    \begin{pmatrix}
        g(s)\\
        e(s)
    \end{pmatrix}.
\]
Applying the quadratic form
\(\frac12\norm e_{C_\mu}^2\) to the second
component gives the signed risk identity of
\cref{thm:risk-balance}.

Thus the transfer of the gradient and that of the
evaluation residual admit a common linear
description along the realized trajectory.
The variation of population risk is recovered
once the initial residual and the evaluation
form are specified.
\end{remark}

\subsection{The quotient by the reservoir}
\label{sec:categorical-quotient}

Fix an interval \([s,t]\) and put \(R=\Ker W_{st}\).
By \cref{thm:reservoir}, the transfer map annihilates
\(R\). Hence it factors through the quotient of
\(V\) by \(R\).

\begin{proposition}
\label{prop:quotient-transfer}
Let \(q:V\to V/R\) be the quotient map. There
exists a unique linear map
\(\overline G:V/R\to E\) such that
\begin{equation}
\label{eq:quotient-G-factorization}
    G_{ts}=\overline Gq,
    \qquad
    \overline G([h])=G_{ts}h.
\end{equation}
Moreover, the formula
\[
    \langle[h],[k]\rangle_W
    =
    \langle h,W_{st}k\rangle,
\]
defines an inner product on \(V/R\). For every
evaluation law \(\mu\), one has
\begin{equation}
\label{eq:quotient-transfer-bound}
    \norm{C_\mu^{1/2}\overline G[h]}^2
    \leq
    \beta_\mu(s,t)\norm{[h]}_W^2.
\end{equation}

With respect to this inner product, the map
\[
    V/R\longrightarrow\ran W_{st}^{1/2},
    \qquad
    [h]\longmapsto W_{st}^{1/2}h,
\]
is an isometry onto \(\ran W_{st}^{1/2}\).
\end{proposition}

\begin{proof}
The existence and uniqueness of \(\overline G\)
follow from the inclusion
\[
    R=\Ker W_{st}\subseteq\Ker G_{ts},
\]
and the universal property of the quotient.

Since \(W_{st}\) is self-adjoint and annihilates
\(R\), the expression
\(\langle h,W_{st}k\rangle\) depends only on
the classes of \(h\) and \(k\) modulo \(R\).
Furthermore,
\[
    \langle h,W_{st}h\rangle=0
    \quad\Longleftrightarrow\quad
    h\in R,
\]
so that the induced form is positive definite
on \(V/R\).

The inequality \eqref{eq:quotient-transfer-bound}
is precisely the dissipation domination estimate
of \cref{thm:domination} expressed on the quotient.
Finally,
\[
    \norm{[h]}_W^2
    =
    \langle h,W_{st}h\rangle
    =
    \norm{W_{st}^{1/2}h}^2,
\]
which proves the isometry assertion.
\end{proof}

The quotient \(V/R\), equipped with its dissipation
inner product, gives an intrinsic formulation of
the factorization in \cref{cor:factorization}.
The training dynamics determine both the quotient
and its metric, while evaluation determines a
bounded map from this quotient to \(E\).

The equivalence of two vectors modulo \(R\)
concerns their action through the finite-window
operators of a fixed trajectory. It does not
assert that distinct initial conditions generate
the same nonlinear training trajectory.

\begin{proposition}
\label{prop:compositional-invisibility}
The morphisms \((P,G,W)\) of \(\mathbf D_E\)
satisfying
\[
    \Ker W\subseteq\Ker G,
\]
form a subcategory of \(\mathbf D_E\).

Let \(C\) be a bounded positive semidefinite
operator on \(E\), and suppose that two composable
morphisms satisfy
\[
    G_i^*CG_i\preceq\beta_iW_i,
    \qquad
    \beta_i\geq0,
    \qquad i=1,2.
\]
Then their composite satisfies
\begin{equation}
\label{eq:composite-domination}
\begin{aligned}
    &(G_1+G_2P_1)^*C(G_1+G_2P_1)\\
    &\qquad\preceq
    (\beta_1+\beta_2)
    (W_1+P_1^*W_2P_1).
\end{aligned}
\end{equation}
\end{proposition}

\begin{proof}
Since both summands of the composite dissipation
operator are positive semidefinite, one has
\[
\begin{aligned}
    \Ker(W_1+P_1^*W_2P_1)
    &=
    \Ker W_1
    \cap
    P_1^{-1}(\Ker W_2).
\end{aligned}
\]
Thus, if \(h\) belongs to the kernel of the
composite dissipation operator, then
\[
    W_1h=0,
    \qquad
    W_2P_1h=0.
\]
By the hypotheses on the two morphisms, it follows
that
\[
    G_1h=0,
    \qquad
    G_2P_1h=0,
\]
and hence
\[
    (G_1+G_2P_1)h=0.
\]
The identity morphisms also satisfy the required
kernel inclusion, which proves the first assertion.

For the second, put
\[
    a=\langle h,W_1h\rangle,
    \qquad
    b=\langle P_1h,W_2P_1h\rangle.
\]
The hypotheses and the triangle inequality give
\[
\begin{aligned}
    \norm{C^{1/2}(G_1+G_2P_1)h}
    &\leq
    \sqrt{\beta_1a}+\sqrt{\beta_2b}\\
    &\leq
    \sqrt{(\beta_1+\beta_2)(a+b)},
\end{aligned}
\]
where the last inequality is Cauchy--Schwarz.
Squaring gives \eqref{eq:composite-domination}.
\end{proof}

For the constants defined in \eqref{eq:beta},
one has
\[
    \beta_\mu(s,t)
    =
    \beta_\mu(s,r)+\beta_\mu(r,t),
\]
since their integrand depends only on the
instantaneous evaluation Jacobian, the preconditioner,
and the evaluation law. The domination estimate is
therefore compatible with the composition of
training intervals.

\subsection{Observability}
\label{sec:observability-structure}

The dissipation operator also admits an interpretation
in the theory of linear systems. Along the realized
trajectory, consider the time-dependent system
\[
    \dot h(t)=A(t)h(t),
    \qquad
    y(t)=D(t)h(t),
\]
where
\[
    A(t)=-B(t)K_{SS}^M(t),
    \qquad
    D(t)=M(t)^{1/2}J_S(t)^*.
\]
Here \(y(t)\) denotes the observed output of the
linear system, rather than the training target.
Its state-transition operator is \(P_{ts}\).

For a fixed interval \([s,T]\), define the
observation operator
\[
    \mathcal O_{sT}:V
    \longrightarrow L^2([s,T];\HP),
\]
by
\[
    (\mathcal O_{sT}h)(t)=D(t)P_{ts}h.
\]
It follows from the definition of \(W_{sT}\) that
\begin{equation}
\label{eq:observability-Gramian}
    W_{sT}=\mathcal O_{sT}^*\mathcal O_{sT},
    \qquad
    \Ker W_{sT}=\Ker\mathcal O_{sT}.
\end{equation}
Thus \(W_{sT}\) is the observability Gramian of
the displayed system, in the usual sense of
linear control theory~\cite{davis2009systems}.

In particular, the reservoir consists precisely
of those initial states whose dissipative observation
vanishes almost everywhere on the interval.

The transfer operator factors through the entire
observation history. To see this, define
\[
    \mathcal L_{sT}:L^2([s,T];\HP)\longrightarrow E
\]
by
\[
    \mathcal L_{sT}v
    =
    \int_s^T J_E(t)M(t)^{1/2}v(t)\,\dd t.
\]
One then has
\begin{equation}
\label{eq:history-transfer-factorization}
    G_{Ts}
    =
    \mathcal L_{sT}\mathcal O_{sT},
    \qquad
    \norm{C_\mu^{1/2}\mathcal L_{sT}}_{\op}^2
    \leq
    \beta_\mu(s,T).
\end{equation}
Indeed, the first assertion follows from
\[
\begin{aligned}
    \mathcal L_{sT}\mathcal O_{sT}h
    &=
    \int_s^T
    J_E(t)M(t)J_S(t)^*P_{ts}h\,\dd t\\
    &=
    G_{Ts}h,
\end{aligned}
\]
while the second follows from the Cauchy--Schwarz
inequality.

Combining \eqref{eq:observability-Gramian} and
\eqref{eq:history-transfer-factorization} gives
another proof of reservoir invisibility and
dissipation domination.

\begin{remark}
The inclusion
\[
    \Ker\mathcal O_{sT}\subseteq\Ker G_{Ts}
\]
may be strict. Indeed, a nonzero observation history
may belong to the kernel of \(\mathcal L_{sT}\),
owing either to the evaluation map or to cancellation
in the time integral.

Thus observability through the dissipative signal
does not imply nonzero accumulated evaluation
transfer. Moreover, as shown in \cref{sec:risk},
nonzero evaluation transfer does not necessarily
imply a decrease in population risk.
\end{remark}

\subsection{The Grassmannian of signal subspaces}
\label{sec:grassmann-structure}

Put \(N=\dim V\), fix \(\alpha>0\), and denote
by
\[
    \Pi_\alpha(T)
    =
    \mathbf1_{(\alpha,\infty)}(W_{sT}),
\]
the spectral projector denoted by \(P_\alpha(T)\)
in \cref{sec:spectral-transport}.

On an interval where \(\Pi_\alpha(T)\) is
continuously differentiable and has constant rank
\(r\), the map
\[
    T\longmapsto\ran\Pi_\alpha(T),
\]
is a curve in the real Grassmannian
\[
    \operatorname{Gr}(r,N)
    \simeq
    O(N)/(O(r)\times O(N-r)).
\]
This description is independent of the choice of
an orthonormal basis of the signal subspace.
We use below the standard realization of the
Grassmannian by orthogonal projectors
\cite{bendokat2024grassmann}.

\begin{proposition}
\label{prop:signal-bundle-transport}
Let \(\Pi(T)=\Pi_\alpha(T)\) be continuously
differentiable and of constant rank on an
interval containing \(T_0\). The spaces
\(\ran\Pi(T)\) form a continuously differentiable
vector subbundle of the trivial bundle with
fiber \(V\).

This subbundle carries a metric connection given
along the interval by
\begin{equation}
\label{eq:signal-bundle-connection}
    \nabla_Tv=\Pi(T)\dot v(T),
    \qquad
    v(T)\in\ran\Pi(T).
\end{equation}
Put
\[
    \mathcal A(T)=[\dot\Pi(T),\Pi(T)],
\]
and let \(U(T)\) be the solution of
\begin{equation}
\label{eq:orthogonal-signal-transport}
    \dot U(T)=\mathcal A(T)U(T),
    \qquad
    U(T_0)=I.
\end{equation}
Then \(U(T)\) is orthogonal and satisfies
\[
    \Pi(T)=U(T)\Pi(T_0)U(T)^*.
\]
For every \(v_0\in\ran\Pi(T_0)\), the section
\(v(T)=U(T)v_0\) is parallel with respect to
the connection \eqref{eq:signal-bundle-connection}.
\end{proposition}

\begin{proof}
Differentiating \(\Pi^2=\Pi\), one obtains
\[
    \dot\Pi\Pi+\Pi\dot\Pi=\dot\Pi,
    \qquad
    \Pi\dot\Pi\Pi=0.
\]
It follows that
\[
    [[\dot\Pi,\Pi],\Pi]=\dot\Pi,
\]
or, equivalently, \(\dot\Pi=[\mathcal A,\Pi]\).

Since \(\Pi\) and \(\dot\Pi\) are self-adjoint,
\(\mathcal A\) is skew-adjoint. Hence
\[
\begin{aligned}
    \frac{\dd}{\dd T}(U^*U)
    &=
    U^*(\mathcal A^*+\mathcal A)U\\
    &=0,
\end{aligned}
\]
and the initial condition gives \(U(T)^*U(T)=I\).

Furthermore,
\[
\begin{aligned}
    \frac{\dd}{\dd T}(U^*\Pi U)
    &=
    U^*(\dot\Pi-[\mathcal A,\Pi])U\\
    &=0,
\end{aligned}
\]
so that
\[
    U(T)^*\Pi(T)U(T)=\Pi(T_0).
\]
This proves the asserted relation between the
projectors. Transporting an orthonormal basis of
\(\ran\Pi(T_0)\) by \(U(T)\) gives a continuously
differentiable orthonormal frame of the subbundle.

If \(v(T)=U(T)v_0\), with
\(v_0\in\ran\Pi(T_0)\), then \(\Pi(T)v(T)=v(T)\).
Consequently,
\[
    \nabla_Tv
    =
    \Pi\dot v
    =
    \Pi\mathcal A\Pi v
    =0,
\]
which proves the parallelism assertion.

Finally, for two differentiable sections \(v\)
and \(w\), one has
\[
\begin{aligned}
    \frac{\dd}{\dd T}\langle v,w\rangle
    &=
    \langle\dot v,w\rangle
    +\langle v,\dot w\rangle\\
    &=
    \langle\nabla_Tv,w\rangle
    +\langle v,\nabla_Tw\rangle,
\end{aligned}
\]
since \(\Pi v=v\) and \(\Pi w=w\).
Thus the connection is compatible with the metric.
\end{proof}

The orthogonal transport in
\eqref{eq:orthogonal-signal-transport} is the
classical construction of Kato~\cite{kato1995}.
The orthonormal \(r\)-frames form a principal
\(O(r)\)-bundle over \(\operatorname{Gr}(r,N)\),
and the group \(O(r)\) describes the freedom in
the choice of a frame of the signal subspace.

This freedom is distinct from a symmetry of the
training problem, which acts on the ambient
training-output space and will be considered below.

The derivative of the spectral projector in
\cref{thm:projector-transport} may therefore be
regarded as a tangent vector to the Grassmannian.
This description applies on intervals of constant
spectral rank. When an eigenvalue passes above the
threshold, the signal subspace belongs to a
Grassmannian of different dimension.

The construction along a single time interval
does not by itself give rise to a nontrivial
curvature or holonomy theory. Such questions
would require a family of signal subspaces
depending on additional parameters.

\subsection{The filtration by exact signal spaces}
\label{sec:signal-persistence}

The exact signal spaces have a simpler structure
than those associated with a positive spectral
threshold. For fixed \(s\), put
\[
    \mathcal S_T
    =
    \ran W_{sT}
    =
    (\Ker W_{sT})^\perp.
\]

\begin{proposition}
\label{prop:exact-signal-persistence}
For \(s\leq T_1\leq T_2\), one has
\[
    \mathcal S_{T_1}\subseteq\mathcal S_{T_2}.
\]
With the inclusion maps, the assignment
\(T\mapsto\mathcal S_T\) defines a persistence
module indexed by the ordered set of training
times.

On every interval on which \(\rank W_{sT}\) is
constant, the signal space \(\mathcal S_T\) is
constant.

Moreover, the reservoir quotients admit canonical
surjections in the opposite direction,
\[
    V/\Ker W_{sT_2}
    \longrightarrow
    V/\Ker W_{sT_1},
    \qquad
    [h]\longmapsto[h].
\]
\end{proposition}

\begin{proof}
By \cref{prop:nested-reservoir}, one has
\[
    \Ker W_{sT_2}\subseteq\Ker W_{sT_1}.
\]
Taking orthogonal complements gives the inclusion
of signal spaces. The functorial properties follow
from those of the inclusion maps.

Since nested finite-dimensional subspaces of equal
dimension coincide, the signal space is constant
whenever the rank of \(W_{sT}\) is constant.
Finally, the same inclusion of kernels gives
the asserted surjection between the quotients.
\end{proof}

The conclusion does not extend to the signal
spaces defined by a positive threshold, even
though their dimensions are nondecreasing.

For instance, consider
\[
    W_1=
    \begin{pmatrix}
        2&0\\
        0&0
    \end{pmatrix},
    \qquad
    W_2=
    \begin{pmatrix}
        3&1\\
        1&1
    \end{pmatrix}.
\]
One has
\[
    W_2-W_1=
    \begin{pmatrix}
        1&1\\
        1&1
    \end{pmatrix}
    \succeq0.
\]
For \(\alpha=1\), however, the corresponding
signal spaces are
\[
    \ran\mathbf1_{(1,\infty)}(W_1)
    =
    \R(1,0),
\]
and
\[
    \ran\mathbf1_{(1,\infty)}(W_2)
    =
    \R(1,\sqrt2-1),
\]
since the eigenvalues of \(W_2\) are
\(2+\sqrt2\) and \(2-\sqrt2\).
These two spaces have the same dimension but
are distinct, so neither contains the other.

Thus the exact signal spaces form a filtration
of \(V\), whereas the effective signal spaces
need not form one. The latter may vary within
a fixed Grassmannian, as in the spectral transport
mechanism of \cref{cor:two-mechanisms}.

The persistence module records the appearance
of exact signal directions, but it does not
determine the transfer operators \(G_{Ts}\).
In particular, no compatibility of these operators
with the inclusion maps of the persistence module
has been assumed.

\subsection{Symmetries of the training dynamics}
\label{sec:task-symmetry}

We now suppose that a finite group \(\Gamma\)
acts orthogonally on \(V\) and \(E\), through
representations denoted by \(U_\gamma\) and
\(V_\gamma\), respectively.

Assume that, for every \(\gamma\in\Gamma\) and
almost every \(t\), one has
\begin{align}
    B(t)U_\gamma
    &=
    U_\gamma B(t),
    &
    K_{SS}^M(t)U_\gamma
    &=
    U_\gamma K_{SS}^M(t),
    \label{eq:symmetry-training-assumptions}\\
    K_{ES}^M(t)U_\gamma
    &=
    V_\gamma K_{ES}^M(t),
    &
    V_\gamma^*C_\mu V_\gamma
    &=
    C_\mu.
    \label{eq:symmetry-evaluation-assumptions}
\end{align}
Under these hypotheses, the operators introduced
above inherit the group action.

\begin{proposition}
\label{prop:equivariant-visibility}
Under \eqref{eq:symmetry-training-assumptions} and
\eqref{eq:symmetry-evaluation-assumptions}, one has
\[
    P_{ts}U_\gamma=U_\gamma P_{ts},
    \qquad
    W_{st}U_\gamma=U_\gamma W_{st},
\]
and
\[
    G_{ts}U_\gamma=V_\gamma G_{ts}.
\]
For every signal filter \(\mathsf F\) obtained
by applying the spectral calculus to \(W_{st}\),
the corresponding visibility operator satisfies
\[
    H_{\mu,\mathsf F}U_\gamma
    =
    U_\gamma H_{\mu,\mathsf F}.
\]
\end{proposition}

\begin{proof}
By \eqref{eq:symmetry-training-assumptions}, the
generator \(-B(t)K_{SS}^M(t)\) commutes with
\(U_\gamma\). The same is therefore true of the
propagator, by uniqueness of the solution of
its initial-value problem.

Since \(U_\gamma\) is orthogonal, one obtains
\[
\begin{aligned}
    U_\gamma^*W_{st}U_\gamma
    &=
    \int_s^t
    P_{\tau s}^*U_\gamma^*
    K_{SS}^M(\tau)U_\gamma
    P_{\tau s}\,\dd\tau\\
    &=
    W_{st},
\end{aligned}
\]
which proves the commutation relation for \(W_{st}\).

Similarly, the first identity in
\eqref{eq:symmetry-evaluation-assumptions} gives
\[
\begin{aligned}
    G_{ts}U_\gamma
    &=
    \int_s^t
    K_{ES}^M(\tau)U_\gamma
    P_{\tau s}\,\dd\tau\\
    &=
    V_\gamma G_{ts}.
\end{aligned}
\]

By spectral calculus, \(\mathsf F\) commutes
with \(U_\gamma\). Consequently,
\[
\begin{aligned}
    U_\gamma^*H_{\mu,\mathsf F}U_\gamma
    &=
    \mathsf F G_{ts}^*V_\gamma^*
    C_\mu V_\gamma G_{ts}\mathsf F\\
    &=
    H_{\mu,\mathsf F},
\end{aligned}
\]
where the last equality follows from the invariance
of the evaluation form. Since \(U_\gamma\) is
orthogonal, this is equivalent to the desired
commutation relation.
\end{proof}

We pass to the complexification \(V_\C\) of \(V\).
By complete reducibility, it admits an orthogonal
isotypic decomposition
\[
    V_\C
    \simeq
    \bigoplus_\lambda
    V_\lambda\otimes\C^{m_\lambda},
\]
where the \(V_\lambda\) are pairwise nonisomorphic
irreducible complex representations of \(\Gamma\)
and \(m_\lambda\) is the multiplicity of
\(V_\lambda\).

By Schur's lemma~\cite{telemanrepresentation},
every endomorphism commuting with the action of
\(\Gamma\) preserves the isotypic components
and acts on each of them through its multiplicity
space. Applied to the operators of
\cref{prop:equivariant-visibility}, this gives
\begin{equation}
\label{eq:isotypic-visibility-decomposition}
\begin{aligned}
    W
    &\simeq
    \bigoplus_\lambda
    I_{V_\lambda}\otimes W_\lambda,\\
    H_{\mu,\mathsf F}
    &\simeq
    \bigoplus_\lambda
    I_{V_\lambda}\otimes H_{\mu,\lambda},
\end{aligned}
\end{equation}
where \(W_\lambda\) and \(H_{\mu,\lambda}\)
are positive semidefinite Hermitian operators on
the corresponding multiplicity spaces.

Thus the spectral analysis of the dissipation
and visibility operators reduces to that of their
multiplicity-space components.

\begin{corollary}
\label{cor:symmetry-task-leakage}
Suppose that the complexification \(\Task_\C\)
of the task subspace is a sum of whole isotypic
components of \(V_\C\). Then
\[
    \Pi_\Task H_{\mu,\mathsf F}\Pi_\Task^\perp
    =
    0.
\]
In particular, the coupling parameter \(\kappa\)
of \cref{thm:block-task} vanishes.

The mere invariance of \(\Task\) under \(\Gamma\)
does not in general imply this conclusion.
\end{corollary}

\begin{proof}
By \eqref{eq:isotypic-visibility-decomposition},
\(H_{\mu,\mathsf F}\) preserves each isotypic
component and therefore any sum of such components,
as well as its orthogonal complement.

On the other hand, an invariant subspace of an
isotypic component may have the form
\[
    V_\lambda\otimes L,
    \qquad
    L\subseteq\C^{m_\lambda},
\]
where \(L\) need not be invariant under
\(H_{\mu,\lambda}\). Thus invariance under
\(\Gamma\) does not suffice when irreducible
representations occur with multiplicity.
\end{proof}

\begin{remark}
If the complex representation is multiplicity free,
each \(m_\lambda\) is equal to one. In this case,
the operators \(W\) and \(H_{\mu,\mathsf F}\) act
by scalars on each irreducible component.

For a finite abelian group, the complex irreducible
representations are characters. In particular,
cyclic and product-cyclic symmetries give a
decomposition into Fourier-character components.
The corresponding visibility operators are scalar
on these components when the representation is
multiplicity free, and are matrix-valued otherwise.

Such a decomposition may be useful in studying
the visibility of individual algebraic components
of a task.
\end{remark}

It is important that the symmetry assumptions
concern the operators along the realized
trajectory, rather than merely the target function.
A symmetric target need not give rise to symmetric
training dynamics, since the training sample or
the initialization may break the symmetry.

The representation-theoretic reduction above is
therefore conditional upon
\eqref{eq:symmetry-training-assumptions} and
\eqref{eq:symmetry-evaluation-assumptions}.

\subsection{Rank loci and spectral degeneracies}
\label{sec:determinantal-geometry}

The dissipation operator \(W_{sT}\) belongs to
the cone \(\operatorname{Sym}_N^+(\R)\) of positive
semidefinite real symmetric matrices. We shall
describe the subsets of the ambient matrix space
on which its rank, or the dimension of a spectral
subspace, may change.

For \(0\leq r<N\), put
\[
    \mathcal D_r
    =
    \left\{
        A\in\operatorname{Sym}_N(\R):
        \rank A\leq r
    \right\}.
\]
This is a real algebraic subset of
\(\operatorname{Sym}_N(\R)\), defined by the
vanishing of all minors of order \(r+1\).
The sets \(\mathcal D_r\) form an increasing
family
\[
    \mathcal D_0
    \subseteq
    \mathcal D_1
    \subseteq
    \cdots
    \subseteq
    \mathcal D_{N-1}.
\]
The set of symmetric matrices of rank exactly
\(r\) is the difference
\[
    \mathcal D_r\setminus\mathcal D_{r-1},
\]
with the convention \(\mathcal D_{-1}=\varnothing\)
and \(\mathcal D_N=\operatorname{Sym}_N(\R)\).

Since positive semidefiniteness is equivalent
to the nonnegativity of all principal minors,
the intersection
\[
    \mathcal D_r\cap\operatorname{Sym}_N^+(\R),
\]
is semialgebraic. These are the usual determinantal
rank conditions~\cite[Section~1.C]{bruns1988determinantal}.

For a positive spectral threshold \(\alpha\),
the relevant algebraic subset is instead
\begin{equation}
\label{eq:algebraic-threshold-locus}
    \Sigma_\alpha
    =
    \left\{
        A\in\operatorname{Sym}_N(\R):
        \det(A-\alpha I)=0
    \right\}.
\end{equation}
On the complement of \(\Sigma_\alpha\), the
number of eigenvalues greater than \(\alpha\)
is locally constant. Moreover, the corresponding
spectral projector depends analytically on \(A\).

Indeed, near any point outside \(\Sigma_\alpha\),
one may choose a fixed contour \(\mathcal C\)
separating the eigenvalues above \(\alpha\)
from the remaining eigenvalues, and write
\[
    \Pi_\alpha(A)
    =
    \frac{1}{2\pi i}
    \oint_{\mathcal C}(zI-A)^{-1}\,\dd z.
\]
The analyticity of the projector follows from
that of the resolvent.

The Grassmannian itself admits the real algebraic
realization
\[
\begin{aligned}
    \operatorname{Gr}(r,N)
    \simeq
    \bigl\{
        \Pi\in\operatorname{Mat}_N(\R):
        \Pi^\top=\Pi,\,
        \Pi^2=\Pi,\,
        \tr\Pi=r
    \bigr\}.
\end{aligned}
\]
Thus, away from the threshold hypersurface,
the effective signal space varies in a fixed
Grassmannian. A change in its dimension can
occur only at the threshold locus.

\begin{remark}
Under additional algebraic hypotheses, the
determinantal description may also be applied
in parameter space.

Suppose that \(\theta\in\R^d\), that the outputs
on a fixed finite collection of training and
evaluation points are polynomial functions of
\(\theta\), and that the entries of the
preconditioner \(M\) are polynomial in \(\theta\).
This includes the case of a constant
preconditioner.

The corresponding Jacobian and kernel matrices
then have polynomial entries. Their rank loci
are therefore algebraic subsets of \(\R^d\),
obtained by pulling back the determinantal
equations along the polynomial matrix maps.

For a square symmetric kernel matrix \(K(\theta)\),
the locus on which \(\alpha\) is an eigenvalue
is similarly given by
\[
    \det(K(\theta)-\alpha I)=0.
\]

These assertions concern the instantaneous
kernel matrices. They do not imply that the
finite-window operator \(W_{sT}\) is polynomial
in \(T\) or in the initial parameters, since its
definition involves a fundamental solution and
an integration along the trajectory.

Even for a constant kernel and squared loss,
one has
\[
    W_{sT}
    =
    \frac12(I-e^{-2(T-s)K}),
\]
which is generally not polynomial in \(T\).

The determinantal loci thus describe the possible
rank and spectral degeneracies in the ambient
matrix space. The training equations determine
whether, and in what manner, the realized
trajectory encounters these loci.
\end{remark}

\subsection{Concluding remarks}

The composition law for training intervals and
the quotient by the reservoir describe two
aspects of the same system. The former specifies
how the transition, transfer, and dissipation
operators combine, while the latter identifies
the directions on which the accumulated
dissipation defines a positive definite form.
The observability interpretation provides a
description of the information retained by this
quotient.

There is a corresponding distinction on the
geometric side. The exact signal spaces form
a filtration of the training-output space,
whereas the effective signal spaces may vary
within a Grassmannian. Symmetries of the
realized dynamics may restrict this variation
to representation-theoretic components and,
in particular, eliminate certain couplings
between the task and its complement.

The algebraic rank and threshold loci describe
the subsets of the ambient matrix space at
which these structures may change. Their
description does not, however, determine the
motion of the operators along the training
trajectory.

None of these constructions replaces the
evaluation transfer, the target residual,
or the signed risk identity. Their purpose
is to describe the composition, invariance,
and degeneracy properties of the operators
on which the preceding estimates depend.

\clearpage

\section{Consequences for the training dynamics}
\label{sec:predictions}

We collect several consequences of the preceding
results that may be examined along a training
trajectory. We distinguish the roles of mobility,
transfer efficiency, and target orientation in
the evolution of population risk, and examine
the contributions of spectral activation and
spectral transport to changes in evaluation
visibility. We also discuss the effects of task
alignment, stochastic fluctuations, and the
distribution of margin crossing times. Finally,
we consider the dependence of generalization on
the evaluation law. These observations identify
quantities that may be measured during training
to distinguish the mechanisms responsible for
delayed, abrupt, and selective generalization.

For a nonzero direction \(h\) outside
the reservoir, the quantities
\[
    \Mob(h),
    \qquad
    \Eff_\mu(h),
    \qquad
    \frac{\norm{Gh}_{C_\mu}^2}{\norm h^2},
\]
describe, respectively, its mobility, its transfer
efficiency, and its normalized evaluation visibility.
By \cref{cor:factorization}, the third is the product
of the first two. When \(h=g(s)\), the signed quantity
\[
    \Prog_\mu(s,T)
    =
    \Risk_\mu(s)-\Risk_\mu(T),
\]
determines the actual variation of population risk.
The distinction between these quantities leads to
the following observations.

\medskip

\noindent
\textit{1. Mobility, transfer, and orientation.}
A small variation of population risk may have
several causes. By \eqref{eq:two-gate}, a small
evaluation displacement may result either from
small mobility or from small transfer efficiency.
Even when neither quantity is small, the displacement
may be poorly oriented relative to the residual,
or its magnitude may be too large for the given
orientation, as in \cref{cor:angle}.
These possibilities should be distinguished in
the study of delayed generalization.

\medskip

\noindent
\textit{2. Visibility and variation of risk.}
An increase in evaluation visibility need not be
accompanied by a decrease in population risk.
Indeed, by \eqref{eq:signed-progress}, the sign
of the risk variation is determined by the
difference
\[
    \ip{e(s)}{C_\mu d(s,T)}
    -
    \frac12\norm{d(s,T)}_{C_\mu}^2,
\]
where \(d(s,T)=G(T,s)g(s)\).
It is therefore necessary to compare the inner
product with the residual to the squared norm
of the displacement, rather than to consider
either quantity separately.

\medskip

\noindent
\textit{3. Variation of the signal filter.}
The two terms in \eqref{eq:energy-balance}
distinguish the contribution of the variation
of \(G\) from that of the variation of the
signal filter. Their respective values determine
whether an increase in filtered visibility
is principally due to the accumulation of
evaluation transfer or to a change in the
spectral weights assigned to the training
directions.

Since both contributions may have either sign,
their magnitudes alone do not determine the
variation of visibility.

\medskip

\noindent
\textit{4. Activation and spectral transport.}
By \cref{thm:projector-transport}, the effective
signal projector \(P_\alpha(T)\) is stationary
precisely when
\[
    (\Id-P_\alpha(T))Q(T;s)P_\alpha(T)=0,
\]
at points where the spectral rank is locally
constant.

If this condition holds throughout an interval
of constant spectral rank, the signal subspace
is fixed. In particular, if \(W(T)\) and \(Q(T;s)\)
commute throughout the training interval, changes
in the effective signal channel can occur only
through eigenvalue activation at the threshold.

\medskip

\noindent
\textit{5. The regularity of classification accuracy.}
The time scale on which population risk decreases
need not coincide with that on which classification
accuracy increases. By \cref{prop:accuracy-cdf},
the latter is related to the distribution of the
examplewise margin crossing times.

The construction of
\cref{thm:two-mode-classification} shows that
differences in spectral learning rates may produce
both an arbitrarily long delay before generalization
and a concentration of the crossing times on an
arbitrarily short interval. Moreover, by
\cref{cor:two-mode-classification-risk}, the
variation of population risk over this interval
may tend to zero while accuracy increases from
zero to one.

The observation of a rapid improvement in
classification accuracy does not therefore suffice
to establish a rapid change in the learned
representation or in the underlying training
dynamics.

\medskip

\noindent
\textit{6. The effect of stochasticity.}
For a fixed residual, mean gradient, and
evaluation transfer, the contribution of gradient
covariance to the conditional expected population
risk is given by
\[
    \frac{\eta^2}{2}
    \tr(A_k^*C_\mu A_k\Sigma_k).
\]
By \cref{prop:noise-placement}, covariance
concentrated in directions associated with large
eigenvalues of \(A_k^*C_\mu A_k\) has a larger
instantaneous cost than covariance concentrated
in directions associated with small eigenvalues.

A comparison of stochastic optimization methods
should therefore distinguish this contribution
from the changes in the transfer operators and
mean gradients induced by subsequent training.

\medskip

\noindent
\textit{7. Task alignment and coupling.}
An increase in the alignment
\(\rho_\Task(h)\) does not in general imply
an increase in evaluation visibility. By
\cref{thm:block-task}, the visibility of a
direction depends also on the extremal spectra
of the task and complementary blocks, and on
the coupling between them.

It is therefore natural to examine the quantities
\[
    b,\quad a,\quad d,\quad c,\quad\kappa,
    \quad\rho_\Task(h),
\]
at successive times along the training trajectory.
Together, they give upper and lower bounds
for the visibility of the direction under
consideration.

\medskip

\noindent
\textit{8. Selective generalization.}
The same training trajectory may give rise to
different generalization times under different
evaluation laws. By \cref{thm:explicit-mu-not-nu},
the conditions
\[
    \Risk_\mu(s)-p_\mu
    \leq
    \frac{c_\mu\gamma_\mu^2\delta}{4},
    \qquad
    m_\nu
    -
    \frac{2a_\nu\norm{g_s}^2}{\gamma_\nu^2}
    >\delta,
\]
are sufficient to ensure that
\[
    A_\mu(T)\geq1-\delta,
    \qquad
    A_\nu(T)<1-\delta.
\]
The first condition concerns the amount of
signed progress under \(\mu\), whereas the
second concerns the evaluation displacement
required to correct the initially negative
margins under \(\nu\).

\medskip

These observations do not provide a universal
prediction of generalization times from the
training loss or the training spectrum alone.
Rather, they identify quantities whose joint
evolution determines, or bounds, the variation
of evaluation visibility and population risk.

Their measurement along a training trajectory
would allow one to distinguish the effects of
small eigenvalues, spectral transport, evaluation
transfer, and target orientation, even when the
corresponding training and accuracy curves are
similar.

\clearpage

\section{A matrix-free empirical protocol}
\label{sec:protocol}

We describe a computational procedure for estimating
the quantities introduced in the preceding sections
without explicitly constructing the associated
operators. We first consider the computation of
the actions of the dissipation and transfer operators
and the application of the resolvent filter. We then
examine how to estimate task visibility, spectral
gaps, and coupling between task directions and their
orthogonal complement. Finally, we describe how
to measure the evolution of evaluation visibility
and signed population-risk reduction under different
evaluation laws. The resulting protocol provides
a means of examining the theoretical mechanisms
of delayed and selective generalization along a
realized training trajectory.

Litman and Guo~\cite[Section~E.8,
Proposition~E.22 and Corollary~E.23]{litmanguo2026}
show that, for finite training and evaluation sets,
the actions of the dissipation and transfer operators
and of their adjoints may be computed by forward
and backward linear differential equations along
the realized training trajectory. We use the same
construction under the operator regularity assumptions
of \cref{sec:setting}, with particular attention to
the quantities associated with task subspaces,
resolvent filtering, and general evaluation laws.

Fix an interval \([s,T]\), an evaluation law \(\mu\),
a task subspace \(\Task\), and a nonzero direction
\(h\in\HS\).
 We describe below a procedure for
estimating the quantities that enter the preceding
results.

\medskip

\noindent
\textit{1. Computation of the dissipation operator.}
By the definition of \(W=\W(s,T)\), one has
\[
    Wh
    =
    \int_s^T
    \Pg(\tau,s)^*
    J_S(\tau)M(\tau)J_S(\tau)^*
    \Pg(\tau,s)h\,\dd\tau.
\]
The action of \(W\) may therefore be approximated
by integrating the gradient propagation equation
and its adjoint, together with the corresponding
Jacobian products. The entries of \(W\) need not
be computed individually.

\medskip

\noindent
\textit{2. Computation of the filtered direction.}
For \(\alpha>0\), one has
\[
    S_\alpha h
    =
    h-\alpha(W+\alpha\Id)^{-1}h.
\]
Since \(W+\alpha\Id\) is positive definite, the
action of its inverse may be computed by the
conjugate gradient method, using only products
with \(W\). Thus the filtered direction may be
obtained without computing the spectral
decomposition of \(W\).

More precisely, if \(v\) approximates the solution
of \((W+\alpha\Id)v=h\), and
\[
    r_\alpha=h-(W+\alpha\Id)v,
\]
then
\[
    \norm{S_\alpha h-(h-\alpha v)}
    \leq\norm{r_\alpha}.
\]
Indeed, this follows from
\(\norm{(W+\alpha\Id)^{-1}}_{\op}\leq1/\alpha\).
The residual of the linear system therefore
provides an error estimate for the computed
filtered direction.

\medskip

\noindent
\textit{3. Computation of the task spectrum.}
Suppose that the task subspace is specified
either by an orthonormal basis or by the action
of its orthogonal projector \(\Pi_\Task\).
The visibility operator is given by
\[
    H_{\mu,\alpha}
    =
    S_\alpha G^*C_\mu G S_\alpha.
\]
Its action on a vector requires only successive
applications of the filter, the transfer operator,
the evaluation metric, and the adjoint transfer.

The extremal eigenvalues of its compressions
to \(\Task\) and \(\Task^\perp\) may be estimated
by the Lanczos method. The coupling coefficient
\[
    \kappa
    =
    \norm{
        \Pi_\Task H_{\mu,\alpha}\Pi_\Task^\perp
    }_{\op},
\]
may likewise be estimated by iteration on the
corresponding product with its adjoint.

The quantities so obtained determine the two
bounds of \cref{thm:block-task}. Both endpoints
of the interval in
\eqref{eq:block-lower}--\eqref{eq:block-upper}
should be reported. If the interval is to be used
as a rigorous bound, the numerical errors in
the spectral estimates must also be controlled.

\medskip

\noindent
\textit{4. Computation of the evaluation transfer.}
For a direction \(v\in\HS\), one has
\[
    Gv
    =
    \int_s^T
    J_E(\tau)M(\tau)J_S(\tau)^*
    \Pg(\tau,s)v\,\dd\tau.
\]
Taking \(v=S_\alpha h\) gives the filtered
evaluation displacement. Taking \(v=h\) gives
the corresponding unfiltered displacement.

The evaluation Jacobian may be computed on an
independent sample from \(\mu\). Alternatively,
a suitable cross-fitting procedure may be used,
provided that the relation between the fitted
trajectories and the evaluation law is taken
into account.

\medskip

\noindent
\textit{5. Mobility, visibility, and signed progress.}
The quantities
\[
    \Mob(h),\qquad
    \Eff_\mu(h),\qquad
    \rho_\Task(h),\qquad
    \norm{Gh}_{C_\mu}^2,
\]
should be recorded separately, with the transfer
efficiency defined whenever \(h\notin\Ker W\).
For such a direction, the identity
\[
    \Mob(h)\Eff_\mu(h)
    =
    \frac{\norm{Gh}_{C_\mu}^2}{\norm h^2},
\]
provides a consistency check on their computation.

For the actual gradient \(h=g(s)\), the signed
variation of population risk is given by
\[
    \Prog_\mu(s,T)
    =
    \ip{e(s)}{C_\mu Gh}
    -
    \frac12\norm{Gh}_{C_\mu}^2.
\]
This quantity should be distinguished from the
unsigned visibility, whether filtered or unfiltered.

\medskip

\noindent
\textit{6. Variation of filtered visibility.}
Keeping \(s\) and \(h\) fixed, let
\[
    d_\alpha(T)=G(T,s)S_\alpha(T)h,
    \qquad
    R_\alpha(T)=(W(T)+\alpha\Id)^{-1}.
\]
By \eqref{eq:d-balance}, one has
\[
    \dot d_\alpha(T)
    =
    L(T;s)S_\alpha(T)h
    +
    \alpha G(T,s)R_\alpha(T)
    Q(T;s)R_\alpha(T)h.
\]
The two terms may be estimated at successive
checkpoints to distinguish the contribution
of the variation of the transfer operator
from that of the signal filter.

Integrating between two checkpoints gives
\[
\begin{aligned}
    d_\alpha(T_2)-d_\alpha(T_1)
    &=
    \int_{T_1}^{T_2}
    L(\tau;s)S_\alpha(\tau)h\,\dd\tau\\
    &\quad+
    \alpha\int_{T_1}^{T_2}
    G(\tau,s)R_\alpha(\tau)
    Q(\tau;s)R_\alpha(\tau)h\,\dd\tau.
\end{aligned}
\]
Comparison with the observed displacement gives
a check on the numerical approximation of the
two contributions.

\medskip

\noindent
\textit{7. Risk and accuracy crossing times.}
The first time at which the signed progress is
at least \(\Risk_\mu(s)-\delta\) coincides, by
\eqref{eq:progress-clock}, with the first time
at which the population risk is at most
\(\delta\).

This equality may be used to check the
reconstruction of the risk from the estimated
transfer operator. In a classification problem,
one may further compare the observed error
with the upper bound
\[
    \Err_\mu(T)
    \leq
    \frac{4\Risk_\mu(T)}{c_\mu\gamma_\mu^2},
\]
when the margin hypothesis of
\cref{thm:classification} is satisfied.

The distribution of the examplewise margin
crossing times may also be estimated. In the
case of nonmonotone margins, one must distinguish
the first crossing time from subsequent crossings
of the decision boundary.

Let \(0<\eta<1/2\), and denote by \(t_\eta\)
and \(t_{1-\eta}\) the first times after
\(t_{\rm mem}(\varepsilon)\) at which classification
accuracy is at least \(\eta\) and \(1-\eta\),
respectively. When both times are finite, one
may estimate the quantities
\[
    D_\eta
    =
    t_\eta-t_{\rm mem}(\varepsilon),
    \qquad
    w_\eta
    =
    t_{1-\eta}-t_\eta,
\]
together with the variation of population risk
\[
    \Delta\Risk_{\mu,\eta}
    =
    \Risk_\mu(t_\eta)-\Risk_\mu(t_{1-\eta}).
\]
For \(D_\eta>0\), the ratio \(w_\eta/D_\eta\)
measures the duration of the accuracy improvement
relative to the preceding delay.

The results of \cref{sec:abrupt} suggest examining
these quantities jointly with the spectral scales
of the training kernel and the concentration of
the distribution of margin crossing times.

\medskip

\noindent
\textit{8. Comparison of evaluation laws.}
For two evaluation laws \(\mu\) and \(\nu\), the
preceding procedure may be carried out with the
same training trajectory and the same signal
filter.

In addition to the corresponding task spectra,
one estimates the signed progress under \(\mu\)
and the measure \(m_\nu\) of examples with
an initially incorrect margin of magnitude at
least \(\gamma_\nu\). The conditions to be examined
are those of \eqref{eq:selective-conditions}.

In particular, the proof of
\cref{thm:explicit-mu-not-nu} gives the estimate
\[
    \nu(B_\nu')
    \leq
    \frac{2\norm{Gg(s)}_{C_\nu}^2}{\gamma_\nu^2}
    \leq
    \frac{2a_\nu\norm{g(s)}^2}{\gamma_\nu^2},
\]
where \(B_\nu'\) is the subset of the initially
misclassified examples in \(B_\nu\) which are
correctly classified at time \(T\).

Thus the observed evaluation displacement gives
an upper bound for the measure of the set of
initial classification errors that can have
been corrected over the interval.

\medskip

The estimates obtained from a finite evaluation
sample concern its empirical measure. Their
extension to the population law requires a
separate statistical argument.

By \cref{thm:no-free}, the evaluation transfer
cannot in general be recovered from the training
trajectory alone. An independent evaluation
sample is one means of obtaining the missing
information. Other procedures, such as
cross-fitting, may serve the same purpose under
appropriate statistical hypotheses.

The essential requirement is not the use of a
particular validation procedure, but the
availability of information relating the
training dynamics to the evaluation law.
No choice of task subspace, in the absence of
such information, removes the identifiability
obstruction.

\clearpage

\section{Conclusion}
\label{sec:conclusion}
\label{sec:novelty}

The starting point of this work is the reservoir
theorem of Litman and Guo~\cite{litmanguo2026}, which
identifies a subspace of training-output directions
that cannot contribute to the evaluation output.
We have extended this description to general
evaluation laws and studied the evolution of the
effective signal channel relative to a given task.
We have also examined the relation between the
visibility of a training direction and its
contribution to the decrease of population risk.

Let \(g=g(s)\), \(W=\W(s,T)\), and \(G=\G(T,s)\).
If \(g\notin\Ker W\), one has
\[
    \frac{\norm{C_\mu^{1/2}Gg}^2}{\norm g^2}
    =
    \frac{\ip{g}{Wg}}{\norm g^2}
    \frac{\norm{C_\mu^{1/2}Gg}^2}{\ip{g}{Wg}},
\]
where the two factors on the right are the mobility
of \(g\) and its transfer efficiency under \(\mu\),
respectively. The former depends only on the training
dynamics, whereas the latter also depends on the
evaluation transfer and the evaluation law.
By \cref{thm:domination}, the transfer efficiency
is bounded above by \(\beta_\mu(s,T)\).

Neither quantity determines the sign of the variation
of population risk. For squared population risk,
this variation is given by
\[
    \Risk_\mu(s)-\Risk_\mu(T)
    =
    \ip{e(s)}{C_\mu Gg}
    -
    \frac12\norm{Gg}_{C_\mu}^2.
\]
Thus, in addition to the size of the evaluation
displacement, one must take into account its
orientation relative to the residual. Much of the
preceding theory consists in studying these two
identities on task subspaces and along training
trajectories.

The bounds of \cref{thm:block-task} describe the
visibility of a direction in terms of the spectra
of the task and complementary compressions, the
coupling between them, and the position of the
direction relative to the task subspace. The
resolvent estimates of
\cref{thm:resolvent,thm:soft-task-stability} show
that the visibility operator and the extremal
spectra of its task compression admit stable
approximations without a spectral gap assumption.

Together with a lower bound for the orientation
of the evaluation displacement, the task estimates
give a lower bound for signed progress
(\cref{thm:task-signed-progress}).
Under suitable margin hypotheses, one obtains
bounds for classification error and sufficient
conditions for attaining a prescribed accuracy
under one evaluation law but not under another
(\cref{thm:classification,thm:explicit-mu-not-nu}).
The latter result uses both a lower bound for the
decrease of population risk under the first law
and an upper bound for the measure of initially
misclassified examples that can be corrected under
the second. The stability of these conditions is
considered in \cref{cor:selective-window}.

The time dependence of evaluation visibility is
described by the differential identities of
\cref{thm:endpoint-balance}, which separate the
variation of the transfer operator from that of
the signal filter. For a fixed initial time,
the exact signal spaces form an increasing
family of subspaces, whereas the spaces defined
by a positive spectral threshold may vary even
when their dimension remains constant.
This distinction leads to the notions of spectral
activation and spectral transport developed in
\cref{sec:spectral-transport}. For the instantaneous
training kernel, the corresponding motion of the
spectral subspaces gives rise to a conservative
transfer of residual energy between spectral
blocks, as shown in \cref{sec:moving-basis}.

These results are related to the study of distinct
time scales in training
\cite{kumar2024lazy,chou2026twospeeds,tan2026twoclocks}
and to that of spectral transitions
\cite{xu2026spectral}. The example of
\cref{thm:two-mode} shows that an arbitrarily long
delay between the attainment of a training-loss
tolerance and that of a population-risk tolerance
may occur even when the kernel is constant.

The results of \cref{sec:abrupt} extend this
observation to classification. We first relate
classification accuracy to the distribution of
the times at which the classification margins
change sign. We then construct a family of
two-mode gradient flows for which
\[
    t_{\rm grok}^{\mu}(\varepsilon,\delta)
    -
    t_{\rm mem}(\varepsilon)
    \longrightarrow+\infty,
    \qquad
    w_s\longrightarrow0,
\]
where \(w_s\) is the length of an interval over
which classification accuracy increases from
zero to one. Furthermore,
\[
    \Risk_\mu(t_-)-\Risk_\mu(t_+)
    \longrightarrow0.
\]
Thus an arbitrarily long delay and an arbitrarily
rapid improvement in classification accuracy
may occur within the same gradient flow, even
in the absence of feature evolution during
training. In this construction, the delay is
governed by the slow spectral mode, while the
concentration of the margin crossing times is
governed by the decay of the fast mode.

In the stochastic case, the conditional identity
of \cref{thm:drift-variance} distinguishes the
contribution of the mean update from that of its
covariance. For the linearized update model, the
latter contributes a nonnegative term to the
conditional expected population risk. Any benefit
of stochasticity of the kind studied in
\cite{acharya2026variance} must therefore be
distinguished from this instantaneous variance
contribution and examined in relation to the
subsequent dynamics.

The constructions of \cref{sec:structures} give
alternative formulations of several of the
preceding results. The interval composition laws
describe the concatenation of the dissipation
and transfer operators, while the quotient by
the reservoir and the observability interpretation
identify the space on which the dissipation form
is positive definite. Grassmannian geometry
describes the motion of the effective signal
subspaces, and representation theory permits
a decomposition of the dissipation and visibility
operators into isotypic blocks under the stated
symmetry hypotheses. The determinantal description
identifies the loci at which ranks and spectral
dimensions may change.

Several of the tools used here are classical.
In particular, the differentiation of spectral
projectors belongs to perturbation theory
\cite{kato1995}, and the estimates for the
variation of spectral subspaces are those of
Davis and Kahan~\cite{daviskahan1970}. The
quadratic risk identity and the conditional
decomposition into mean and variance are elementary
Hilbert-space identities. The contribution of
the present work lies in their application to
the dissipation and transfer operators and in
the resulting estimates for task visibility,
population risk, and classification accuracy.

The task visibility-gap estimate and the abstract
criterion for selective generalization appeared
in an earlier version of this work and are retained
in \cref{cor:task-gap,thm:abstract-mu-not-nu}.
The block estimates, the resolvent approximation,
the signed-progress bounds, and the explicit
margin criterion extend those results.

\subsection{Further questions}

The preceding results leave several questions open.
The first concerns the evolution of the dissipation
and transfer operators themselves. For a fixed
initial time \(s\), their derivatives are given by
\[
    \frac{\dd}{\dd T}\W(s,T)=Q(T;s),
    \qquad
    \frac{\dd}{\dd T}\G(T,s)=L(T;s).
\]
These equations describe the variation of the
operators along a given training trajectory,
but do not constitute a closed system for them.
Indeed, the operators on the right depend on the
Jacobians, the preconditioner, and the propagator
along the trajectory.

It would be of interest to find hypotheses on
the model and the training problem under which
these equations, possibly supplemented by a
finite number of additional variables, form a
closed system. One could then seek estimates for
the generalization time in terms of the initial
data, rather than in terms of operators evaluated
along the entire trajectory.

There is, however, a restriction on what can be
expected from such a system. The results of
\cref{thm:no-free,cor:training-only} show that
even complete knowledge of the training trajectory
does not determine evaluation visibility or
generalization time for an arbitrary evaluation
law. A closed system involving only training-side
quantities would therefore not suffice in general.
Additional hypotheses relating the training and
evaluation spaces would be necessary.

A second question concerns the choice of the task
subspace \(\Task\). Throughout the paper, this
subspace is given independently of the estimates
that are to be established on it. The symmetry
considerations of \cref{sec:task-symmetry} suggest
that, for certain classes of problems, it may be
possible to associate such a subspace intrinsically
with the structure of the target.

For instance, when the target admits a decomposition
into components associated with a group action,
one may examine whether the corresponding
decomposition of the training-output space is
preserved by the dissipation and visibility
operators. It remains to determine to what extent
an analogous construction is possible when the
symmetries are only approximate, or when no group
action is given. The task estimates of
\cref{sec:task} provide conditions with which such
a construction would have to be compatible.

The example of \cref{thm:two-mode} raises a further
asymptotic question. In that example, the difference
between the population-risk and training-loss
passage times satisfies
\[
    t_{\mathrm{grok}}(\delta)
    -
    t_{\mathrm{mem}}(\eps)
    \sim
    \frac{1}{2\kappa_s}
    \log\left(\frac{c^2b^2}{2\delta}\right),
\]
as \(\kappa_s\) tends to zero, the other quantities
being fixed. It would be of interest to determine
whether analogous asymptotic laws hold for
nonconstant kernels and for task subspaces of
higher dimension.

More generally, consider a family of training
problems indexed by \(n\), where \(n\) may denote
the sample size, the width of the model, or a
parameter governing the separation of spectral
scales. One may ask for conditions under which
\[
    t_{\mathrm{grok}}^{\mu,(n)}(\eps,\delta)
    -
    t_{\mathrm{mem}}^{(n)}(\eps)
    \longrightarrow+\infty,
\]
with both passage times finite for every \(n\).

The classification model of
\cref{thm:two-mode-classification} provides an example
in which the delay before generalization tends to
infinity while the length of the interval containing
the classification margin crossing times tends to
zero. More precisely, one has
\[
    \frac{w_s}
    {t_{\rm grok}^{\mu}(\varepsilon,\delta)
    -t_{\rm mem}(\varepsilon)}
    \sim
    \frac{1}{2\log 2}
    \exp\left(
        -\frac{\kappa_f\log 2}{2\kappa_s}
    \right),
\]
as \(\kappa_s\) tends to zero.

It remains to determine under what hypotheses an
analogous behaviour occurs for more general families
of learning problems. Let \(\tau_n\) denote the
time at which classification accuracy reaches a
prescribed lower level, and let \(w_n\) be the
time required to reach a prescribed higher level.
One may ask for conditions under which
\[
    \frac{w_n}
    {\tau_n-t_{\rm mem}^{(n)}(\varepsilon)}
    \longrightarrow0,
\]
when the denominator tends to infinity.

The crossing-time formulation of
\cref{sec:abrupt} suggests that this question
requires estimates relating the spectral scales
of the training dynamics to the concentration
of the distribution of margin crossing times.
It would be of particular interest to obtain
such estimates for evolving kernels and to
determine how the evolution of the learned
representation affects this concentration.

Another question concerns the passage from
continuous gradient flows to discrete optimization.
Most of the identities established here concern
continuous preconditioned dynamics, while the
stochastic identity of \cref{sec:stochastic}
is exact for a linearized update model. For a
nonlinear output map, a finite parameter update
contains a remainder whose contribution to
population risk must also be taken into account.

It remains to obtain estimates for the accumulation
of these remainders over intervals whose length
may be comparable to the generalization time.
The inclusion of momentum, explicit regularization,
and adaptive preconditioning would require a
corresponding extension of the state equations
and of the dissipation and transfer operators.
In particular, it would be of interest to determine
which of the reservoir and spectral transport
properties persist under these modifications.

Finally, there is the question of estimating the
population quantities from finite samples.
Let \(\widehat H_{\mu,\alpha}\) denote the
visibility operator obtained by replacing the
evaluation law \(\mu\) by an empirical measure.
One would like to obtain, under appropriate
assumptions, bounds of the form
\[
    \Pr\left(
        \norm{
            \widehat H_{\mu,\alpha}
            -
            H_{\mu,\alpha}
        }_{\op}
        >\varepsilon
    \right)
    \leq\delta.
\]
Combined with the perturbation estimates of
\cref{sec:resolvent,sec:shift}, such bounds would
give statistical control of the estimated task
spectra. Their application to selective
generalization would further require estimates
for the target orientation and the distribution
of the margins.

The matrix-free procedure of \cref{sec:protocol}
provides a means of examining these quantities
along realized training trajectories. It remains
to determine, for particular classes of learning
problems, which of the mechanisms described in
this paper account for the observed separation
between training and generalization times.

The theory developed here gives exact relations
between training dissipation, evaluation visibility,
and population risk, together with conditions
under which these relations lead to delayed or
selective generalization. A further step would
be to deduce the hypotheses of these results
from the structure of the learning problem
itself. The questions above concern precisely
this passage from a description of generalization
along a given trajectory to its determination
from the model, the task, and the initial data.

\clearpage

\appendix

\section{Notation}
\label{app:notation}

We collect here the principal notation used in the paper.
Unless otherwise specified, the operators depending on time
are evaluated along a fixed training trajectory.

\begingroup
\small
\renewcommand{\arraystretch}{1.2}

\begin{longtable}{@{}p{0.25\textwidth}p{0.70\textwidth}@{}}

\toprule
Notation & Definition or meaning\\
\midrule
\endfirsthead

\toprule
Notation & Definition or meaning\\
\midrule
\endhead

\bottomrule
\endfoot

\multicolumn{2}{@{}l}{\textit{Spaces and operators}}\\
\addlinespace[3pt]

\(\HP\) & Hilbert space of parameters\\

\(\HS\) & Space of training outputs\\

\(\HE\) & Evaluation space \(L^2(\lambda;\R^p)\)\\

\(C_\mu\) & Multiplication operator associated with \(\mu\)\\

\(J_S,J_E\) & Derivatives of the training and evaluation maps\\

\(M(t)\) & Positive semidefinite preconditioner\\

\(\Pg(t,s)\) & Propagator of the output gradient\\

\(W=\W(s,T)\) & Cumulative dissipation operator\\

\(G=\G(T,s)\) & Evaluation-transfer operator\\

\(\beta_\mu(s,T)\) & Constant in the domination estimate
\eqref{eq:domination}\\

\(P_0,P_+\) & Orthogonal projectors onto \(\Ker W\)
and \(\ran W\), respectively\\

\(\mathsf F\) & Signal filter\\

\(\Salpha(W)\) & Resolvent filter
\(W(W+\alpha\Id)^{-1}\)\\

\(H_{\mu,\mathsf F}\) & Visibility operator
\(\mathsf F G^*C_\mu G\mathsf F\)\\

\midrule
\multicolumn{2}{@{}l}{\textit{Task subspaces and visibility}}\\
\addlinespace[3pt]

\(\Task\) & Task subspace of \(\HS\)\\

\(\Pi_\Task\) & Orthogonal projector onto \(\Task\)\\

\(\Pi_\Task^\perp\) & Orthogonal projector onto
\(\Task^\perp\)\\

\(\rho_\Task(h)\) &
Task alignment \(\norm{\Pi_\Task h}^2/\norm h^2\)\\

\(\lambda_{\min}^{\Task}(H)\) &
Smallest eigenvalue of the compression of \(H\)
to \(\Task\)\\

\(\lambda_{\max}^{\Task}(H)\) &
Largest eigenvalue of the compression of \(H\)
to \(\Task\)\\

\(\kappa\) &
Coupling coefficient
\(\norm{\Pi_\Task H\Pi_\Task^\perp}_{\op}\)\\

\(\Geo_{\mu,\mathsf F}(h)\) &
Normalized visibility
\(\ip{h}{H_{\mu,\mathsf F}h}/\norm h^2\)\\

\(\Mob(h)\) &
Training mobility \(\ip{h}{Wh}/\norm h^2\)\\

\(\Eff_\mu(h)\) &
Transfer efficiency
\(\norm{Gh}_{C_\mu}^2/\ip{h}{Wh}\)\\

\midrule
\multicolumn{2}{@{}l}{\textit{Population risk and time dependence}}\\
\addlinespace[3pt]

\(g(t)\) & Gradient with respect to the training outputs\\

\(g_s\) & Initial gradient \(g(s)\)\\

\(f^\star\) & Evaluation target function\\

\(e(t)\) & Evaluation residual
\(U_E(\theta(t))-f^\star\)\\

\(\Loss_S(t)\) & Empirical training loss\\

\(\Risk_\mu(t)\) & Squared population risk
\(\frac12\norm{e(t)}_{C_\mu}^2\)\\

\(\Prog_\mu(s,T)\) & Signed progress
\(\Risk_\mu(s)-\Risk_\mu(T)\)\\

\(A_\mu(t)\) & Classification accuracy under \(\mu\)\\

\(\Err_\mu(t)\) & Classification error under \(\mu\)\\

\(m_\nu\) & Measure \(\nu(B_\nu)\) of the set of
initially misclassified examples for which a competing
class exceeds the correct class by a margin of at
least \(\gamma_\nu\) in \cref{thm:explicit-mu-not-nu}\\

\(t_{\rm mem}(\varepsilon)\) & Training-loss passage time\\

\(t_{\rm grok}^{\mu}(\varepsilon,\delta)\) &
Accuracy passage time after memorization\\

\(Q(T;s)\) & Derivative \(\partial_T\W(s,T)\)\\

\(L(T;s)\) & Derivative \(\partial_T\G(T,s)\)\\

\(\Ralpha(T)\) & Resolvent
\((W(T)+\alpha\Id)^{-1}\)\\

\midrule
\multicolumn{2}{@{}l}{
\textit{Spectral delay and classification margins}}\\
\addlinespace[3pt]

\(\kappa_f,\kappa_s\) &
Fast and slow eigenvalues of the training kernel
in the two-mode models of
\cref{sec:two-mode,sec:abrupt}\\

\(a,b\) & Components of the training target
in the two-mode models\\

\(c\) & Evaluation scaling coefficient in
\cref{thm:two-mode}\\

\(u_1(w),u_2(w)\) &
Training outputs associated with the fast and slow
spectral modes, respectively\\

\(t_{\rm grok}(\delta)\) &
Population-risk passage time in \cref{thm:two-mode}\\

\(m(t,z)\) & Signed classification margin at
evaluation input \(z\) and time \(t\)\\

\(m_\infty(z)\) & Limiting signed margin in
\cref{prop:accuracy-cdf}\\

\(d(z)\) & Amplitude of the exponentially decaying
component of the margin in \cref{prop:accuracy-cdf}\\

\(\tau(z)\) & Margin crossing time, taken to be zero
for examples with nonnegative initial margin in
\cref{prop:accuracy-cdf}\\

\(\nu_\mu\) & Image measure \(\tau_*\mu\) describing
the distribution of margin crossing times\\

\(p_\mu(t)\) & Density of the crossing-time distribution
when \(\nu_\mu\) is absolutely continuous\\

\(\mathcal Z_E\) & Evaluation input space
\([-2,-1]\cup[1,2]\) in the two-mode
classification model\\

\(v(t)\) & Slow-mode contribution
\(1-2e^{-\kappa_s t}\) in
\cref{thm:two-mode-classification}\\

\(q(t)\) & Fast-mode contribution
\(\frac14e^{-\kappa_f t}\) in
\cref{thm:two-mode-classification}\\

\(t_\star\) & Reference crossing time
\((\log 2)/\kappa_s\)\\

\(q_s\) & Upper bound
\(\frac14e^{-\kappa_f t_\star/2}\)
for \(q(t)\) when \(t\geq t_\star/2\)\\

\(t_-,t_+\) &
Lower and upper bounds for the margin crossing times
defined in \eqref{eq:two-mode-crossing-interval}\\

\(w_s\) & Length \(t_+-t_-\) of the interval
containing all margin crossing times in
\cref{thm:two-mode-classification}\\

\(t_\eta,t_{1-\eta}\) &
First passage times after memorization for
classification accuracy levels \(\eta\) and
\(1-\eta\), respectively\\

\(D_\eta\) & Delay
\(t_\eta-t_{\rm mem}(\varepsilon)\)
before reaching accuracy level \(\eta\)\\

\(w_\eta\) & Time
\(t_{1-\eta}-t_\eta\) between the two
accuracy passage levels\\

\(\Delta\Risk_{\mu,\eta}\) &
Signed population-risk reduction
\(\Risk_\mu(t_\eta)-\Risk_\mu(t_{1-\eta})\)\\

\midrule
\multicolumn{2}{@{}l}{
\textit{Categorical, geometric, and symmetry structures}}\\
\addlinespace[3pt]

\(V,E\) & Training and evaluation spaces
\(\HS\) and \(\HE\), respectively\\

\(N\) & Dimension of \(V\)\\

\(P_{ts}\) & Propagator \(\Pg(t,s)\)\\

\(G_{ts}\) & Evaluation-transfer operator \(\G(t,s)\)\\

\(W_{st}\) & Cumulative dissipation operator \(\W(s,t)\)\\

\(\mathbf{Time}\) &
Category associated with the ordered set of times\\

\(\mathbf D_E\) & Category of triples \((P,G,W)\)\\

\(\mathfrak F\) & Functor
\(\mathbf{Time}\longrightarrow\mathbf D_E\)\\

\(q\) & Quotient map \(V\longrightarrow V/\Ker W_{st}\)\\

\(\mathcal O_{sT}\) & Observation operator\\

\(\mathcal L_{sT}\) & Transfer from observation histories
to \(E\)\\

\(\Pi_\alpha(T)\) & Spectral projector
\(\1_{(\alpha,\infty)}(W_{sT})\)\\

\(\operatorname{Gr}(r,N)\) &
Grassmannian of \(r\)-planes in \(\R^N\)\\

\(\mathcal S_T\) & Exact signal space \(\ran W_{sT}\)\\

\(\Gamma\) & Finite group acting on \(V\) and \(E\)\\

\(\mathcal D_r\) & Locus of symmetric matrices
of rank at most \(r\)\\

\(\Sigma_\alpha\) & Locus of symmetric matrices
having \(\alpha\) as an eigenvalue\\

\end{longtable}

\endgroup

The quantities \(\rho_\Task(h)\),
\(\Geo_{\mu,\mathsf F}(h)\), and \(\Mob(h)\)
are defined for \(h\ne0\), whereas
\(\Eff_\mu(h)\) is defined for \(h\notin\Ker W\).
The quantities \(w_\eta\) and
\(\Delta\Risk_{\mu,\eta}\) are considered when both
accuracy passage times are finite, and the ratio
\(w_\eta/D_\eta\) is considered when \(D_\eta>0\).

Certain symbols have a local meaning within individual
sections. In particular, the coupling coefficient
\(\kappa\) is distinct from the kernel eigenvalues
\(\kappa_f,\kappa_s\), and the scalar function \(q(t)\)
in the two-mode classification model is distinct
from the quotient map \(q\) used in the categorical
formulation.


\clearpage

















\begin{thebibliography}{99}
\setlength{\itemsep}{0pt}
\setlength{\parsep}{0pt}
\setlength{\baselineskip}{0.92\baselineskip}

\bibitem{acharya2026variance}
P.~Acharya and H.~Dhakal,
\emph{Grokking as a variance-limited phase transition: Spectral gating and
the epsilon-stability threshold},
arXiv:2603.15492, 2026.

\bibitem{bartlett2017margin}
P.~L.~Bartlett, D.~J.~Foster, and M.~J.~Telgarsky,
\emph{Spectrally-normalized margin bounds for neural networks},
Advances in Neural Information Processing Systems \textbf{30} (2017),
6240--6249.

\bibitem{bartlett2020benign}
P.~L.~Bartlett, P.~M.~Long, G.~Lugosi, and A.~Tsigler,
\emph{Benign overfitting in linear regression},
Proceedings of the National Academy of Sciences \textbf{117} (2020),
30063--30070.

\bibitem{belkin2019reconciling}
M.~Belkin, D.~Hsu, S.~Ma, and S.~Mandal,
\emph{Reconciling modern machine-learning practice and the classical
bias--variance trade-off},
Proceedings of the National Academy of Sciences \textbf{116} (2019),
15849--15854.

\bibitem{bendokat2024grassmann}
T.~Bendokat, R.~Zimmermann, and P.-A.~Absil,
\emph{A Grassmann manifold handbook: Basic geometry
and computational aspects},
Advances in Computational Mathematics \textbf{50}
(2024), no.~1, Article 6.
\url{https://doi.org/10.1007/s10444-023-10090-8}.
Corrigendum, \textbf{50} (2024), no.~2, Article 20.
\url{https://doi.org/10.1007/s10444-024-10115-w}.

\bibitem{bhatia1997}
R.~Bhatia,
\emph{Matrix Analysis},
Graduate Texts in Mathematics 169,
Springer, New York, 1997.

\bibitem{bordelon2020spectrum}
B.~Bordelon, A.~Canatar, and C.~Pehlevan,
\emph{Spectrum dependent learning curves in kernel regression and wide
neural networks},
Proceedings of the 37th International Conference on Machine Learning,
PMLR \textbf{119} (2020), 1024--1034.

\bibitem{bruns1988determinantal}
W.~Bruns and U.~Vetter,
\emph{Determinantal Rings},
Lecture Notes in Mathematics 1327,
Springer, Berlin, 1988.
\url{https://doi.org/10.1007/BFb0080378}.

\bibitem{canatar2021alignment}
A.~Canatar, B.~Bordelon, and C.~Pehlevan,
\emph{Spectral bias and task-model alignment explain generalization in
kernel regression and infinitely wide neural networks},
Nature Communications \textbf{12} (2021), Article 2914.

\bibitem{chizat2019lazy}
L.~Chizat, E.~Oyallon, and F.~Bach,
\emph{On lazy training in differentiable programming},
Advances in Neural Information Processing Systems
\textbf{32} (2019), 2937--2947.

\bibitem{chou2026twospeeds}
C.-N.~Chou, O.~Uzdelewicz, N.-C.~Chiu, Y.-Y.~Yang, and S.~Chung,
\emph{Two speeds of learning: A representation--readout decomposition of
grokking and double descent},
arXiv:2605.27078, 2026.

\bibitem{daviskahan1970}
C.~Davis and W.~M.~Kahan,
\emph{The rotation of eigenvectors by a perturbation. III},
SIAM Journal on Numerical Analysis \textbf{7}
(1970), no.~1, 1--46.
\url{https://doi.org/10.1137/0707001}.

\bibitem{davis2009systems}
J.~M.~Davis, I.~A.~Gravagne, B.~J.~Jackson, and R.~J.~Marks II,
\emph{Controllability, observability, realizability, and stability of
dynamic linear systems},
Electronic Journal of Differential Equations \textbf{2009} (2009),
no.~37, 1--32.
arXiv:0901.3764.

\bibitem{douglas1966}
R.~G.~Douglas,
\emph{On majorization, factorization, and range
inclusion of operators on Hilbert space},
Proceedings of the American Mathematical Society
\textbf{17} (1966), no.~2, 413--415.
\url{https://doi.org/10.1090/S0002-9939-1966-0203464-1}.

\bibitem{fong2019backprop}
B.~Fong, D.~I.~Spivak, and R.~Tuy\'eras,
\emph{Backprop as functor: A compositional perspective on supervised
learning},
Proceedings of the 34th Annual ACM/IEEE Symposium on Logic in
Computer Science (LICS), 2019, 1--13.
\url{https://doi.org/10.1109/LICS.2019.8785665}.
arXiv:1711.10455.

\bibitem{hardt2016stability}
M.~Hardt, B.~Recht, and Y.~Singer,
\emph{Train faster, generalize better: Stability of stochastic gradient
descent},
Proceedings of the 33rd International Conference on Machine Learning,
PMLR \textbf{48} (2016), 1225--1234.

\bibitem{jacot2018ntk}
A.~Jacot, F.~Gabriel, and C.~Hongler,
\emph{Neural tangent kernel: Convergence and generalization in neural
networks},
Advances in Neural Information Processing Systems \textbf{31} (2018),
8571--8580.

\bibitem{kato1995}
T.~Kato,
\emph{Perturbation Theory for Linear Operators},
2nd ed., Classics in Mathematics,
Springer, Berlin--Heidelberg, 1995.
\url{https://doi.org/10.1007/978-3-642-66282-9}.

\bibitem{kumar2024lazy}
T.~Kumar, B.~Bordelon, S.~J.~Gershman, and C.~Pehlevan,
\emph{Grokking as the transition from lazy to rich
training dynamics},
International Conference on Learning Representations,
2024, 23010--23035.

\bibitem{lee2019wide}
J.~Lee, L.~Xiao, S.~S.~Schoenholz, Y.~Bahri, R.~Novak,
J.~Sohl-Dickstein, and J.~Pennington,
\emph{Wide neural networks of any depth evolve as linear models under
gradient descent},
Advances in Neural Information Processing Systems \textbf{32} (2019),
8572--8583.

\bibitem{litmanguo2026}
E.~Litman and G.~Guo,
\emph{A theory of generalization in deep learning},
arXiv:2605.01172v1, 2026.
\url{https://arxiv.org/abs/2605.01172v1}.

\bibitem{liu2022effective}
Z.~Liu, O.~Kitouni, N.~S.~Nolte, E.~J.~Michaud,
M.~Tegmark, and M.~Williams,
\emph{Towards understanding grokking: An effective
theory of representation learning},
Advances in Neural Information Processing Systems
\textbf{35} (2022), 34651--34663.
\url{https://doi.org/10.52202/068431-2511}.

\bibitem{nakkiran2020deep}
P.~Nakkiran, G.~Kaplun, Y.~Bansal, T.~Yang, B.~Barak, and I.~Sutskever,
\emph{Deep double descent: Where bigger models and more data hurt},
International Conference on Learning Representations, 2020.

\bibitem{nanda2023progress}
N.~Nanda, L.~Chan, T.~Lieberum, J.~Smith, and J.~Steinhardt,
\emph{Progress measures for grokking via mechanistic interpretability},
International Conference on Learning Representations, 2023.

\bibitem{power2022grokking}
A.~Power, Y.~Burda, H.~Edwards, I.~Babuschkin, and V.~Misra,
\emph{Grokking: Generalization beyond overfitting on small algorithmic
datasets},
arXiv:2201.02177, 2022.

\bibitem{rubin2024phase}
N.~Rubin, I.~Seroussi, and Z.~Ringel,
\emph{Grokking as a first order phase transition in two layer networks},
International Conference on Learning Representations, 2024.

\bibitem{soudry2018implicit}
D.~Soudry, E.~Hoffer, M.~S.~Nacson, S.~Gunasekar, and N.~Srebro,
\emph{The implicit bias of gradient descent on separable data},
Journal of Machine Learning Research \textbf{19} (2018),
no.~70, 1--57.

\bibitem{tan2026twoclocks}
H.~Tan, K.~Gai, and S.~Zhang,
\emph{Deciphering two training clocks in grokking via deep linear network
theory with conditional ReLU reduction},
arXiv:2606.05863, 2026.

\bibitem{telemanrepresentation}
C.~Teleman,
\emph{Representation Theory},
Cambridge lecture notes, Lent Term, 2005.
\url{https://math.berkeley.edu/~teleman/math/RepThry.pdf}.

\bibitem{tian2025li2}
Y.~Tian,
\emph{$\mathbf{Li}_2$: A framework on dynamics of feature emergence
and delayed generalization},
International Conference on Learning Representations, 2026.
arXiv:2509.21519.

\bibitem{xu2026spectral}
Y.~Xu,
\emph{Spectral edge dynamics: An analytical--empirical study of phase
transitions in neural network training},
arXiv:2603.28964, 2026.

\bibitem{zhang2017rethinking}
C.~Zhang, S.~Bengio, M.~Hardt, B.~Recht, and O.~Vinyals,
\emph{Understanding deep learning requires rethinking generalization},
International Conference on Learning Representations, 2017.

\end{thebibliography}
\end{document}